\documentclass{article} 
\usepackage{iclr2027_conference,times}

\usepackage{amsmath,amsfonts,bm}

\def\eqref#1{equation~\ref{#1}}

\def\1{\bm{1}}

\DeclareMathAlphabet{\mathsfit}{\encodingdefault}{\sfdefault}{m}{sl}
\SetMathAlphabet{\mathsfit}{bold}{\encodingdefault}{\sfdefault}{bx}{n}

\usepackage{hyperref}
\usepackage{enumitem}
\usepackage{url}
\usepackage{graphicx}
\usepackage{subcaption}
\usepackage{caption}
\usepackage{booktabs}
\usepackage{multirow}
\usepackage{makecell}
\usepackage{threeparttable}
\usepackage{adjustbox}
\usepackage{xcolor}
\usepackage{booktabs}
\usepackage[table]{xcolor}
\usepackage{xcolor}
\definecolor{interpcolor}{HTML}{3E648D}
\definecolor{extrapcolor}{HTML}{A85F32}

\title{Cross-View Time-Frequency Diffusion for Extreme-Aware Time-Series Generation}

\author{
Lin Jiang \\
Department of Computer Science \\
Florida State University \\
Tallahassee, FL 32306, USA \\
\texttt{lin.jiang@fsu.edu} \\
\And
Rongchao Xu \\
Department of Computer Science \\
Florida State University \\
Tallahassee, FL 32306, USA \\
\texttt{rx21a@fsu.edu} \\
\And
Guang Wang\thanks{Corresponding author.} \\
Department of Computer Science \\
Florida State University \\
Tallahassee, FL 32306, USA \\
\texttt{guang@cs.fsu.edu}
}

\iclrfinalcopy 
\begin{document}

\maketitle
\lhead{}

\begin{abstract}
Generating realistic time-series requires faithfully reproducing rare but high-impact extreme events. Existing approaches often rely on specialized representations or extreme-focused training objectives, which may restrict the diversity of captured extremes or compromise overall distributional fidelity. We propose \textbf{Cross-View Time-Frequency Diffusion (CrossTF)}, a framework that exploits discrepancies between time- and frequency-view denoisers for extreme-aware time-series generation. Our key empirical finding is \emph{extreme-sensitive cross-view asymmetry}: moving along the direction connecting the two views' clean-sample estimates substantially changes extreme fidelity while having limited effects on overall generation quality. We further identify a late-stage \emph{cross-view window} in which this discrepancy stabilizes and cross-view adjustments have a particularly strong effect on extreme fidelity. These findings motivate a staged optimization strategy in CrossTF: after training and freezing both denoisers, it only needs to learn step-specific fusion parameters within this window using an Extreme Value Theory-based objective under an energy-distance constraint. This formulation supports interpolation between and extrapolation beyond the view-specific estimates while limiting degradation in overall generation quality. Extensive experiments on seven real-world datasets against 11 baselines show that CrossTF improves extreme fidelity by 26.1\% and overall performance by 7.4\% relative to the strongest baselines. These results highlight cross-view discrepancy as an effective basis for low-dimensional optimization in extreme-aware time-series generation.
\end{abstract}

\section{Introduction}\label{sec:introduction}

Time-series generation has emerged as an important research direction
~\citep{yuan2024diffusion,naiman2024utilizing,naimangenerative,rousseau2025forging} because it can support a wide range of applications, including data augmentation~\citep{lou2023pcf}, system simulation~\citep{nikitin2024tsgm}, and stress testing~\citep{rizzato2022stress}. Within this field, extreme-aware time-series generation has attracted increasing attention~\citep{galib2024fide,pandey2025heavy}, driven by the need to faithfully reproduce high-impact events such as severe heatwaves, intense rainfall, and abrupt demand surges~\citep{gu2025beyond}. However, these events are often rare and temporally localized, making their distinctive patterns particularly susceptible to smoothing or distortion during generation~\citep{jiang2026e4gen}.

Existing approaches to improving extreme-aware time-series generation broadly fall into two categories. \textbf{Representation-based methods} leverage specialized representations, such as high-frequency components in FIDE~\citep{galib2024fide} and time-domain shapelets in E4GEN~\citep{jiang2026e4gen}. While effective for certain patterns, these predefined representations introduce inductive biases that may limit their ability to capture the diversity of extreme events. \textbf{Objective-based methods}, such as ExGAN~\citep{bhatia2021exgan} and Tail-GAN~\citep{cont2026tail}, emphasize extremes through customized training objectives. However, explicitly prioritizing sparse extremes may divert optimization from modeling the full data distribution, potentially reducing overall generation fidelity. This trade-off motivates an alternative optimization strategy that improves the generation fidelity of extremes without sacrificing the overall generation fidelity.

\begin{figure*}[b]
\vspace{-5pt}
    \centering
    \begin{minipage}[t]{0.29\linewidth}
        \centering
        \includegraphics[width=\linewidth]{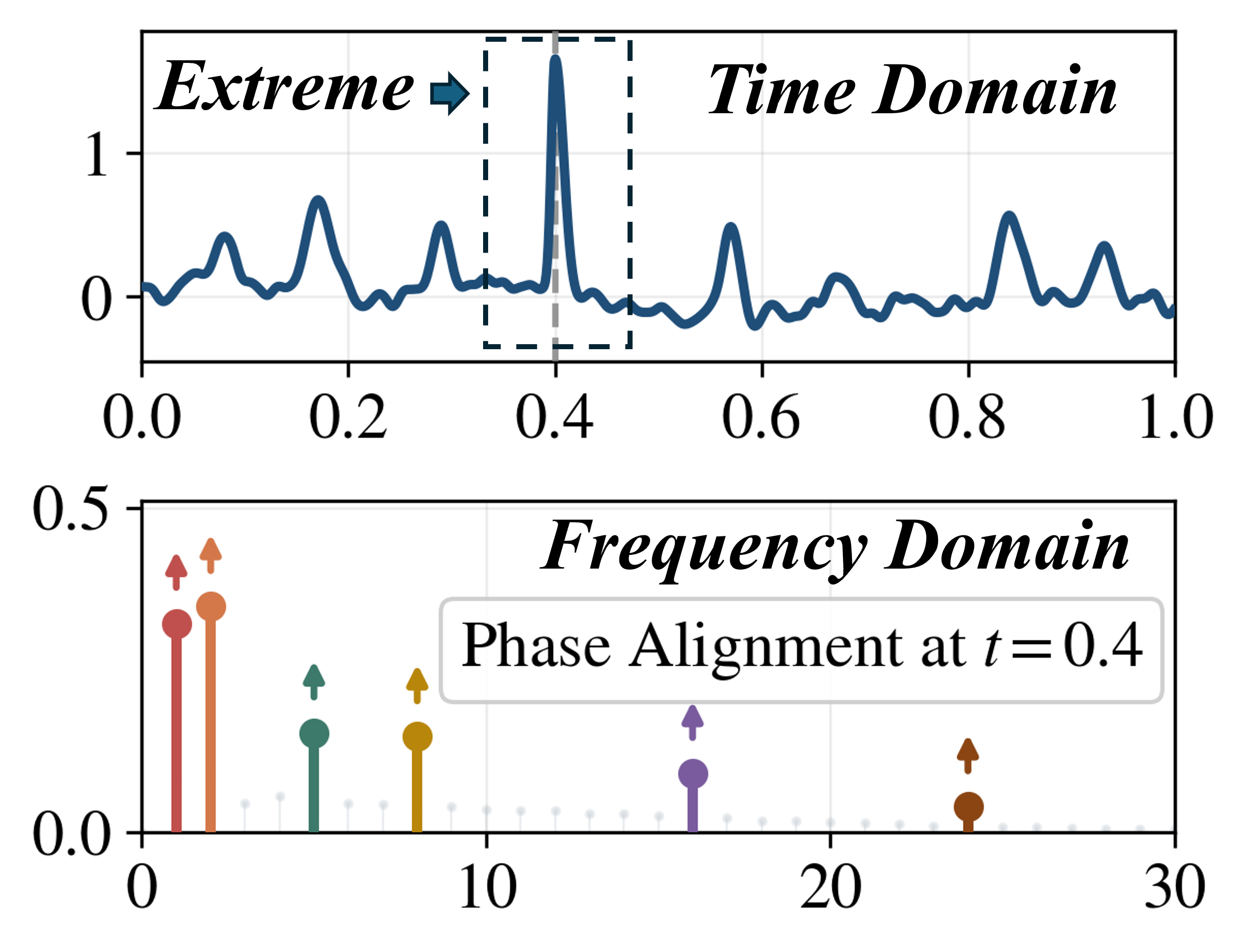}
        \captionof{figure}{Extreme Representations}
        \label{fig:extreme_representation}
    \end{minipage}
    \hfill
    \begin{minipage}[t]{0.69\linewidth}
        \centering
        \includegraphics[width=\linewidth]{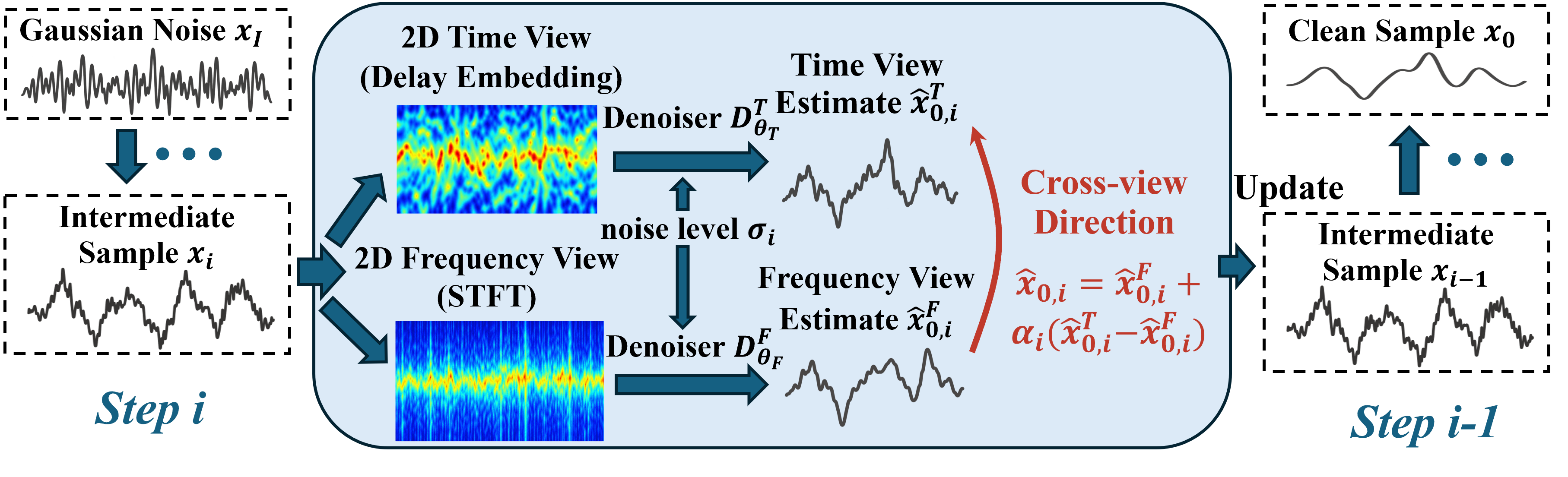}
        \captionof{figure}{Pipeline of CrossTF}
        \label{fig:CrossTF}
    \end{minipage}%
    \vspace{-10pt}
\end{figure*}

To this end, we investigate whether discrepancies between time- and frequency-domain generative models can guide optimization toward better extreme-aware generation~\citep{kollovieh2023predict,crabbe2024time}. Although these models operate on representations that are information-equivalent under an invertible Fourier transform~\citep{shatkay1995fourier}, they encode extremes through distinct structures. An abrupt peak is localized in time, while its spectral representation distributes this structure across the amplitudes and phases of multiple frequency components, whose constructive superposition produces the peak, as illustrated in Figure~\ref{fig:extreme_representation}. These structural differences may give the time-view and frequency-view complementary strengths in capturing extremes, suggesting that their \textbf{cross-view discrepancy} may encode informative signals beyond any single view. This prompts a fundamental question: \textit{Can cross-view discrepancy define an optimization direction that improves extreme modeling while largely preserving overall generation quality?}


To investigate this question, we develop \textbf{Cross-View Time-Frequency Diffusion (CrossTF)}, a framework that characterizes cross-view discrepancy and exploits it for extreme-aware time-series generation. We present the framework through three key questions.

\textbf{How is the cross-view direction defined?} At each denoising step along a shared diffusion trajectory, CrossTF constructs two views of the same intermediate sample using delay embedding~\citep{packard1980geometry} for the time view and STFT~\citep{allen1977unified} for the frequency view. Each view-specific denoiser produces a clean-sample estimate that is then mapped back to the time domain. We define the vector from the frequency-view estimate to the time-view estimate as the \textbf{cross-view direction}, as in Figure~\ref{fig:CrossTF}. A step-specific fusion parameter $\alpha_i$ controls movement along this direction, with $\alpha_i=0$ and $\alpha_i=1$ denoting the frequency- and time-view estimates.


\textbf{Why and when is this direction useful?} We probe the cross-view direction by varying $\alpha_i$ between the pure frequency view and time view, revealing a pronounced asymmetry: \textit{the denoising loss remains nearly flat, while metrics of extreme fidelity change substantially}. 
Across seven datasets, the median relative variation of the extreme-fidelity metric is 16.74$\times$ that of the denoising loss and 11.56$\times$ that of the overall-quality metric. This asymmetry creates an opportunity to improve extreme fidelity with limited impact on overall generation quality. 

We further examine how the cross-view discrepancy evolves during denoising. In the early stages, the discrepancy is largely driven by noise and does not yet reliably reflect the systematic differences between the two views. 
As denoising progresses, the discrepancy stabilizes as the influence of noise diminishes, increasingly reflecting view-specific differences.
Interestingly, the onset of this stable stage is highly consistent across the evaluated datasets. We define this late-stage interval as the \textbf{cross-view window}. Further experiments show that optimization within this window has a substantially stronger effect on extreme fidelity than optimization outside it. These motivate optimizing extreme fidelity along the cross-view direction specifically within this window.

\textbf{How do we optimize along this direction?} CrossTF uses a staged strategy that decouples distribution learning from extreme-focused optimization. We first train the time-view and frequency-view denoisers using the standard MSE-based denoising objective and then freeze their parameters. Then, a lightweight constrained cross-view optimizer learns the fusion parameters ${\alpha_i}$ within the cross-view window using an extreme-focused objective inspired by Extreme Value Theory (EVT)~\citep{coles2001introduction}. Notably, CrossTF allows $\alpha_i$ to take values outside $[0,1]$, enabling extrapolation beyond the two view-specific estimates. This flexibility is particularly useful when both views overestimate or both underestimate the extremes, as interpolation alone may not provide a sufficient correction.

The key contributions of this paper are summarized as follows:


(i) \textbf{Conceptually}, we identify a time-frequency cross-view optimization direction for extreme-aware modeling to which the standard denoising loss is largely insensitive. Along this cross-view direction, we observe a pronounced asymmetry: the denoising loss remains nearly flat, while metrics of extreme fidelity change substantially.
This finding reveals an underexplored degree of freedom for improving extreme-aware time-series generation while largely preserving overall generation quality.

(ii) \textbf{Technically}, we propose CrossTF, a diffusion framework that turns the time--frequency discrepancy into an optimizable cross-view direction. CrossTF identifies a \textbf{cross-view window} and uses a constrained EVT-based optimizer to learn step-specific movement along the cross-view direction, supporting both interpolation and extrapolation while keeping the pretrained denoisers frozen.

(iii) \textbf{Experimentally}, we evaluate CrossTF on seven real-world datasets against 11 representative baselines using 10 metrics. Across all 770 (dataset, metric, baseline) comparisons, CrossTF wins 730 (94.8\%). Relative to the strongest baseline on each dataset, it improves extreme fidelity by 26.1\% and overall performance by 7.4\% on average. These results demonstrate the effectiveness and broad applicability of CrossTF. Our code and resources are available at \href{https://github.com/LinJiang18/CrossTF}{\textcolor{blue}{\textbf{Anonymous Repository}}}.

\section{Preliminaries}\label{sec:preliminary}

\subsection{Extremes in Time Series} \label{sec:extreme_definition}
We first examine how extremes manifest in time series. Across different scenarios, extremes take diverse forms, such as impulse spikes, sustained high values, and abrupt drops, with varying durations and local structures, as shown in Figure~\ref{fig:representative_extreme_events} in Appendix~\ref{appendix:representative_extreme_event}. Such diversity makes it challenging to characterize extremes using a single event representation. The Peaks-Over-Threshold (POT) framework in Extreme Value Theory (EVT) provides a general approach to identifying extremes as exceedances above a high threshold. However, this pointwise definition primarily captures extreme magnitudes and overlooks extreme changes between consecutive observations. To account for both, we characterize extremes along two complementary dimensions: \textbf{extreme magnitude} and \textbf{extreme change}, as illustrated in Figure~\ref{fig:extreme_sample}.

Given a time series $\mathbf{x}=\{x_t\}_{t=1}^{T}$, we define normalized scores for the two dimensions:
\begin{equation}
    s_t^{\mathrm{mag}}=|x_t-\mathrm{median}(\mathbf{x})|/\mathrm{std}(\mathbf{x}), \qquad
s_t^{\mathrm{chg}}=|x_t-x_{t-1}|/\mathrm{std}(\mathbf{x}).
\end{equation}
The former measures deviations from the sequence baseline, capturing both unusually high and unusually low values, while the latter measures abrupt temporal transitions that magnitude alone may not capture. For each score type, we identify extremes as exceedances above an upper-tail threshold estimated from the real-data scores, using the 95th percentile as a representative setting. These two complementary dimensions guide extreme-aware optimization throughout CrossTF.


\subsection{Time and Frequency Views} \label{sec:temporal_frequency_views}
\begin{figure*}[t]
    \centering
    \captionsetup{
        justification=centering,
        singlelinecheck=false,
        skip=2pt
    }

    \begin{minipage}[t]{0.325\textwidth}
        \centering
        \includegraphics[width=\linewidth]{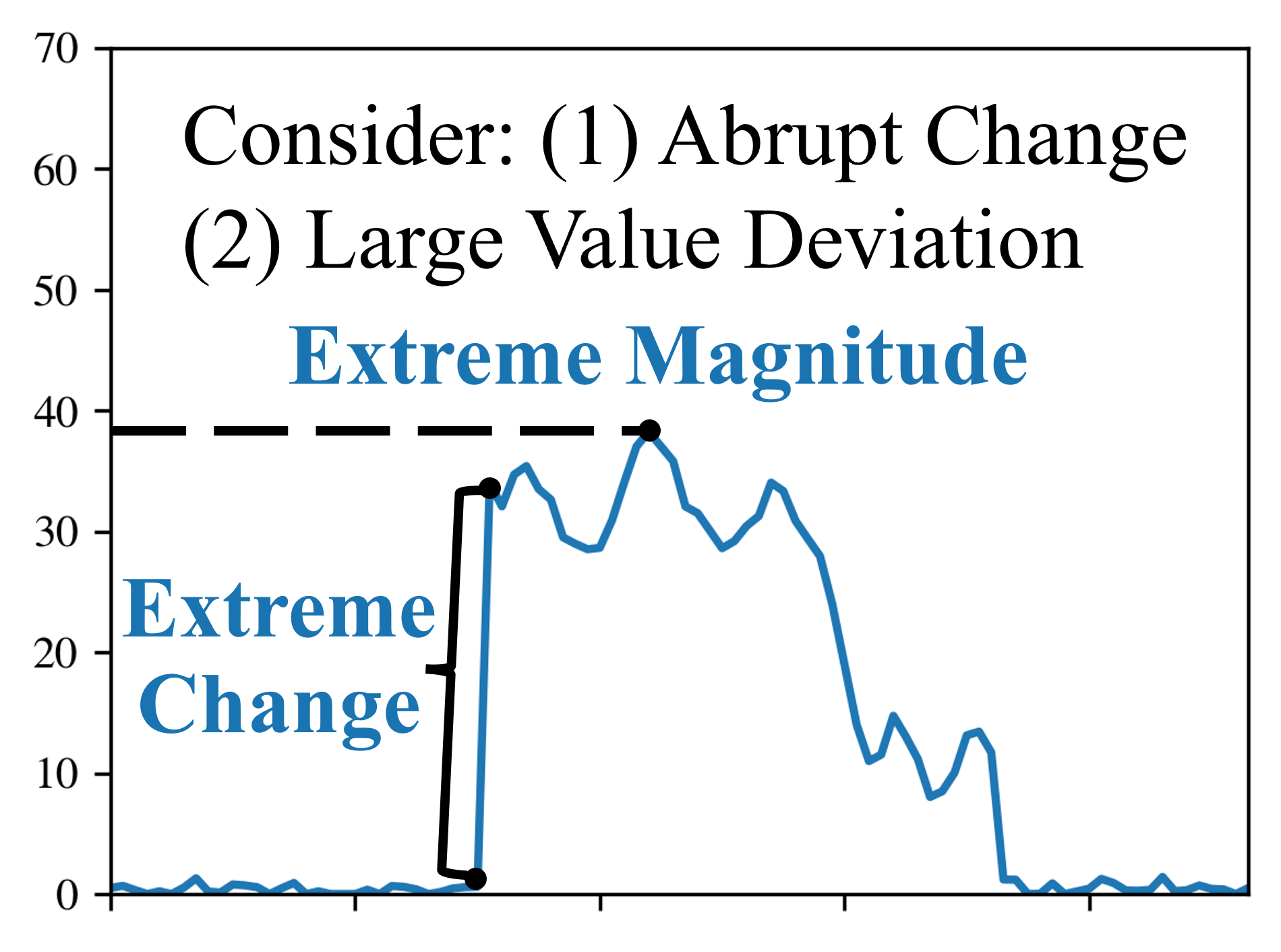}
        \captionof{figure}{Two Different Aspects of Extremes}
        \label{fig:extreme_sample}
    \end{minipage}
    \hfill
    \begin{minipage}[t]{0.325\textwidth}
        \centering
        \includegraphics[width=\linewidth]{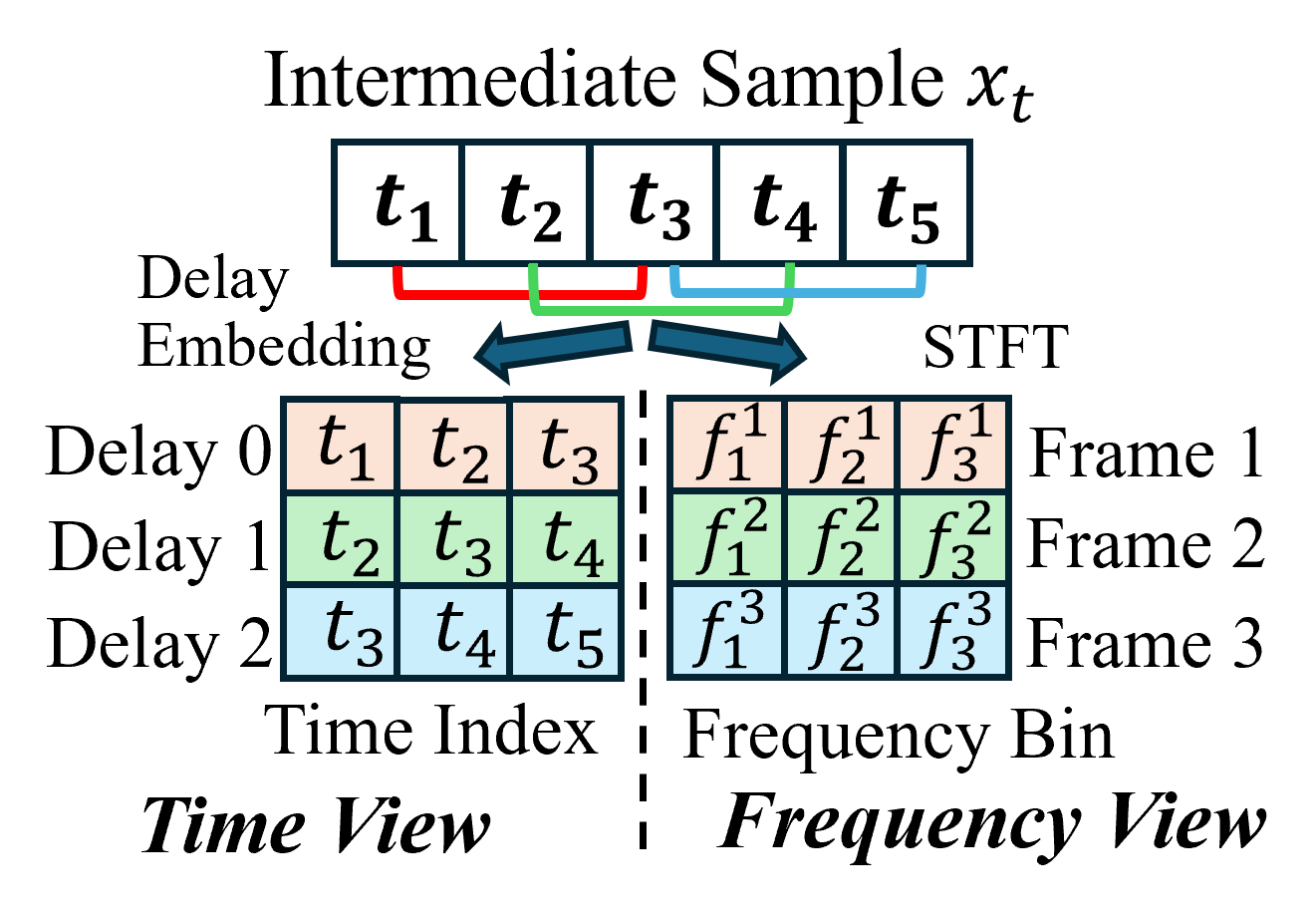}
        \captionof{figure}{Two-Dimensional Time and Frequency Views}
        \label{fig:TFView}
    \end{minipage}
    \hfill
    \begin{minipage}[t]{0.325\textwidth}
        \centering
        \includegraphics[width=\linewidth]{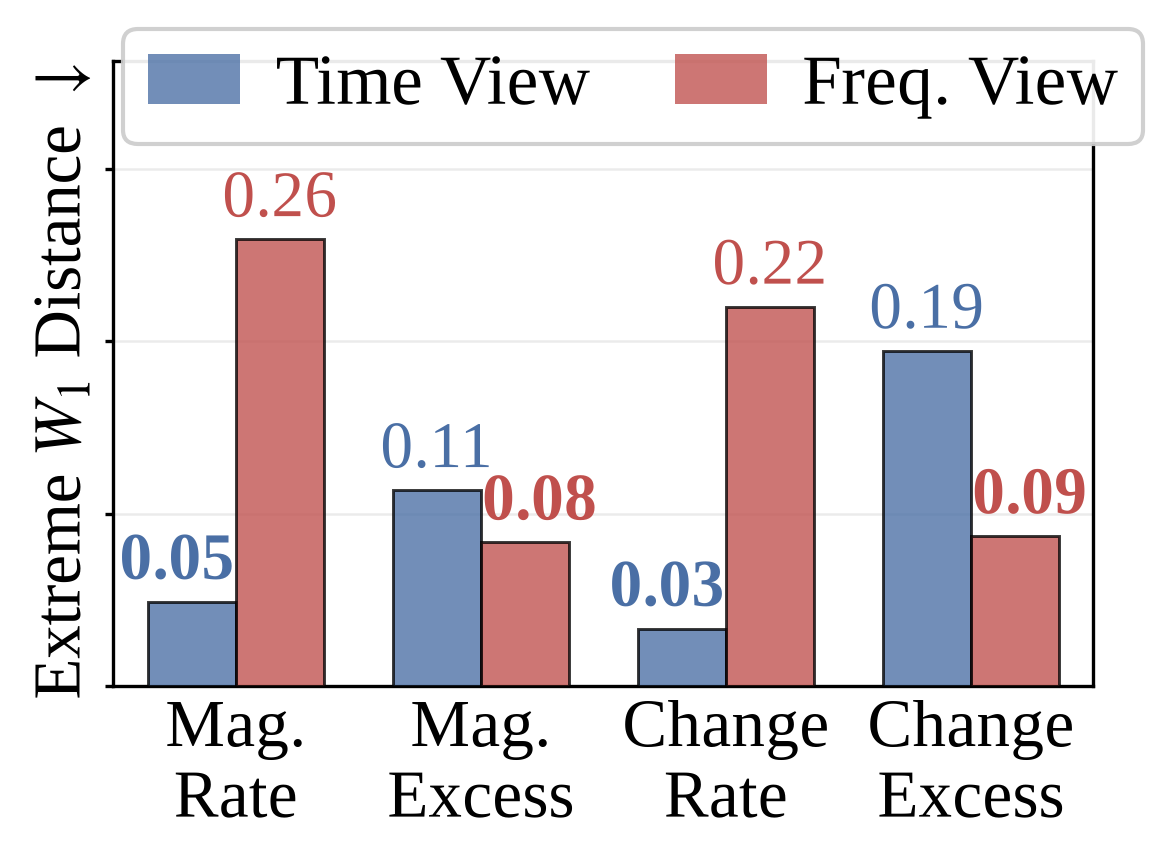}
        \captionof{figure}{Complementary Advantages for Extreme-Aware Generation}
        \label{fig:TimeFrequencyAdvantage}
    \end{minipage}

    \vspace{-15pt}
\end{figure*}

As illustrated in Figure~\ref{fig:CrossTF}, CrossTF maintains the intermediate sample $\mathbf{x}_i$ as a one-dimensional (1D) time-domain sequence throughout the denoising process. At each denoising step from $\mathbf{x}_i$ to $\mathbf{x}_{i-1}$, $\mathbf{x}_i$ is transformed into both time and frequency views, yielding view-specific representations $X_i^T$ and $X_i^F$. 
To ensure comparable computational budgets across the two views, both are encoded as equally sized two-dimensional (2D) representations~\citep{naiman2024utilizing}. Specifically, as shown in Figure~\ref{fig:TFView}, the time view is constructed using delay embedding~\citep{packard1980geometry}, where each row represents the sequence under a different temporal delay. For example, with a one-step delay, the first row contains $(t_1,t_2,t_3)$, while the next contains $(t_2,t_3,t_4)$. In parallel, the frequency view is constructed using the short-time Fourier transform (STFT)~\citep{allen1977unified}, which decomposes each local time frame into frequency components. For example, the first frame $(t_1,t_2,t_3)$ is mapped to its frequency coefficients $(f^{1}_1,\dots,f^{1}_K)$, where $f^{k}_j$ denotes the coefficient at frequency bin $j$ in frame $k$, and $K$ is the number of frequency bins. The transformation is then applied to each subsequent overlapping frame.

We further examine the complementary advantages of the two views for extreme-aware generation. The time view directly represents extreme magnitudes and local changes, whereas the frequency view captures extremes through distributed cross-frequency patterns. Empirically, the two views exhibit complementary characteristics for extreme-aware generation across datasets. Taking the Precipitation dataset as an example, Figure~\ref{fig:TimeFrequencyAdvantage} compares the two views using four magnitude- and change-based evaluation metrics defined in Section~\ref{sec:experiment}, where a lower $W_1$ distance indicates better fidelity. The time view performs better on Magnitude Rate and Change Rate, whereas the frequency view achieves lower errors on Magnitude Excess and Change Excess. This complementarity motivates the key design in CrossTF: modeling the same intermediate sample through both views to leverage their advantages and exploiting the cross-view direction to improve extreme fidelity.

\subsection{EDM-Based Diffusion for Time-Series Generation} \label{sec:EDM_based_diffusion}
CrossTF adopts EDM~\citep{karras2022elucidating} as its underlying diffusion framework. During sampling, EDM starts from Gaussian noise $\mathbf{x}_I$ and progressively denoises it along decreasing noise levels until reaching the final sample $\mathbf{x}_0$. At each denoising step $i$, the denoiser produces a clean-sample estimate $\hat{\mathbf{x}}_{0,i}$ given the current state $\mathbf{x}_i$ and noise level $\sigma_i$. Based on this estimate, EDM computes the local direction of the probability-flow ODE and performs an Euler prediction toward the next noise level:
\begin{equation} \label{eq:edm_di}
\mathbf{d}_i=(\mathbf{x}_i-\hat{\mathbf{x}}_{0,i})/\sigma_i,
\qquad
\tilde{\mathbf{x}}_{i-1}=\mathbf{x}_i+(\sigma_{i-1}-\sigma_i)\mathbf{d}_i.
\end{equation}
Rather than using this first-order prediction, EDM applies a second-order Heun correction~\citep{hairer1993solving} when $\sigma_{i-1}>0$. It recomputes the direction at $(\tilde{\mathbf{x}}_{i-1},\sigma_{i-1})$ and averages the two directions to obtain
\begin{equation} \label{eq:xi_update}
\mathbf{x}_{i-1}
=
\mathbf{x}_i+
(\sigma_{i-1}-\sigma_i)(\mathbf{d}_i+\tilde{\mathbf{d}}_{i-1})/2.
\end{equation}
When $\sigma_{i-1}=0$ at the final step, the Heun correction is omitted and the Euler prediction~\citep{karras2022elucidating} is used directly. Together, these updates progressively transform $\mathbf{x}_I$ into $\mathbf{x}_0$.

\section{Proposed CrossTF}\label{sec:ctfdiff}

\subsection{Cross-View Direction Formulation}\label{sec:what}
CrossTF uses the discrepancy between time- and frequency-view estimates to define a direction for adjusting the clean-sample estimate $\hat{\mathbf{x}}_{0,i}$. As shown in Figure~\ref{fig:CrossTF}, at each denoising step $i$, the same intermediate sample $\mathbf{x}_i \in \mathbb{R}^{T}$ is transformed into time and frequency representations $X_i^T$ and $X_i^F$ through delay embedding and STFT, respectively, as introduced in Section~\ref{sec:temporal_frequency_views}. The view-specific denoisers $D_{\theta_T}^T$ and $D_{\theta_F}^F$ then independently produce the time- and frequency-view clean-sample estimates:
\begin{equation}  \label{eq:denoise_def}
\hat{\mathbf{x}}_{0,i}^T = D_{\theta_T}^T(X_i^T, \sigma_i) \in \mathbb{R}^{T},
\qquad
\hat{\mathbf{x}}_{0,i}^F = D_{\theta_F}^F(X_i^F, \sigma_i) \in \mathbb{R}^{T}.
\end{equation}
Here, each denoiser processes its corresponding 2D view and maps the output back to the 1D time domain through a fixed, non-learned inverse transform, making the two estimates directly comparable. We define their difference
$\Delta_i = \hat{\mathbf{x}}_{0,i}^T - \hat{\mathbf{x}}_{0,i}^F$,
as the cross-view direction from the frequency-view estimate to the time-view estimate, with the two estimates as its anchors. At each denoising step, a step-specific fusion parameter $\alpha_i$ determines the position along this direction, yielding the cross-view clean-sample estimate $\hat{\mathbf{x}}_{0,i}$:
\begin{equation}
\label{eq:cross_view_fusion}
\hat{\mathbf{x}}_{0,i}
=
\hat{\mathbf{x}}_{0,i}^F
+
\alpha_i \Delta_i
=
\hat{\mathbf{x}}_{0,i}^F
+
\alpha_i
\left(
\hat{\mathbf{x}}_{0,i}^T
-
\hat{\mathbf{x}}_{0,i}^F
\right).
\end{equation}
Under this parameterization, $\alpha_i=0$ and $1$ correspond to the pure frequency-view and time-view estimates, respectively. The resulting estimate is then passed to the EDM sampler in Section~\ref{sec:EDM_based_diffusion}, which updates $\mathbf{x}_i$ to $\mathbf{x}_{i-1}$ using Equations~\ref{eq:edm_di} and ~\ref{eq:xi_update}. Further architectural details of the view-specific denoisers are provided in Appendix~\ref{Appendix:EDM}.

\subsection{Cross-View Asymmetry and Window Identification}\label{sec:why}
We next examine why the cross-view direction is useful for extreme-aware generation and when to apply it during denoising. Although other effective directions may exist in the high-dimensional sample space, this cross-view direction offers two practical advantages.
First, the time and frequency views represent the same underlying sequence in different forms, providing complementary perspectives on its structure. 
Second, their clean-sample estimates anchor a sample-dependent direction along which adjustments can be controlled by a single scalar at each denoising step.
Beyond these advantages, our empirical analysis reveals a key property particularly relevant to extreme-focused optimization, which we term \emph{Extreme-Sensitive Cross-View Asymmetry}.

\textbf{Extreme-Sensitive Cross-View Asymmetry.} At denoising step $i$, varying the fusion parameter $\alpha_i$ moves the clean-sample estimate along the cross-view direction. Empirically, this movement produces substantial changes in extreme-fidelity metrics, while the denoising loss and overall-quality metrics vary considerably less. Thus, extreme fidelity is markedly more sensitive to movement along this direction than overall generation quality is. This differential sensitivity creates a key optimization opportunity for CrossTF: extreme-focused optimization can be performed separately along this direction after overall generation capability has stabilized, avoiding direct competition with the primary objective during joint optimization. The two objectives can thus be largely decoupled, enabling improvements in extreme fidelity with limited impact on the learned overall distribution.

\begin{table}[t]
\centering
\caption{\textbf{Extreme-sensitive cross-view asymmetry} across seven datasets.}
\label{tab:asymmetry}
\small
\renewcommand{\arraystretch}{0.93}
\begin{tabular}{lccccc}
\toprule
& \multicolumn{3}{c}{Relative Max--Min Range ($\alpha_i \in [0,1]$)} & \multicolumn{2}{c}{Ratio} \\
\cmidrule(lr){2-4}\cmidrule(lr){5-6}
Dataset & Loss & Overall & Extreme & Ext./Loss & Ext./Overall \\
\midrule
Traffic         & 4.5\%  & 8.4\%   & 290.4\% & 65.22$\times$ & 34.44$\times$ \\
Transportation  & 7.5\%  & 51.1\%  & 254.2\% & 34.11$\times$ & 4.97$\times$ \\
Temperature     & 7.3\%  & 12.6\%  & 145.6\% & 19.91$\times$ & 11.52$\times$ \\
Electricity     & 8.7\%  & 78.4\%  & 85.2\%  & 9.82$\times$  & 1.09$\times$ \\
Precipitation   & 33.3\% & 339.9\% & 160.3\% & 4.81$\times$  & 0.47$\times$ \\
ECG             & 16.7\% & 3.9\%   & 38.2\%  & 2.29$\times$  & 9.93$\times$ \\
Building        & 12.9\% & 10.8\%  & 22.8\%  & 1.77$\times$  & 2.10$\times$ \\
\midrule
\textbf{Median} & 8.7\%  & 12.6\%  & 145.6\% & \textbf{16.74}$\times$ & \textbf{11.56}$\times$ \\

\bottomrule
\end{tabular}
\vspace{-10pt}
\end{table}

To examine this asymmetry, Table~\ref{tab:asymmetry} reports results across seven datasets. Here, the \emph{Relative Max--Min Range} measures the difference between the maximum and minimum values observed for each quantity while varying $\alpha_i$ over $[0,1]$, normalized by its pure time-view value. For simplicity, we use the same $\alpha_i$ across the denoising steps considered in this analysis. We calculate the \emph{denoising loss} using the standard MSE-based one-step denoising objective in EDM-based diffusion, the \emph{overall-quality metric} using the Wasserstein distance, and the \emph{extreme-fidelity metric} using the average of four EVT-based metrics introduced in Section~\ref{sec:experiment}. Across the seven datasets, the median relative range of the extreme-fidelity metric is \textbf{16.74}$\times$ that of the denoising loss and \textbf{11.56}$\times$ that of the overall-quality metric. Notably, the extreme-fidelity metric exhibits a larger relative range than the denoising loss on all seven datasets and the overall-quality metric on  six. The only exception is Precipitation, where extremes dominate the signal variation, making changes in extreme behavior more directly reflected in the overall distribution. These findings provide empirical support for optimizing extreme fidelity along the cross-view direction. Further details are provided in Appendix~\ref{appendix:cross_view_asymmetry}.

\begin{figure*}[b]
    \centering
    \vspace{-10pt}
    \begin{minipage}[t]{0.53\linewidth}
        \vspace{0pt}
        \centering
        \includegraphics[height=4.5cm]{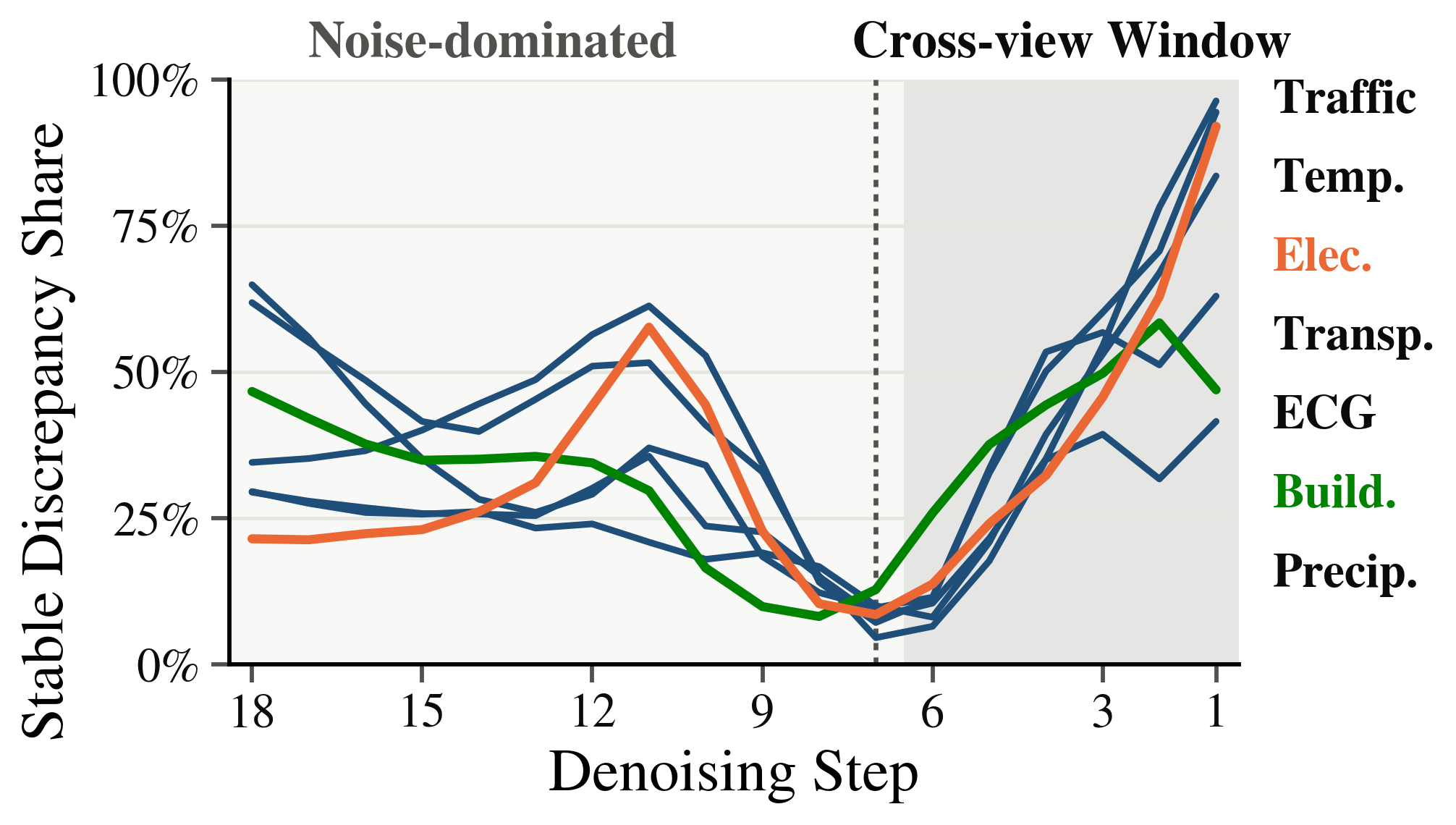}
        \captionof{figure}{Cross-View Window Identification.}
        \label{fig:cross_view_window_share}
    \end{minipage}
    \hfill
    \begin{minipage}[t]{0.45\linewidth}
        \vspace{0pt}
        \centering
        \includegraphics[height=4.5cm]{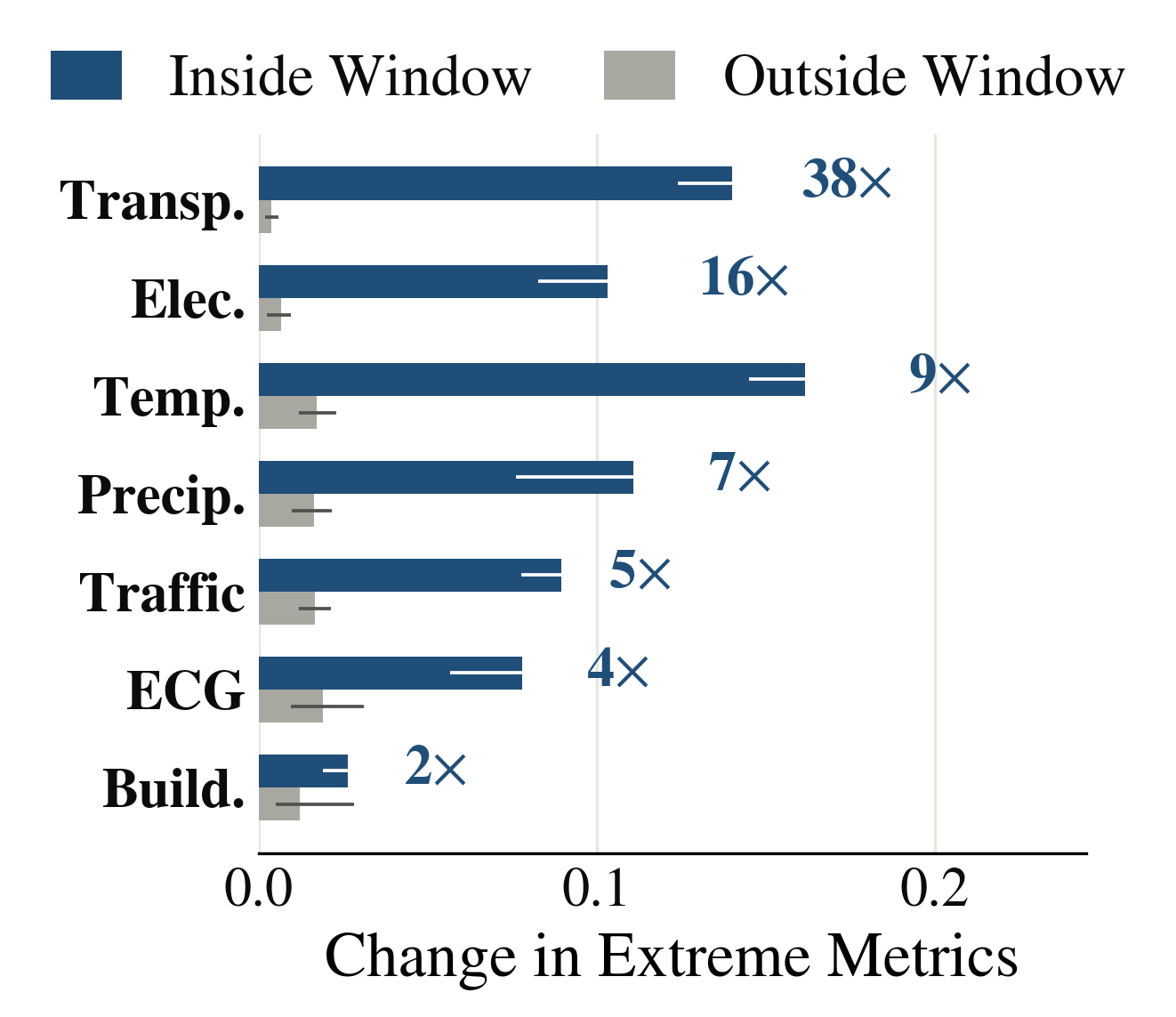}
        \captionof{figure}{Cross-View Window Sensitivity.}
        \label{fig:cross_view_window_effect}
    \end{minipage}

    \vspace{-10pt}
\end{figure*}

\textbf{Identifying the Cross-View Window.} Having established the utility of the cross-view direction for extreme-focused optimization, we next ask when to apply this optimization during denoising. Because the direction is defined by the discrepancy between the two views, we examine how this discrepancy evolves along the denoising process and identify when it stabilizes sufficiently to provide a reliable cross-view direction.


At each denoising step, we keep the real samples and noise level fixed and evaluate the cross-view discrepancy over 16 independent noise realizations for each of three independently trained seeds, with the results aggregated across seeds. We decompose the discrepancy into a \emph{stable component}, which remains consistent across noise realizations, and a \emph{deviation component}, which varies with the noise realization.
We then quantify the proportion of the total discrepancy contributed by the stable component, denoted as the \emph{Stable Discrepancy Share}. From Figure~\ref{fig:cross_view_window_share}, we observe that although the share can be high at the earliest denoising stages, it varies substantially across training seeds, indicating that the apparent stability is not reproducible and thus not yet meaningful. The share then drops to a low level, indicating that a stable cross-view component has not yet emerged. In the later stages, however, the share rises sharply and remains high, indicating that a persistent cross-view discrepancy begins to emerge across different noise realizations. Interestingly, this pattern is broadly observed across all seven datasets. Based on this observation, we define this late-stage interval as the \emph{cross-view window}, where a reliable cross-view discrepancy emerges, and restrict cross-view optimization to this window. In our setting with $I=18$ denoising steps, the cross-view window corresponds to the final six steps ($i=6,\ldots,1$), where $\sigma_i \leq 0.35$.

To further validate the effectiveness of the cross-view window, we conduct an ablation study by switching the generation view from the pure time view to the pure frequency view either within or outside the window, while keeping all other settings unchanged, and compare the resulting changes in extreme fidelity. As shown in Figure~\ref{fig:cross_view_window_effect}, the changes within the window are substantially larger than those outside the window across all seven datasets, with differences of up to $38\times$. This confirms that the cross-view direction is substantially more effective within the cross-view window. Restricting optimization to this window also reduces the search space of the fusion parameters and simplifies subsequent optimization. 
Additional analyses and discussion are provided in Appendix~\ref{appendix:cross_view_window}.

\subsection{Lightweight Cross-View Optimization}\label{Cross_View_Optimization}

Finally, we determine how far to move along the cross-view direction by learning step-wise fusion parameters $\alpha_i$ within the cross-view window. Since fusion outside this window offers little benefit to extreme fidelity (Figure~\ref{fig:cross_view_window_effect}), we use the time view alone for the remaining steps by setting $\alpha_i=1$.We first independently train the time- and frequency-view denoisers, $D^T_{\theta_T}$ and $D^F_{\theta_F}$ from Eq.~\ref{eq:denoise_def}, with the EDM denoising objective and freeze their parameters before optimizing $\alpha_i$. Here, $X_\sigma^v$ denotes the corresponding 2D view representation at noise level $\sigma$. The training objective is a noise-weighted MSE between the denoised estimate and the clean sample:
\begin{equation}
\mathcal{L}_{\mathrm{denoise}}^{v}
=
\mathbb{E}_{\mathbf{x}_0,\sigma,\boldsymbol{\epsilon}}
\left[
\lambda(\sigma)
\left\|
D_{\theta_v}^{v}
\left(
X_\sigma^v,\sigma
\right)
-\mathbf{x}_0
\right\|_2^2
\right],
\qquad v\in\{T,F\},
\label{eq:denoising_objective}
\end{equation}
where $\boldsymbol{\epsilon}\sim\mathcal{N}(\mathbf{0},\mathbf{I})$ and $\lambda(\sigma)$ is the noise-dependent weighting function in EDM. Further details on the training and sampling configurations of the EDM-based denoiser are in Appendix~\ref{Appendix:EDM}.

After training and freezing both denoisers, the only learnable variables are six scalar fusion parameters, $\boldsymbol{\alpha}=\{\alpha_i\}_{i=1}^{6}$, one for each denoising step in the six-step cross-view window. To optimize these parameters, we formulate a Peaks-Over-Threshold (POT)-based objective grounded in EVT~\citep{coles2001introduction}. For each extreme dimension $q\in\{\mathrm{mag},\mathrm{chg}\}$, POT approximates the distribution of excesses above a sufficiently high threshold with a Generalized Pareto Distribution (GPD), whose shape parameter $\xi^q$ characterizes tail behavior across pooled exceedances. Rather than relying on this pooled tail descriptor alone, we further characterize extremes at the sequence level through their occurrence frequency and severity. Specifically, the per-sequence exceedance count $C^q$ is modeled by a Negative Binomial distribution~\citep{hilbe2011negative}, while the mean excess $M^q$ is modeled by a Gamma distribution~\citep{johnson1994continuous}. The resulting EVT-based objective is formulated as:
\begin{equation}
\mathcal{L}_{\mathrm{EVT}}(\boldsymbol{\alpha})
=
\sum\nolimits_{q\in\{\mathrm{mag},\mathrm{chg}\}}
\Big[
|\xi_g^q-\xi_r^q|
+
\widetilde{W}_1(\mathrm{NegBin}_g^q,\mathrm{NegBin}_r^q)
+
\widetilde{W}_1(\Gamma_g^q,\Gamma_r^q)
\Big],
\label{eq:evt_objective}
\end{equation}
where subscripts $g$ and $r$ denote generated and real data, respectively, $\mathrm{NegBin}^q$ and $\Gamma^q$ denote the corresponding Negative Binomial and Gamma distributions, and $\widetilde{W}_1$ denotes the Wasserstein distance normalized by the mean of the corresponding real data. The three terms quantify discrepancies in \textbf{\textit{tail shape}}, \textit{\textbf{extreme occurrence frequency}}, and \textbf{\textit{extreme severity}}, respectively. All three terms are dimensionless and are combined with equal weights.

Because CrossTF allows extrapolation beyond the two view-specific estimates with $\alpha_i\notin[0,1]$, we further constrain the energy distance $\mathcal{E}$ between the generated and real distributions to limit the potential degradation in overall generation quality. The resulting optimization problem is:
\begin{equation}
\boldsymbol{\alpha}^{*}
=
\arg\min_{\boldsymbol{\alpha}}
\mathcal{L}_{\mathrm{EVT}}(\boldsymbol{\alpha})
\quad
\mathrm{s.t.}\quad
\mathcal{E}(\boldsymbol{\alpha})
\leq
\mathcal{E}_T
+
\kappa\,\mathrm{SE}(\mathcal{E}_T).
\label{eq:cross_view_optimization}
\end{equation}
Here, $\mathcal{E}_T$ denotes the energy distance of the pure time-view baseline, and $\mathrm{SE}(\mathcal{E}_T)$ denotes its estimated standard error accounting for randomness in training and generation. We set $\kappa=2$, allowing the optimized energy distance to exceed the baseline by at most two standard errors, thereby limiting the overall performance degradation introduced by cross-view extrapolation to the baseline's natural statistical fluctuation range. We solve the resulting constrained nonlinear optimization problem using Sequential Least Squares Programming (SLSQP)~\citep{kraft1988software}, which iteratively solves local quadratic subproblems with finite-difference gradient estimates. With both denoisers frozen, this stage reduces to a lightweight optimization problem involving only six scalar fusion parameters.

\section{Experiments} \label{sec:experiment}
\textbf{Research Questions (RQs).} 
Our evaluation addresses seven key research questions.
\textbf{RQ1:} Does CrossTF improve extreme fidelity?
\textbf{RQ2:} Does CrossTF preserve overall generation fidelity?
\textbf{RQ3:} Does CrossTF generate useful synthetic data for downstream tasks?
\textbf{RQ4:} How does cross-view fusion compare with generation using either view alone?
\textbf{RQ5:} Do the learned step-wise fusion parameters $\alpha_i$ exhibit both interpolation and extrapolation?
\textbf{RQ6:} Does cross-view complementarity provide benefits beyond model diversity?
\textbf{RQ7:} How computationally efficient is CrossTF?


\textbf{Datasets.} We evaluate CrossTF on \textbf{seven real-world datasets} across diverse domains:
\href{https://github.com/liyaguang/DCRNN}{\textsc{Traffic}},
\href{https://docs.deweydata.io/docs/custom-weather-weather-data}{\textsc{Precip.}},
\href{https://github.com/buds-lab/building-data-genome-project-2}{\textsc{Build.}},
\href{https://docs.deweydata.io/docs/custom-weather-weather-data}{\textsc{Temp.}},
\href{https://archive.ics.uci.edu/dataset/235/individual+household+electric+power+consumption}{\textsc{Elect.}},
\href{https://archive.ics.uci.edu/ml/datasets/pems-sf}{\textsc{Transp.}}, and
\href{https://www.physionet.org/physiobank/database/ltstdb/}{\textsc{ECG}}.
Here, \textsc{Traffic}, \textsc{Precip.}, \textsc{Build.}, \textsc{Temp.}, \textsc{Elect.}, \textsc{Transp.}, and \textsc{ECG}
denote Traffic Speed, Precipitation, Building Power, Temperature, Electricity, Traffic Occupancy, and Electrocardiogram, respectively.
These datasets capture diverse real-world extremes, including abrupt traffic-speed changes, intense precipitation, building-load surges, unusual temperature deviations, power surges, high freeway occupancy, and pronounced ECG deviations. Detailed descriptions and sources are provided in Appendix~\ref{appendix:datasets}.

\textbf{Metrics.} We evaluate CrossTF with \textbf{ten metrics} for extreme fidelity and overall performance. Extreme fidelity uses four metrics over extreme magnitude and change: \textsc{Mag-Rate} and \textsc{Chg-Rate} measure occurrence-frequency discrepancies, while \textsc{Mag-Excess} and \textsc{Chg-Excess} measure exceedance severity. Overall performance uses three fidelity metrics (\textsc{Wass.}~\citep{villani2009optimal}, \textsc{Marg.}~\citep{naiman2024utilizing}, and \textsc{ACD}~\citep{ang2023tsgbench}) and three downstream metrics (\textsc{Context-FID}~\citep{yuan2024diffusion}, \textsc{Pred}~\citep{naiman2024utilizing}, and \textsc{Disc}~\citep{naiman2024utilizing}) for representation, prediction, and discrimination, respectively. \textsc{Pred} follows the Train-on-Synthetic, Test-on-Real protocol. All metrics are lower-is-better. Details are in Appendix~\ref{appendix:metrics}.

\textbf{Baselines.} We compare CrossTF with \textbf{11 representative baselines} spanning GAN, VAE, LLM, diffusion, time-frequency-based, and extreme-aware generation approaches, including TimeGAN~\citep{yoon2019time}, KoVAE~\citep{naimangenerative}, SDForger~\citep{rousseau2025forging}, DiffTime~\citep{coletta2023constrained}, Diffusion-TS~\citep{yuan2024diffusion}, FourierDiff~\citep{crabbe2024time}, ImagenTime~\citep{naiman2024utilizing}, RF-Diffusion~\citep{chi2024rf}, FIDE~\citep{galib2024fide}, HeavyDiff~\citep{pandey2025heavy}, and E4GEN~\citep{jiang2026e4gen}. Details are in Appendix~\ref{appendix:baselines}.

\subsection{Overall Performance}
\begin{table*}[t]
\centering
\caption{
Summary of comparisons on seven datasets.
\textit{Best} denotes the strongest baseline.
\textit{W--L} counts Wins-Losses against 11 baselines across 10 metrics, yielding 110 pairwise comparisons per dataset.
Lower is better.
}
\label{tab:summary_results}

\fontsize{8}{9}\selectfont
\setlength{\tabcolsep}{2.5pt}
\renewcommand{\arraystretch}{1.15}

\begin{tabular*}{\textwidth}{
@{\extracolsep{\fill}}
lccccccccc
@{}
}
\toprule

\multirow{2}{*}{\textbf{Dataset}}
& \multicolumn{4}{c}{\textbf{Extreme Fidelity}}
& \multicolumn{4}{c}{\textbf{Overall Performance}}
& \multirow{2}{*}{\textbf{W--L}} \\

\cmidrule(lr){2-5}
\cmidrule(lr){6-9}

& \textbf{Best}
& \textbf{Ours}
& \textbf{Baseline}
& \textbf{Improve}
& \textbf{Best}
& \textbf{Ours}
& \textbf{Baseline}
& \textbf{Improve}
& \\

\midrule

\textsc{Traffic}
& ImagenTime
& \textbf{0.0343}
& 0.0591
& \textbf{+42.0\%}
& FourierDiff
& \textbf{0.0083}
& 0.0138
& \textbf{+39.9\%}
& 110--0 \\

\textsc{Precip.}
& ImagenTime
& \textbf{0.0755}
& 0.1086
& \textbf{+30.5\%}
& ImagenTime
& \textbf{0.0050}
& 0.0051
& \textbf{+2.0\%}
& 103--7 \\

\textsc{Build.}
& ImagenTime
& \textbf{0.0762}
& 0.1012
& \textbf{+24.7\%}
& ImagenTime
& \textbf{0.0040}
& 0.0046
& \textbf{+13.0\%}
& 102--8 \\

\textsc{Temp.}
& ImagenTime
& \textbf{0.0884}
& 0.1222
& \textbf{+27.7\%}
& FIDE
& \textbf{0.0131}
& 0.0165
& \textbf{+20.6\%}
& 101--9 \\

\textsc{Elect.}
& ImagenTime
& \textbf{0.0691}
& 0.0734
& \textbf{+5.9\%}
& ImagenTime
& 0.0220
& \textbf{0.0193}
& $-14.0\%$
& 102--8 \\

\textsc{Transp.}
& ImagenTime
& \textbf{0.0477}
& 0.0691
& \textbf{+31.0\%}
& ImagenTime
& 0.0120
& \textbf{0.0114}
& $-5.3\%$
& 107--3 \\

\textsc{ECG}
& ImagenTime
& \textbf{0.1421}
& 0.1807
& \textbf{+21.4\%}
& ImagenTime
& 0.0201
& \textbf{0.0193}
& $-4.1\%$
& 105--5 \\

\midrule

\textbf{Summary}
& --
& --
& --
& \textbf{+26.1\%}
& --
& --
& --
& \textbf{+7.4\%}
& \textbf{730--40} \\

\bottomrule
\end{tabular*}
\end{table*}
\begin{table*}[t]
\centering
\caption{Generation performance on the \textsc{Traffic} dataset.
All metrics are lower-is-better.}
\label{tab:traffic_results}

\fontsize{8}{9}\selectfont
\setlength{\tabcolsep}{2.5pt}
\renewcommand{\arraystretch}{1.05}

\begin{tabular*}{\textwidth}{
@{\extracolsep{\fill}}
cclccccccc
@{}
}
\toprule

\multirow{2}{*}{\textbf{Dataset}}
& \multirow{2}{*}{\textbf{Type}}
& \multirow{2}{*}{\textbf{Method}}
& \multicolumn{4}{c}{\textbf{Extreme Fidelity}}
& \multicolumn{3}{c}{\textbf{Overall Fidelity}} \\

\cmidrule(lr){4-7}
\cmidrule(lr){8-10}

&
&
& \textbf{Mag-Rate}
& \textbf{Chg-Rate}
& \textbf{Mag-Excess}
& \textbf{Chg-Excess}
& \textbf{Wass.}
& \textbf{Marg.}
& \textbf{ACD} \\

\midrule

\multirow{12}{*}{\textsc{Traffic}}
& \textbf{GAN}
& TimeGAN
& 0.6852 & 0.6682 & 1.2587 & 0.7178 & 0.1240 & 0.0027 & 0.1393 \\

\cmidrule(lr){2-10}

& \textbf{VAE}
& KoVAE
& 0.6755 & 0.9931 & 0.6919 & 1.0784 & 0.3751 & 0.0100 & 0.1384 \\

\cmidrule(lr){2-10}

& \textbf{LLM}
& SDForger
& 0.1641 & 0.7892 & 0.2229 & 0.9543 & 0.2521 & 0.0053 & 0.0690 \\

\cmidrule(lr){2-10}

& \multirow{4}{*}{\textbf{Diffusion}}
& DiffTime
& 0.2229 & 0.2625 & 0.2053 & 0.2194 & 0.1098 & 0.0020 & 0.0258 \\

&
& Diffusion-TS
& 0.2756 & 0.6143 & 0.2531 & 0.3993 & 0.1817 & 0.0036 & 0.0388 \\

&
& FourierDiff
& 0.0749 & 0.1563 & 0.1359 & 0.6281 & \underline{0.0373} & \underline{0.0017} & 0.0064 \\

&
& ImagenTime
& \underline{0.0634} & \underline{0.0532} & 0.0681 & \underline{0.0532} & 0.0471 & 0.0023 & \underline{0.0059} \\

\cmidrule(lr){2-10}

& \textbf{\makecell{TF-Based}}
& RF-Diffusion
& 0.4571 & 7.2173 & 0.5824 & 1.3235 & 0.2573 & 0.0082 & 0.2225 \\

\cmidrule(lr){2-10}

& \multirow{3}{*}{\textbf{\makecell{Extreme-\\Aware}}}
& FIDE
& 0.5573 & 5.3031 & 0.9103 & 2.2618 & 0.3703 & 0.0072 & 0.1041 \\

&
& HeavyDiff
& 0.3541 & 0.3541 & 0.4583 & 0.9180 & 0.1088 & 0.0022 & 0.0273 \\

&
& E4GEN
& 0.0901 & 0.5107 & \underline{0.0644} & 0.1965 & 0.1133 & 0.0048 & 0.0725 \\

\midrule

& \textbf{Ours}
& \textbf{CrossTF}
& \textbf{0.0323}
& \textbf{0.0416}
& \textbf{0.0282}
& \textbf{0.0367}
& \textbf{0.0208}
& \textbf{0.0014}
& \textbf{0.0041} \\

\bottomrule
\end{tabular*}
\vspace{-10pt}
\end{table*}


To answer RQs~1--3, Table~\ref{tab:summary_results} summarizes comparisons between CrossTF and 11 baselines across seven real-world datasets and ten metrics, yielding 770 pairwise comparisons. CrossTF achieves 730 wins, outperforming the baselines in 94.8\% of these comparisons. Relative to the strongest baseline on each dataset, CrossTF improves extreme fidelity by 26.1\% overall, with gains across all seven datasets, and overall performance by 7.4\%. 
Among the seven datasets, \textsc{Precipitation} and \textsc{Electricity}, which exhibit the weakest Ext./Overall ratios (0.47$\times$ and 1.09$\times$) in Table~\ref{tab:asymmetry}, also show some of the clearest degradations in individual overall-fidelity metrics. In contrast, \textsc{Traffic} exhibits the strongest asymmetry (34.44$\times$) and achieves the largest overall-performance improvement (+39.9\%). This correspondence supports our interpretation that stronger asymmetry provides greater room to improve extreme fidelity without degrading overall performance, whereas weaker asymmetry may require a greater trade-off between extreme and overall performance. We first use \textsc{Traffic} as a representative case to examine extreme and overall fidelity in detail, followed by a cross-dataset analysis. Detailed fidelity results for the remaining datasets are provided in Appendix~\ref{appendix:more_fidelity}, while downstream utility is reported separately in Table~\ref{tab:downstream_utility} in Appendix~\ref{appendix:downstream}.

As shown in Table~\ref{tab:traffic_results}, CrossTF achieves the best results on \textsc{Traffic} across all four extreme-fidelity and three overall-fidelity metrics. Compared with the strongest baseline for each metric, CrossTF reduces the four extreme-fidelity metrics by 21.8\%--56.2\% and the three overall-fidelity metrics by 17.6\%--44.2\%. Among the baselines, ImagenTime and E4GEN are competitive in extreme fidelity, while FourierDiff and ImagenTime perform strongly in overall fidelity. CrossTF improves upon these methods across both dimensions. Across all seven datasets, CrossTF shows its most consistent advantage on \textsc{Chg-Excess}, ranking first on every dataset with an average relative improvement of 32.6\% over the strongest baseline and at least a 22.1\% gain. In contrast, its gains on \textsc{Chg-Rate} are more modest, with an average relative improvement of 1.1\% and the best result on five of seven datasets. This suggests that cross-view optimization is more effective at controlling how far abrupt changes exceed the threshold than how often they occur.

For downstream utility, we utilize commonly adopted public evaluators in time-series generation studies to compute \textsc{Disc}, \textsc{Pred}, and \textsc{Context-FID}. 
As shown in Table~\ref{tab:downstream_utility}, CrossTF achieves the best or tied-best result in 18 of 21 dataset--task combinations, including all seven \textsc{Context-FID}, six \textsc{Pred}, and five \textsc{Disc} cases. These results demonstrate strong downstream utility of CrossTF-generated data across representation learning, prediction, and real-vs-synthetic discrimination tasks.

\subsection{Further Analysis} 
To answer RQ4, we compare CrossTF with \textsc{Time-Only} and \textsc{Frequency-Only} to examine the effect of cross-view fusion. Compared with the stronger single-view variant, \textsc{Time-Only}, CrossTF improves extreme fidelity by \textbf{18.0}\% on average, while overall fidelity changes by only \textbf{0.1}\%. This shows that cross-view fusion substantially improves extreme generation while largely maintaining overall distributional fidelity. Details are shown in Appendix~\ref{appendix:ablation}.

To answer RQ5, we analyze the learned step-wise fusion parameters $\alpha_i$ within the cross-view window across seven datasets. Both interpolation and extrapolation appear frequently, with $\alpha_i$ generally biased toward the time view and \textbf{52.4\%} of the parameters satisfying $\alpha_i>1$, corresponding to extrapolation beyond the time view. This shows that cross-view fusion is not a simple convex combination and that allowing extrapolation can further exploit the cross-view discrepancy to improve extremes. Details are shown in Appendix~\ref{appendix:fusion_parameters}.

To answer RQ6, we compare time--frequency cross-view fusion with time--time fusion using two independently trained time-view denoisers, thereby controlling for the effect of same-view model diversity. The cross-view direction exhibits an extreme-fidelity variation \textbf{11.25$\times$} larger than the same-view direction, while all same-view variations remain within sampling noise. These results show that the \emph{Extreme-Sensitive Cross-View Asymmetry} is a characteristic of the cross-view complementarity between the time and frequency views, rather than a generic effect of model ensemble or model diversity. Details are shown in Appendix~\ref{appendix:model_diversity}.

To answer RQ7, we compare the training and sampling efficiency of CrossTF and all baselines on a single NVIDIA RTX PRO 6000 Blackwell Server GPU. CrossTF requires only \textbf{3.7 seconds} to generate 1,000 samples, whereas the more competitive ImagenTime, FourierDiff, and E4GEN require approximately 6.03$\times$, 6.27$\times$, and 20.14$\times$ longer sampling time, respectively. These results show that CrossTF maintains fast sampling and practical computational efficiency while achieving strong generation performance. Details are shown in Appendix~\ref{appendix:computation}.

\section{Related Work}\label{sec:related_work}
Recent studies have explored diverse approaches to time-series generation, including GANs, VAEs, and diffusion models, with a primary focus on reproducing overall data distributions.
\textbf{Extreme-aware generation} instead focuses on rare events. FIDE emphasizes high-frequency components, E4GEN uses event-level shapelets~\citep{jiang2026e4gen}, and Tail-GAN focuses on tail statistics~\citep{cont2026tail}.
In contrast, CrossTF freezes denoisers trained with the standard denoising objective and optimizes only their fusion to improve extreme fidelity under an overall-fidelity constraint.
\textbf{Time-frequency modeling} incorporates spectral structure into time-series generation: FourierDiff denoises spectral coefficients~\citep{crabbe2024time}, while RF-Diffusion couples time and frequency domains through a prescribed process for radio signals~\citep{chi2024rf}.
CrossTF instead produces complete time- and frequency-view clean-sample estimates from the same intermediate sample at each denoising step and learns how far to move along the direction defined by their discrepancy to improve extreme fidelity.
Further discussion is provided in Appendix~\ref{related_work_details}.

\section{Conclusion}\label{sec:conclusion}
In this paper, we presented CrossTF, a diffusion framework that exploits time--frequency cross-view discrepancy for extreme-aware time-series generation. Our empirical analysis revealed \emph{extreme-sensitive cross-view asymmetry}: movement along the cross-view direction substantially affects extreme fidelity while having comparatively limited effects on overall generation quality. We further identified a late-stage \textit{cross-view window} in which the discrepancy stabilizes and cross-view adjustments become particularly effective. Building on these findings, CrossTF decouples denoiser training from extreme-focused optimization by freezing both denoisers and learning only six fusion parameters. An EVT-based objective guides interpolation and extrapolation, while an energy-distance constraint limits degradation in overall generation quality. Experiments on seven real-world datasets against 11 baselines demonstrated improvements in extreme fidelity and overall performance. These findings establish time--frequency cross-view discrepancy as an effective optimization mechanism for improving extreme-event time-series generation without retraining the underlying denoisers.

\clearpage

\section*{AI Use Statement}
In this work, we used generative AI tools, including large language models, to assist with experimental code generation and editing, parts of mathematical derivations, manuscript editing for improved readability, literature retrieval, and reference formatting. The authors reviewed all AI-assisted work, including verifying AI-assisted code and mathematical derivations for correctness, checking that language edits preserved the intended scientific meaning, and confirming the relevance and bibliographic accuracy of suggested references. The authors take full responsibility for the final content of this work and its associated materials, including the text, mathematical arguments, references, implementation, experimental results, and scientific claims, regardless of whether generative AI assistance was involved.

\section*{Ethics statement}
All authors have read and agree to abide by the ICLR Code of Ethics. This study does not involve the collection of private or personally identifiable information. The work focuses on methodological development for extreme-aware time-series generation and does not involve direct interaction with human participants.

\section*{Reproducibility Statement}
To support full reproduction of our results, we release the complete implementation and experimental materials of CrossTF in an anonymous repository:
\href{https://github.com/LinJiang18/CrossTF}{\textcolor{blue}{\textbf{Anonymous Repository}}}.
The repository covers the complete experimental pipeline, from raw data and model configurations to training, sampling, evaluation, and table generation. Specifically, \texttt{data/} provides the raw data used for the seven datasets; \texttt{configs/} contains the complete CrossTF configurations for each dataset; \texttt{baselines/} provides the implementations and corresponding configurations of all 11 baseline methods; \texttt{results/} contains the generated samples from CrossTF and all 11 baselines; and \texttt{evaluation/} provides the implementations of the evaluation metrics and result aggregation code. We further provide \texttt{evaluation/make\_table.py}, which can directly reproduce the main experimental results for all seven datasets reported in Tables~\ref{tab:traffic_results}, \ref{tab:precip_results}, \ref{tab:build_results}, \ref{tab:temp_results}, \ref{tab:elect_results}, \ref{tab:transp_results}, and~\ref{tab:ecg_results} from the released generated samples. With these materials, readers can either rerun the full training and sampling pipeline from scratch or directly reproduce the main quantitative results from the provided samples.

Our experiments were conducted on multiple high-performance computing clusters, primarily using NVIDIA RTX PRO 6000 Blackwell, NVIDIA B200, and NVIDIA A100 GPUs. To ensure fair comparison across methods, all methods follow the unified data splits, evaluation metrics, and experimental protocols reported in the paper. The main hyperparameters and implementation details required for model architecture, training, sampling, optimization, and evaluation are provided in the main text, appendices, and anonymous code repository.

\bibliography{iclr2027_conference}
\bibliographystyle{iclr2027_conference}

\clearpage
\newpage
\appendix
\newpage

\section{Related Work}\label{related_work_details}

\subsection{Time-series Generation}
Time-series generation aims to synthesize sequences that preserve the temporal dynamics and distributional properties of real data~\citep{nikitin2024tsgm}.
In the early years, GAN-based methods were widely applied to this task, combining recurrent generators and discriminators to capture temporal dependencies~\citep{esteban2017real}.
TimeGAN~\citep{yoon2019time} augments adversarial training with a supervised loss in a jointly learned embedding space. Subsequent methods introduce alternative distribution-matching objectives based on causal optimal transport~\citep{xu2020cot} or characteristic functions on path space~\citep{lou2023pcf}.
GT-GAN~\citep{jeon2022gt} extends the paradigm to irregularly sampled time-series through a continuous-time formulation.
Although effective, adversarial approaches can suffer from unstable training and incomplete mode coverage.

Likelihood-based approaches offer an alternative through explicit or variational objectives. TimeVAE~\citep{desai2021timevae} reconstructs sequences through interpretable components such as level, trend, and seasonality, while Koopman VAE~\citep{naimangenerative} introduces structured latent dynamics based on Koopman operators. Normalizing flows enable exact likelihood evaluation, with Fourier flows~\citep{alaa2021generative} modeling sequences through spectral representations. These approaches support likelihood-based learning, although preserving fine-grained local structure remains challenging, particularly for models with restrictive latent or reconstruction assumptions.

Diffusion models have become a prominent approach following the success of iterative denoising in waveform synthesis~\citep{kong2021diffwave}. Diffusion-TS~\citep{yuan2024diffusion} combines a decomposition-based denoiser with interpretable trend and seasonal components, while self-guided formulations reuse an unconditional model for generation and refinement~\citep{kollovieh2023predict}. Other methods change the representation in which denoising occurs, either transforming sequences into two-dimensional representations~\citep{naiman2024utilizing} or modeling spectral coefficients directly~\citep{crabbe2024time}. Iterative refinement supports strong generation fidelity but typically requires multiple denoiser evaluations during sampling. Beyond these paradigms, time-series generation has also been explored through contrastive imitation~\citep{jarrett2021time} and language modeling over tokenized sequences~\citep{rousseau2025forging,das2024decoder}.

\subsection{Extreme-Aware Time-Series Generation}\label{sec:related_extreme}
Existing extreme-aware generation methods can be broadly grouped into representation-based and objective-based approaches.

\textbf{Representation-based methods.} These methods incorporate extreme-aware structures into the data representation or generative process. FIDE~\citep{galib2024fide} amplifies high-frequency components to retain spectral patterns associated with extremes during denoising, while E4GEN~\citep{jiang2026e4gen} uses time-domain shapelets to represent extremes as event-level patterns that can be individually interpreted and enhanced. Heavy-tailed diffusion~\citep{pandey2025heavy} instead introduces extreme-aware assumptions into the generative process by replacing the Gaussian prior and perturbation kernel with their Student-$t$ counterparts to better accommodate heavy-tailed data. These design choices introduce inductive biases that may not fully capture the diversity of extreme patterns across datasets.

\textbf{Objective-based methods.} These methods emphasize extreme behavior through tailored learning objectives. ExGAN~\citep{bhatia2021exgan} conditions adversarial generation on a target extremeness level, enabling sampling at a prescribed severity, while Tail-GAN~\citep{cont2026tail} uses a risk-aware objective based on tail statistics to improve the reproduction of tail risk. Prioritizing rare extremes, however, may introduce a trade-off with fidelity to the broader data distribution.

CrossTF differs from both categories in the variables it optimizes and the stage at which extreme-focused optimization occurs. It first trains the view-specific denoisers using the standard denoising objective and then freezes them, without introducing a predefined extreme-specific representation or generative assumption. Extreme fidelity is subsequently refined through a small set of fusion parameters that control movement along the cross-view direction, to which the denoising loss is comparatively insensitive. This avoids committing to a particular form of extremes and allows the optimization to adapt to diverse extreme patterns across datasets, while an explicit distributional constraint limits degradation in overall generation quality.

\subsection{Time-Frequency Modeling for Time-Series Generation}\label{sec:related_timefreq}

The time and frequency domains provide mathematically equivalent descriptions of the same sequence~\citep{shatkay1995fourier}, but organize its structure differently for generative modeling. Existing approaches broadly fall into two categories: representation-centric and integration-centric methods.

\textbf{Representation-centric methods.} These methods choose a spectral or time--frequency representation as the primary generative space. Frequency-domain diffusion, for example, directly denoises Fourier coefficients~\citep{crabbe2024time}, whereas WaveletDiff operates on multilevel wavelet coefficients and exchanges information across resolutions~\citep{wang2026waveletdiff}. Such representations expose compact spectral or multiscale structure, but the roles of time and frequency information are determined by the chosen transform rather than modeled as two parallel generative views.

\textbf{Integration-centric methods.} These methods combine the two domains within a model while explicitly specifying how they interact. Frequency-conditioned diffusion retains a time-domain denoising path and injects adaptively separated spectral components as conditions~\citep{jeong2025frequency}. Related generative models for imputation similarly fuse frequency-derived features into a temporal modeling pipeline~\citep{yang2024frequency}. More tightly coupled designs incorporate both domains directly into the diffusion process. RF-Diffusion develops a time--frequency formulation specialized for RF signals~\citep{chi2024rf}, while HyFAD performs sequential cross-domain denoising with predefined coarse-to-fine roles~\citep{gao2026hyfad}. Despite differences in coupling strength, these methods specify in advance how and where each domain contributes, such as through conditioning or stage-specific processing.

CrossTF instead treats the two domains as parallel generative views of the same intermediate sample. At each denoising step, the time and frequency views independently produce complete clean-sample estimates along the same sampling trajectory, without assigning either view a permanent primary or auxiliary role. Their discrepancy therefore defines a sample-dependent cross-view direction rather than a fixed backbone--auxiliary hierarchy. This design leverages the complementary properties characterized in Section~\ref{sec:temporal_frequency_views} without designating either domain as the primary pathway.

\section{How CrossTF Differs from Guidance and Multi-Model Combination Methods}
\label{appendix:guidance_relation}

CrossTF's cross-view fusion is algebraically similar to guidance-based sampling~\citep{dhariwal2021diffusion,ho2022classifier,karras2024guiding}. As defined in Equation~\ref{eq:cross_view_fusion}, CrossTF linearly combines two clean-sample estimates and allows extrapolation along their difference when $\alpha_i\notin[0,1]$. This naturally raises the question of whether CrossTF is simply another form of guidance or multi-model combination~\citep{balaji2022ediffi,liu2022compositional}. The key distinction lies not in the linear combination itself, but in how the two estimates are obtained, why their difference is useful, and how the fusion parameters are determined.

\textbf{Difference from guidance-based sampling.}
Classifier guidance steers the generation process using the gradient of an auxiliary classifier~\citep{dhariwal2021diffusion}, while classifier-free guidance uses the difference between conditional and unconditional predictions to strengthen conditioning~\citep{ho2022classifier}. Autoguidance further uses a deliberately smaller or less-trained model as a reference and guides sampling away from its errors~\citep{karras2024guiding}. CrossTF relies on neither a conditioning difference nor a deliberately weakened reference model. Its time- and frequency-view denoisers are independently trained on the same data using the same EDM denoising objective, noise distribution, and network configuration; they differ only in the representations of the same intermediate sample they receive. Therefore, the cross-view direction in CrossTF represents neither a conditional--unconditional difference nor a strong--weak model difference, but rather the discrepancy between two parallel representation views of the same sample.

Importantly, CrossTF does not assume that one view is uniformly superior to the other. On the Precipitation dataset, for example, the time view performs better on magnitude and change rates, whereas the frequency view performs better on magnitude and change excesses, indicating complementary strengths for different types of extreme behavior. Moreover, Table~\ref{tab:asymmetry} shows that, across seven datasets, varying the fusion parameter along the cross-view direction changes extreme fidelity substantially more than the denoising loss. CrossTF therefore exploits a direction that naturally arises from representation discrepancy and is particularly sensitive to extreme fidelity, rather than relying on a predefined guidance direction.

\textbf{Difference from post-training sampler optimization.}
CrossTF is also related to methods that optimize the sampling procedure after model training. Align Your Steps, for example, optimizes the discretization schedule of a fixed diffusion model~\citep{sabour2024align}. CrossTF likewise freezes the trained denoisers, but optimizes six step-wise fusion parameters, $\boldsymbol{\alpha}=\{\alpha_i\}_{i=1}^{6}$, within the cross-view window rather than the noise schedule. These parameters are optimized using the EVT-based extreme objective under an energy-distance constraint that limits degradation in overall distributional fidelity. Consequently, whether a step performs interpolation or extrapolation is determined by the constrained optimization rather than specified in advance as a guidance scale. Outside the cross-view window, CrossTF uses the time view alone.

\textbf{Difference from multi-model combination.}
CrossTF also differs from methods in which multiple denoisers are assigned different generative roles. For example, eDiff-I trains experts for different noise-level intervals and activates the corresponding expert at each stage~\citep{balaji2022ediffi}, while compositional diffusion combines models associated with different conditions to generate samples satisfying their conjunction~\citep{liu2022compositional}. In CrossTF, the two denoisers correspond neither to different noise stages nor to different semantic conditions. Both model the same unconditional data distribution and simultaneously produce complete clean-sample estimates of the same intermediate sample within the cross-view window. Their representation-specific discrepancy then defines the cross-view direction used to optimize extreme fidelity.

In summary, the similarity between CrossTF and guidance or multi-model methods is primarily algebraic: all involve combining multiple estimates. Their underlying mechanisms are different. In CrossTF, the two endpoints arise from parallel time and frequency views of the same data distribution, the cross-view direction is induced by representation discrepancy and empirically exhibits strong sensitivity to extreme fidelity, and the step-wise fusion parameters are learned through post-training optimization under an overall-fidelity constraint. CrossTF therefore does not rely on conditioning differences, deliberately weakened reference models, or predefined expert roles.

\section{Additional Preliminaries}

\subsection{Visualization of Representative Extremes in Time-series} \label{appendix:representative_extreme_event}
\begin{figure*}[htbp]
    \centering

    \begin{minipage}[t]{0.32\textwidth}
        \centering
        \includegraphics[width=\linewidth]{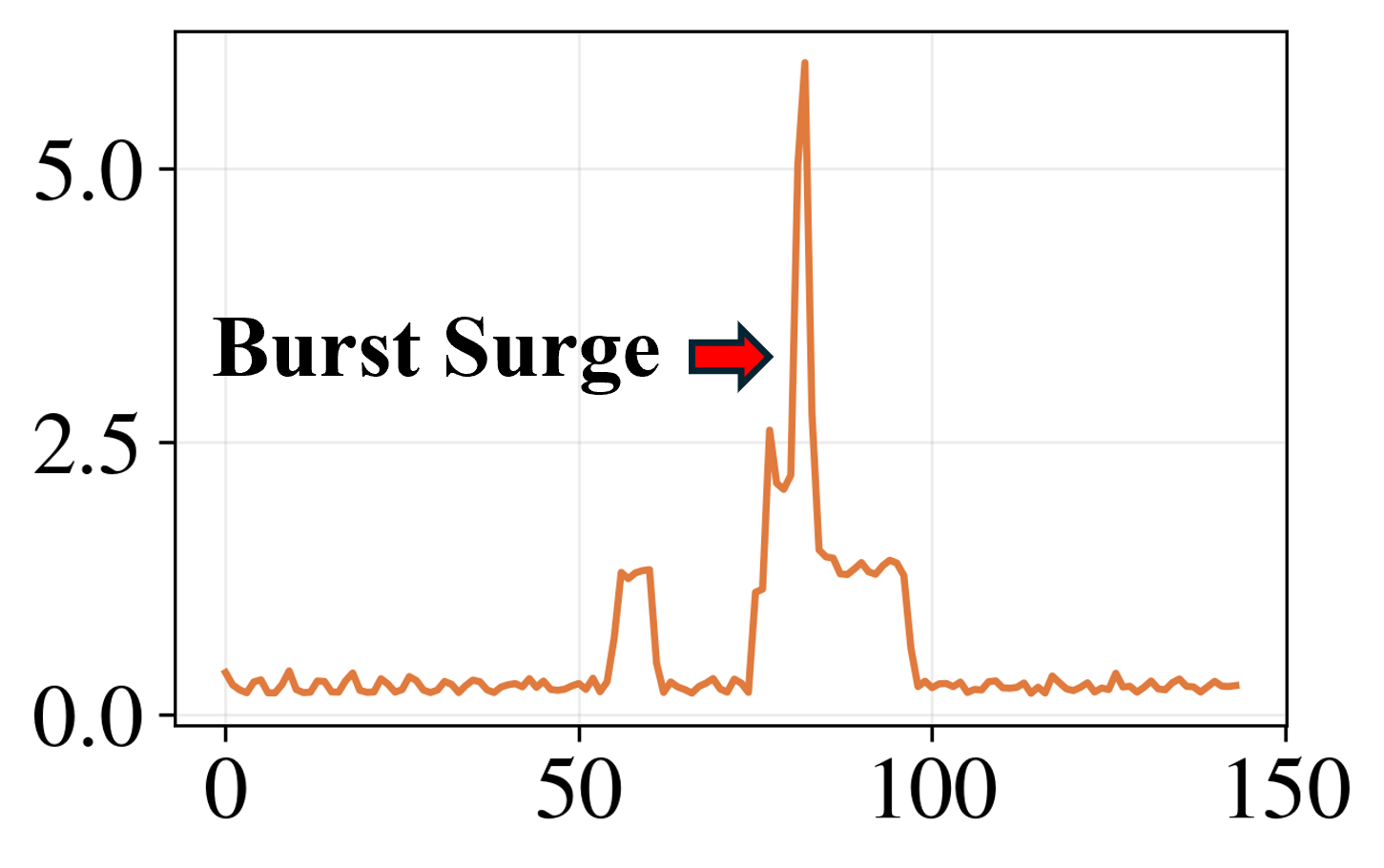}
        \vspace{-14pt}
        \caption*{(a) Electricity}
    \end{minipage}
    \hfill
    \begin{minipage}[t]{0.32\textwidth}
        \centering
        \includegraphics[width=\linewidth]{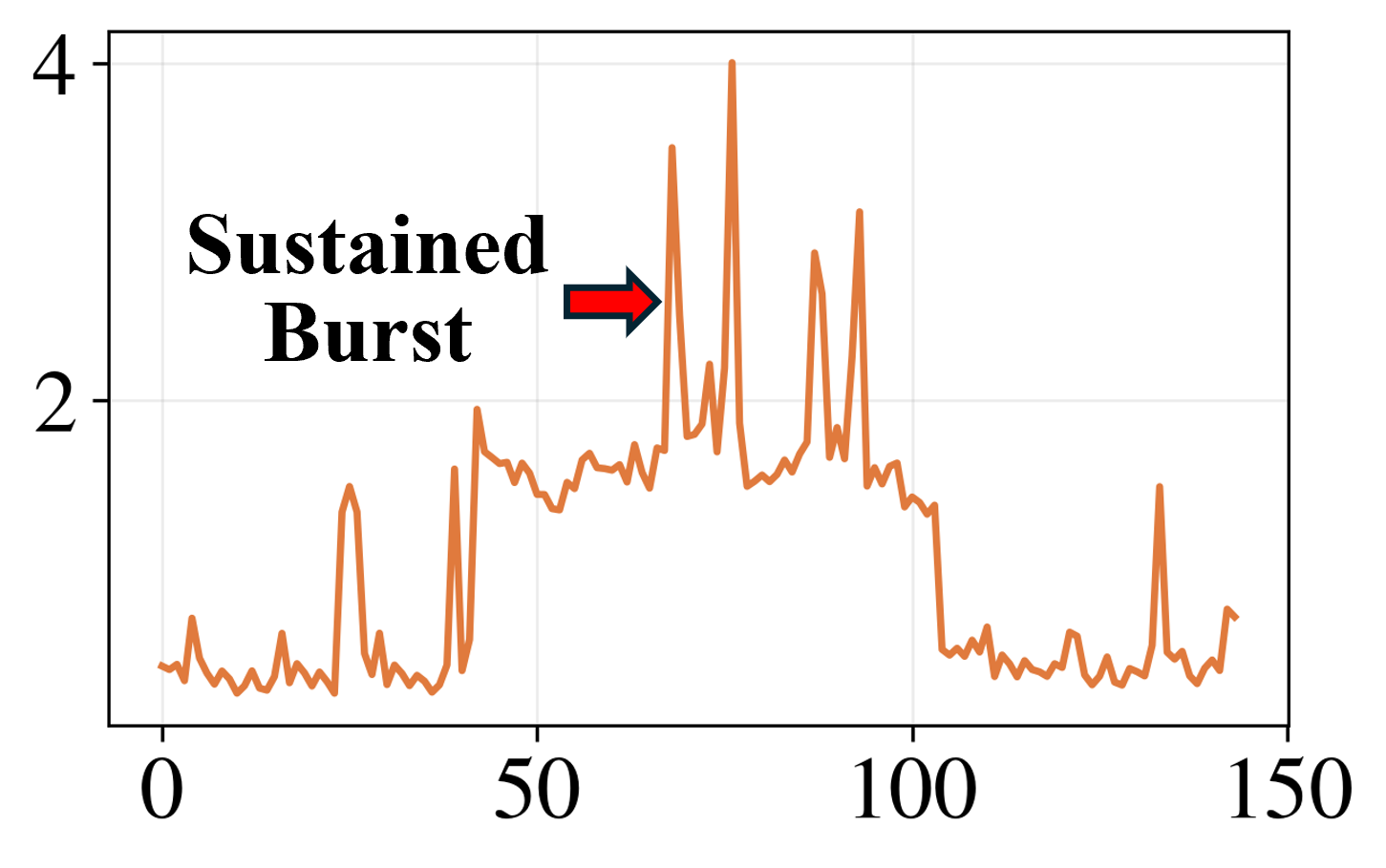}
        \vspace{-14pt}
        \caption*{(b) Traffic Flow}
    \end{minipage}
    \hfill
    \begin{minipage}[t]{0.30\textwidth}
        \centering
        \includegraphics[width=\linewidth]{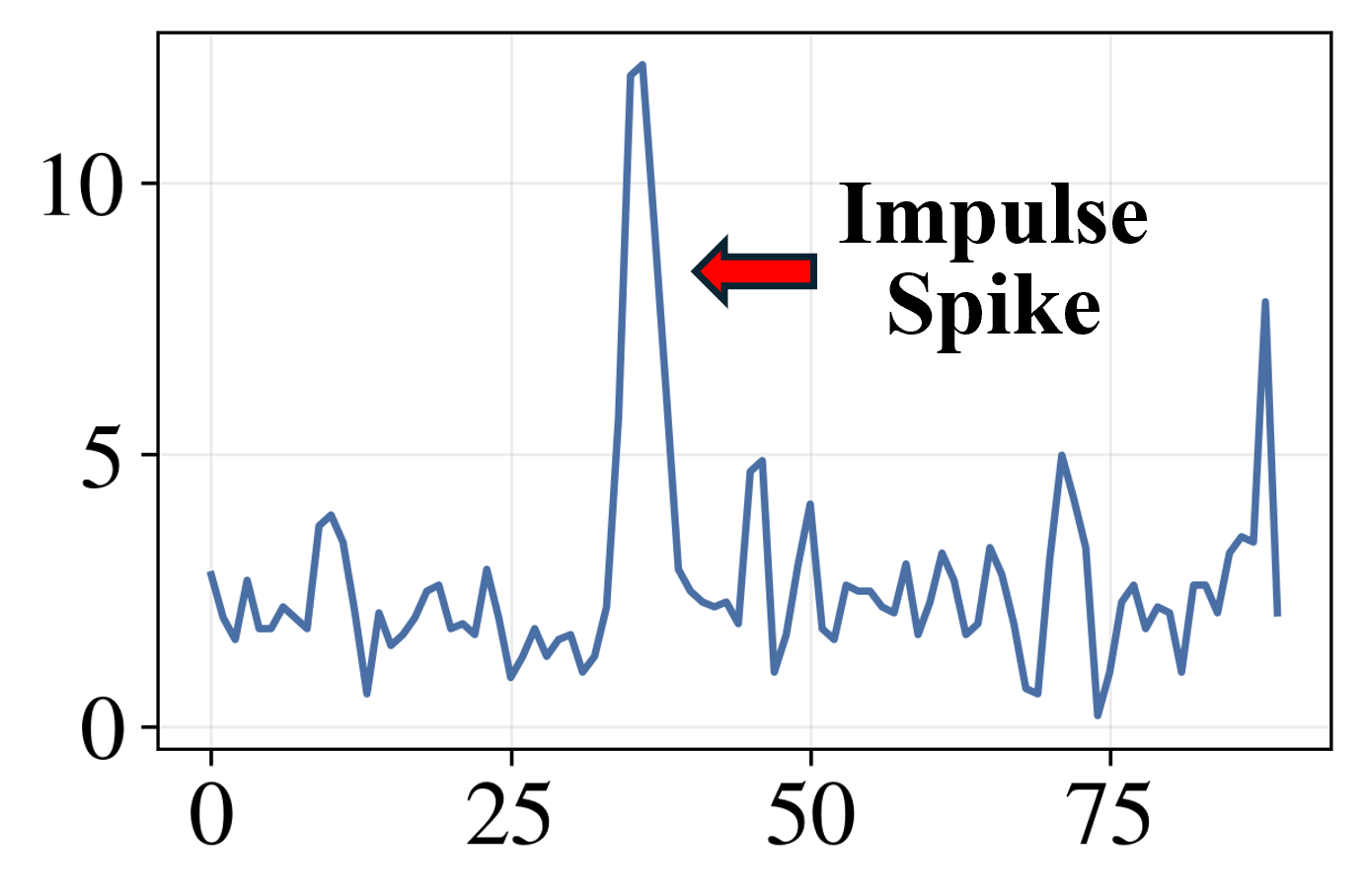}
        \vspace{-14pt}
        \caption*{(c) Wind Speed}
    \end{minipage}

    \vspace{0pt}

    \begin{minipage}[t]{0.30\textwidth}
        \centering
        \includegraphics[width=\linewidth]{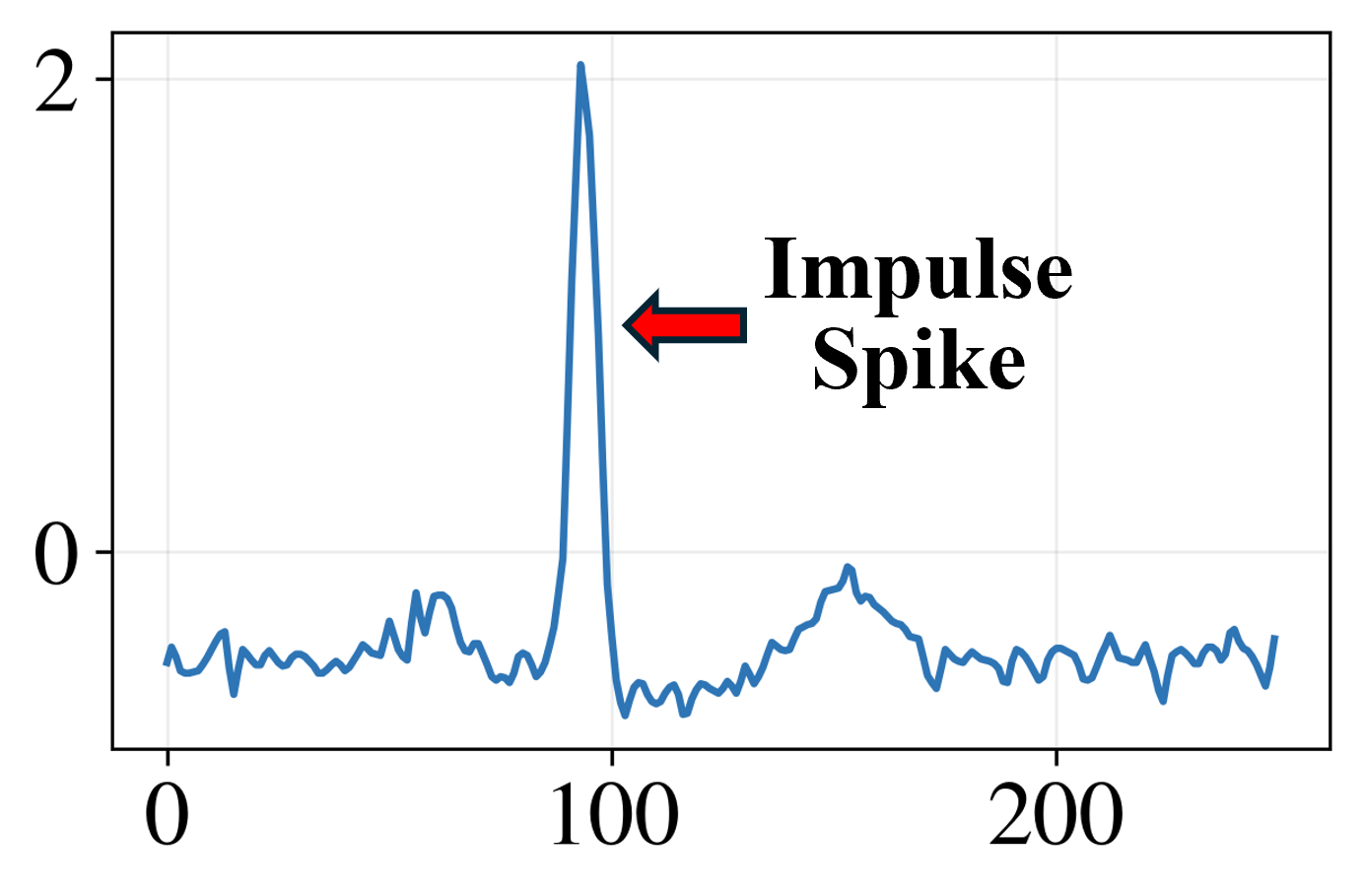}
        \vspace{-14pt}
        \caption*{(d) ECG}
    \end{minipage}
    \hfill
    \begin{minipage}[t]{0.32\textwidth}
        \centering
        \includegraphics[width=\linewidth]{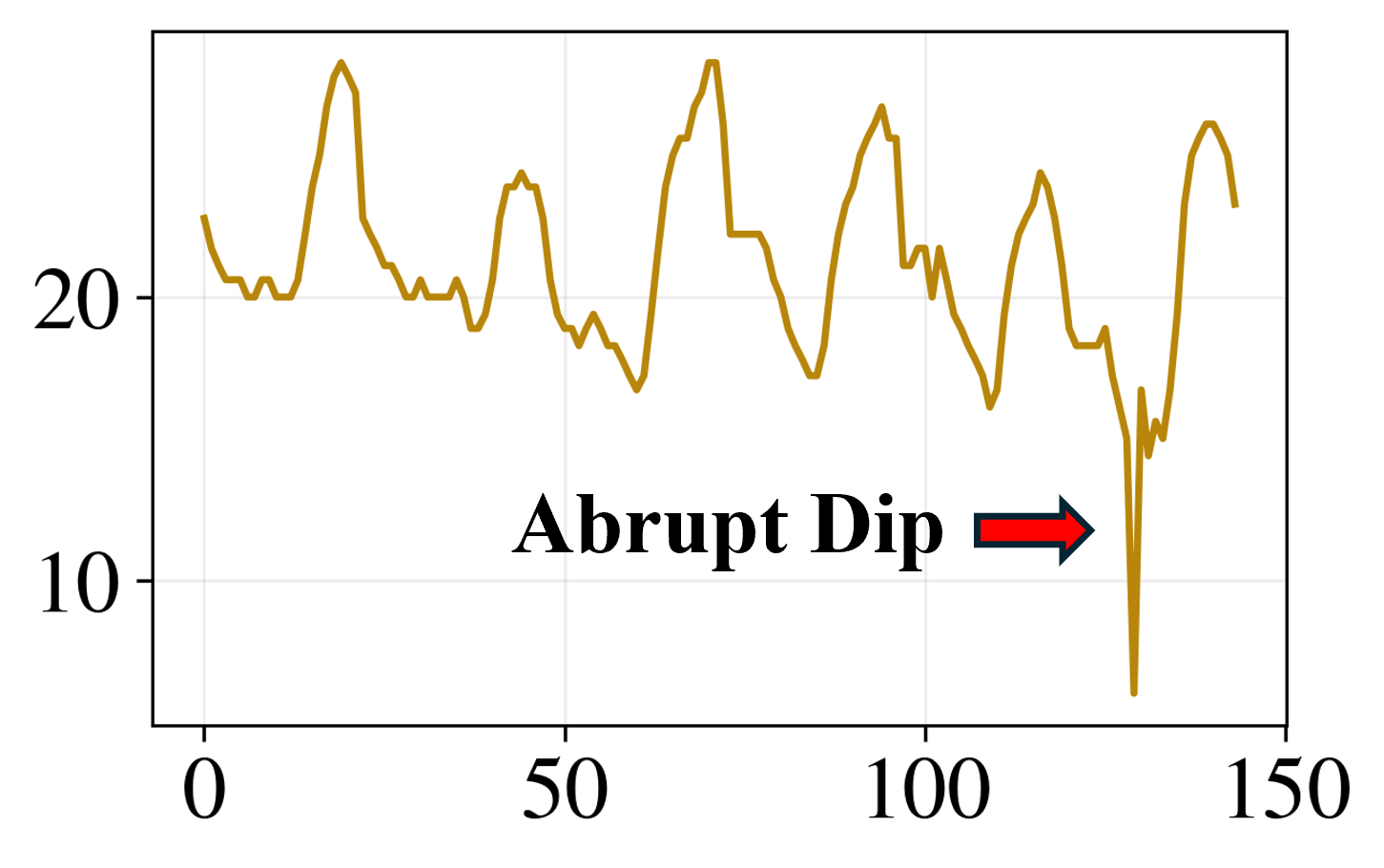}
        \vspace{-14pt}
        \caption*{(e) Temperature}
    \end{minipage}
    \hfill
    \begin{minipage}[t]{0.32\textwidth}
        \centering
        \includegraphics[width=\linewidth]{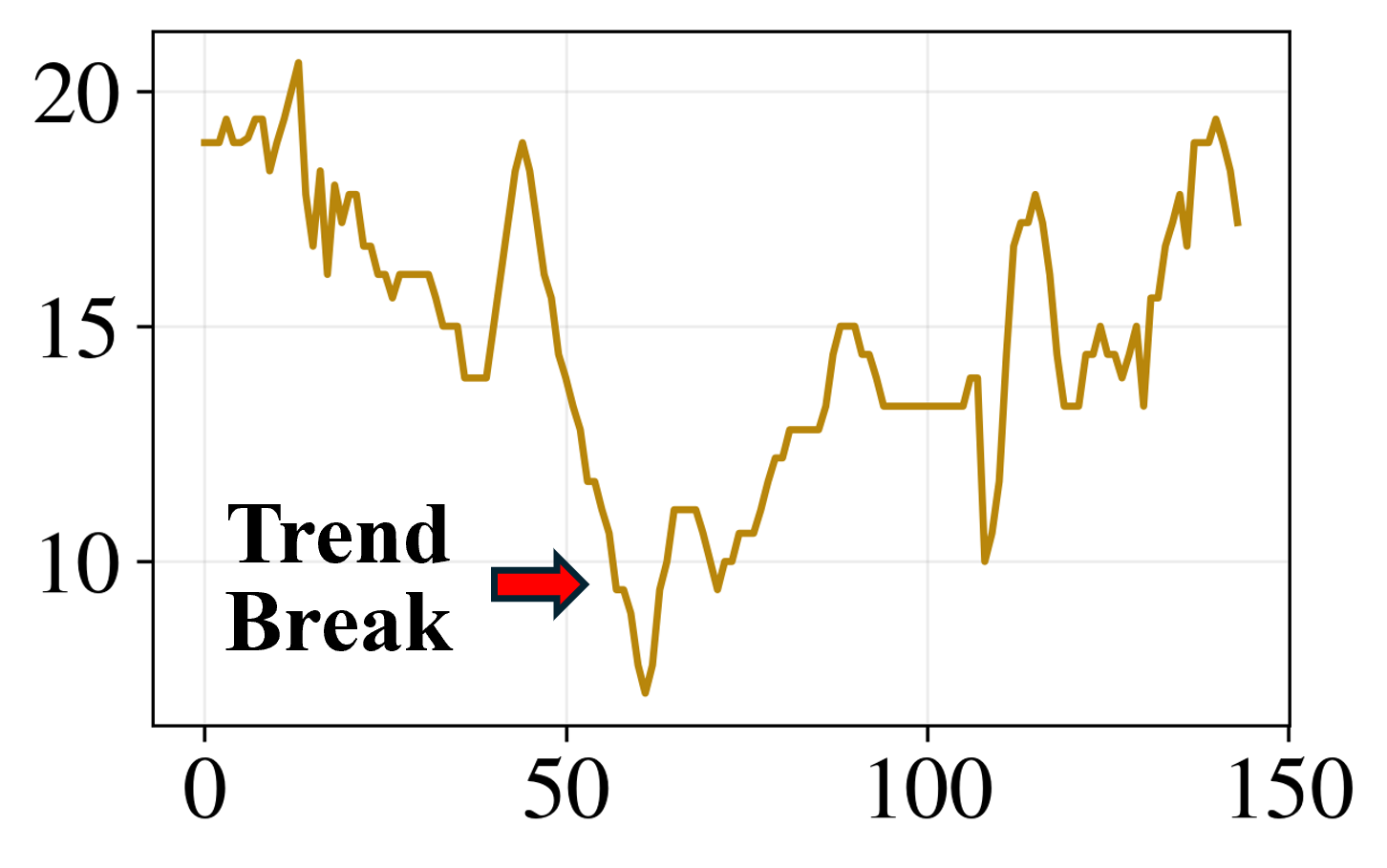}
        \vspace{-14pt}
        \caption*{(f) Temperature}
    \end{minipage}

    \vspace{-4pt}
    \caption{Representative extremes across different real-world time-series datasets.}
    \label{fig:representative_extreme_events}
\end{figure*}

We visualize representative extremes from five real-world datasets, covering diverse patterns such as burst surges, sustained bursts, impulse spikes, abrupt dips, and trend breaks. As shown in Figure~\ref{fig:representative_extreme_events}, these extremes exhibit substantially different shapes, durations, and local structures, making it difficult to describe them through a single event pattern. Despite this diversity, their extreme behavior can be largely reflected through two fundamental aspects: unusually large deviations from the typical level and abrupt changes between consecutive time steps. These examples further motivate our characterization of extremes through \emph{extreme magnitude} and \emph{extreme change}.

\subsection{Additional Analysis of Cross-View Asymmetry}\label{appendix:cross_view_asymmetry}

This section provides additional details on the experimental setup and statistical procedure used for the Extreme-Sensitive Cross-View Asymmetry analysis in Table~\ref{tab:asymmetry}.

For the asymmetry analysis, we follow the cross-view formulation in Section~\ref{sec:what}. At denoising step $i$, the fused clean-sample estimate is
\begin{equation}
\hat{\mathbf{x}}_{0,i}
=
\hat{\mathbf{x}}_{0,i}^F
+
\alpha_i
\left(
\hat{\mathbf{x}}_{0,i}^T
-
\hat{\mathbf{x}}_{0,i}^F
\right),
\end{equation}
where $\alpha_i=0$ and $\alpha_i=1$ correspond to the pure frequency-view and time-view estimates, respectively. In this analysis, we use a shared fusion parameter $\alpha_i=\alpha$ across all steps within the cross-view window and consider
$\alpha\in\{0,0.25,0.5,0.75,0.9,1.0\}$.

We use three independently trained time/frequency denoiser pairs with training seeds 4, 5, and 6, and evaluate their EMA weights~\citep{karras2022elucidating}. Generation uses two fixed sampling seeds, 101 and 102. For each training and sampling seed, we first execute the denoising steps preceding the cross-view window and cache the resulting intermediate state. All $\alpha$ values then continue sampling from the same cached state, ensuring that their differences arise only from cross-view fusion within the window rather than from different sampling trajectories.

For any quantity $q$, we define its Relative Max--Min Range as
\begin{equation}
\operatorname{Range}(q)
=
\frac{
\max_{\alpha}q(\alpha)-\min_{\alpha}q(\alpha)
}{
|q(\alpha=1)|
},
\end{equation}
where the pure time-view value is used for normalization. We first compute the range for each individual $\alpha$-response curve and then average the resulting ranges, rather than averaging the curves before computing the range. Specifically, \emph{Loss} is evaluated separately for each training seed and each noise level $\sigma$ within the cross-view window; \emph{Extreme} is evaluated separately for each of the four EVT-based metrics and each training seed; and \emph{Overall} is evaluated for each independent generation run. The metrics used for \emph{Extreme} and \emph{Overall} follow the definitions in the main evaluation.

The denoising loss is evaluated on real data. For each noise level $\sigma_i$ within the cross-view window, the time and frequency branches use the same noise realization and produce clean-sample estimates $\hat{\mathbf{x}}_{0,i}^T$ and $\hat{\mathbf{x}}_{0,i}^F$. We define their denoising errors as
\begin{equation}
e_T=\hat{\mathbf{x}}_{0,i}^T-\mathbf{x}_0,
\qquad
e_F=\hat{\mathbf{x}}_{0,i}^F-\mathbf{x}_0.
\end{equation}
Following the cross-view fusion in Section~\ref{sec:what}, the fused error is therefore
\begin{equation}
e_{\alpha}
=
e_F+\alpha(e_T-e_F),
\end{equation}
and the corresponding loss is
\begin{equation}
L(\alpha)
=
\left\|
e_F+\alpha(e_T-e_F)
\right\|^2.
\end{equation}
Since $L(\alpha)$ is quadratic in $\alpha$, we compute its coefficients once and evaluate it analytically over the entire $\alpha$ grid, eliminating the need for additional forward passes at each $\alpha$ value.

In Table~\ref{tab:asymmetry}, Ext./Loss and Ext./Overall denote, for each dataset,
the ratios of the Extreme range to the Loss and Overall ranges, respectively.
In the Median row, the first three entries report the median Loss, Overall, and
Extreme ranges across the seven datasets, and the last two entries divide the
median Extreme range by the median Loss and Overall ranges, respectively, using
the unrounded values.

To further assess whether the observed variation is attributable to $\alpha$ rather than ordinary seed-level variability, Table~\ref{tab:alpha_seed_variation} compares the $\alpha$-induced range with the variation across training seeds. For \emph{Extreme}, the variation induced by $\alpha$ is consistently larger than the seed variation across all seven datasets, often by a substantial margin, indicating that the observed sensitivity is not simply caused by model instability. For \emph{Overall}, the $\alpha$-induced variation is comparable to or smaller than the seed variation on Traffic, Temperature, ECG, and Building, while larger changes are observed on Transportation, Electricity, and Precipitation, suggesting that overall quality is relatively stable on some datasets but not uniformly invariant. For \emph{Loss}, the absolute variation remains modest on most datasets, with Precipitation showing the largest change. 
Taken together, these results reinforce that the cross-view direction has a substantially stronger and more consistent effect on extreme fidelity than would be expected from seed-level variability.

\begin{table}[t]
\centering
\caption{\textbf{Comparison between $\alpha$-induced variation and seed variation} across seven datasets.}
\label{tab:alpha_seed_variation}
\small
\renewcommand{\arraystretch}{0.95}
\begin{tabular}{lcccccc}
\toprule
& \multicolumn{2}{c}{Loss} 
& \multicolumn{2}{c}{Overall} 
& \multicolumn{2}{c}{Extreme} \\
\cmidrule(lr){2-3}
\cmidrule(lr){4-5}
\cmidrule(lr){6-7}
Dataset 
& $\alpha$ Var. & Seed Var.
& $\alpha$ Var. & Seed Var.
& $\alpha$ Var. & Seed Var. \\
\midrule
Traffic        & 4.5\%  & 1.45\% & 8.4\%   & 9.21\%  & 290.4\% & 17.10\% \\
Transportation & 7.5\%  & 1.70\% & 51.1\%  & 12.50\% & 254.2\% & 10.81\% \\
Temperature    & 7.3\%  & 1.70\% & 12.6\%  & 14.49\% & 145.6\% & 9.62\%  \\
Electricity    & 8.7\%  & 0.28\% & 78.4\%  & 3.76\%  & 85.2\%  & 4.50\%  \\
Precipitation  & 33.3\% & 1.01\% & 339.9\% & 4.63\%  & 160.3\% & 14.73\% \\
ECG            & 16.7\% & 1.20\% & 3.9\%   & 14.08\% & 38.2\%  & 9.15\%  \\
Building       & 12.9\% & 2.76\% & 10.8\%  & 9.87\%  & 22.8\%  & 6.10\%  \\
\bottomrule
\end{tabular}
\end{table}

\subsection{Details of the Cross-View Window}
\label{appendix:cross_view_window}

We provide the definition and estimation of Stable Discrepancy Share, followed by the experimental settings of the view-switching ablation.

\textbf{Stable Component:} For a fixed real sample $\mathbf{x}_0^{(n)}$, noise level $\sigma_i$, and trained denoiser pair, we construct time and frequency representations from the same noisy input. Let $\boldsymbol{\Delta}_i^{(n)}(\boldsymbol{\epsilon})$ denote the difference between their clean-sample estimates, where $n$ indexes real samples, $i$ denotes the denoising step, and $\boldsymbol{\epsilon}\sim\mathcal{N}(\mathbf{0},\mathbf{I})$ is Gaussian noise. We define the stable component as the expected discrepancy over noise realizations:
\begin{equation}
\boldsymbol{\mu}_i^{(n)}
=
\mathbb{E}_{\boldsymbol{\epsilon}}
\left[\boldsymbol{\Delta}_i^{(n)}(\boldsymbol{\epsilon})\right].
\end{equation}
Here, $\boldsymbol{\mu}_i^{(n)}$ captures the mean cross-view discrepancy for a fixed sample and noise level, independently of any particular noise realization.

\textbf{Deviation Component:} Each discrepancy can then be expressed as the sum of its stable component and a deviation component:
\begin{equation}
\boldsymbol{\xi}_i^{(n)}(\boldsymbol{\epsilon})
=
\boldsymbol{\Delta}_i^{(n)}(\boldsymbol{\epsilon})
-
\boldsymbol{\mu}_i^{(n)},
\qquad
\mathbb{E}_{\boldsymbol{\epsilon}}
\left[\boldsymbol{\xi}_i^{(n)}(\boldsymbol{\epsilon})\right]
=
\mathbf{0}.
\end{equation}
The residual $\boldsymbol{\xi}_i^{(n)}(\boldsymbol{\epsilon})$ captures fluctuations across noise realizations. Its zero expectation implies that the expected squared norm of the discrepancy decomposes into the energies of the stable and deviation components.

\textbf{Stable Discrepancy Share:} We therefore define the proportion of total discrepancy energy contributed by the stable component as
\begin{equation}
s_i
=
\frac{
\sum_{n=1}^{N}\left\|\boldsymbol{\mu}_i^{(n)}\right\|_2^2
}{
\sum_{n=1}^{N}\mathbb{E}_{\boldsymbol{\epsilon}}
\left[\left\|\boldsymbol{\Delta}_i^{(n)}(\boldsymbol{\epsilon})\right\|_2^2\right]
}.
\end{equation}
Here, $N$ is the number of real samples, $\|\cdot\|_2^2$ denotes the squared Euclidean norm, and $s_i$ is the Stable Discrepancy Share at step $i$. When the total energy is nonzero, $s_i\in[0,1]$, with larger values indicating a greater stable proportion. Energies are summed across samples before taking the ratio.

In practice, we fix $N=128$ real series and draw $K=16$ noise realizations per series at each noise level, using identical noisy inputs for both views. We estimate the stable component by the sample mean of these discrepancies. To correct the upward bias from finite sampling, we subtract $1/K$ of the summed unbiased sample variances across dimensions from the squared norm of this mean. The share is computed separately for each training seed before aggregation, allowing us to assess both stability across noise realizations and the reproducibility of this stability across training seeds.

\textbf{View-Switching Ablation:} The above statistic measures discrepancy stability at individual noise levels. To examine how view selection within and outside the window affects the final generated samples, we evaluate four configurations along complete sampling trajectories: Time-Only, Frequency-Only, time outside and frequency inside the window, and frequency outside and time inside the window. We use the window defined in the main text, $\mathcal{W}=\{i:\sigma_i\leq0.35\}=\{6,\ldots,1\}$.

Let $M_{ab}^{q}$ denote extreme-fidelity metric $q$ when view $a$ is used outside the window and view $b$ inside it, where $a,b\in\{T,F\}$ denote the time and frequency views. We compute the switching effect for each segment by averaging over the two view assignments in the other segment:
\begin{align}
\delta_{\mathrm{in}}^{q}
&=
\frac{(M_{TT}^{q}-M_{TF}^{q})+(M_{FT}^{q}-M_{FF}^{q})}{2},\\
\delta_{\mathrm{out}}^{q}
&=
\frac{(M_{TT}^{q}-M_{FT}^{q})+(M_{TF}^{q}-M_{FF}^{q})}{2}.
\end{align}
Here, $\delta_{\mathrm{in}}^{q}$ and $\delta_{\mathrm{out}}^{q}$ are the average switching effects inside and outside the window, respectively. Within each run, we average their absolute values over Mag-Rate, Chg-Rate, Mag-Excess, and Chg-Excess to prevent cancellation across metrics. We then aggregate over nine runs formed by three training seeds and three sampling seeds. The ratio of the resulting mean effect magnitudes inside and outside the window gives the fold difference reported in the main text. Within each run, all four configurations share the initial noise, denoisers, and deterministic sampler, differing only in view assignment. This ablation measures the magnitude of changes in extreme fidelity, rather than improvements from switching views.

\subsection{Implementation Details of EDM-Based Diffusion}
\label{Appendix:EDM}

We provide additional implementation details of the EDM-based diffusion framework used in CrossTF. For each view $v\in\{T,F\}$, the denoiser processes the corresponding 2D representation $X_\sigma^v$ and maps its output back to the 1D time domain through a fixed, non-learned inverse transform to obtain the clean-sample estimate. Both the time- and frequency-view denoisers follow the standard EDM preconditioning and sampling configuration~\citep{karras2022elucidating}, while differing in their view-specific 2D representations.

\textbf{EDM Preconditioning.}
For each view, EDM preconditioning is applied to the corresponding 2D representation $X_\sigma^v$:
\begin{equation}
\widetilde{D}_{\theta_v}^{v}(X_\sigma^v,\sigma)
=
c_{\mathrm{skip}}X_\sigma^v
+
c_{\mathrm{out}}
F_{\theta_v}^{v}
\left(
c_{\mathrm{in}}X_\sigma^v,
c_{\mathrm{noise}}
\right),
\end{equation}
where
\begin{equation}
c_{\mathrm{skip}}
=
\frac{\sigma_{\mathrm{data}}^2}
{\sigma^2+\sigma_{\mathrm{data}}^2},
\qquad
c_{\mathrm{out}}
=
\frac{\sigma\sigma_{\mathrm{data}}}
{\sqrt{\sigma^2+\sigma_{\mathrm{data}}^2}},
\end{equation}
and
\begin{equation}
c_{\mathrm{in}}
=
\frac{1}
{\sqrt{\sigma^2+\sigma_{\mathrm{data}}^2}},
\qquad
c_{\mathrm{noise}}
=
\frac{1}{4}\log\sigma.
\end{equation}
Following the standard EDM configuration~\citep{karras2022elucidating}, we set $\sigma_{\mathrm{data}}=0.5$. The preconditioned output $\widetilde{D}_{\theta_v}^{v}$ remains in the corresponding 2D representation space. It is then mapped back to the 1D time domain through the fixed, non-learned inverse transform, yielding $D_{\theta_v}^{v}(X_\sigma^v,\sigma)\in\mathbb{R}^{T}$, the clean-sample estimate used in Section~\ref{sec:what}.

\textbf{View-Specific Denoiser Architecture.}
The time- and frequency-view denoisers are implemented as two independently trained models with the same backbone configuration. Each denoiser takes the corresponding 2D representation $X_\sigma^v$ as input and processes it using a 2D DhariwalUNet~\citep{dhariwal2021diffusion}. 
The two branches are trained independently and stored as separate checkpoints, and are combined only during CrossTF sampling.

Both branches share the same DhariwalUNet backbone configuration, with a base channel width of $64$, four resolution levels with channel multipliers $[1,2,2,2]$, and one residual block per level. For the $48\times48$ configuration, self-attention is applied at spatial resolutions $24\times24$, $12\times12$, and $6\times6$, with $64$ channels per attention head. The network uses a dropout rate of $0.1$ and adaptive scale-and-shift conditioning based on the EDM noise embedding. For each dataset, the time and frequency branches use matching spatial resolutions, with their view-specific configurations summarized in Table~\ref{tab:view_configuration}.

\begin{table}[t]
\centering  
\caption{View-specific representation configurations across datasets.}
\label{tab:view_configuration}
\small
\begin{tabular}{lcccccc}
\toprule
Dataset
& $T$
& Delay
& Emb.
& Time Res.
& $(n_{\mathrm{fft}}, h)$
& Frequency Res. \\
\midrule
\textsc{Traffic}
& 144 & 3 & 48 & $48\times48$ & $(95,3)$ & $48\times48$ \\
\textsc{Precip.}
& 90 & 2 & 32 & $32\times32$ & $(63,3)$ & $32\times32$ \\
\textsc{Build.}
& 144 & 3 & 48 & $48\times48$ & $(95,3)$ & $48\times48$ \\
\textsc{Temp.}
& 144 & 3 & 48 & $48\times48$ & $(95,3)$ & $48\times48$ \\
\textsc{Elect.}
& 144 & 3 & 48 & $48\times48$ & $(95,3)$ & $48\times48$ \\
\textsc{Transp.}
& 144 & 3 & 48 & $48\times48$ & $(95,3)$ & $48\times48$ \\
\textsc{ECG}
& 250 & 3 & 64 & $64\times64$ & $(127,4)$ & $64\times64$ \\
\bottomrule
\end{tabular}
\end{table}

\textbf{Training Configuration.}
Both view-specific denoisers are trained using the EDM denoising objective in Equation~\ref{eq:denoising_objective}. 
During training, the noise level is sampled as
\begin{equation}
\log\sigma \sim \mathcal{N}(-1.2,1.2^2),
\end{equation}
with the standard EDM weighting function
\begin{equation}
\lambda(\sigma)
=
\frac{\sigma^2+\sigma_{\mathrm{data}}^2}
{(\sigma\sigma_{\mathrm{data}})^2}.
\end{equation}
The same noise distribution, loss formulation, and network configuration are used for both views, while their view-specific representation configurations follow Table~\ref{tab:view_configuration}.

\textbf{Sampling Configuration.}
For generation, we use an 18-step Karras noise schedule~\citep{karras2022elucidating} with $\rho=7$, $\sigma_{\max}=80$, and $\sigma_{\min}=0.002$. 
We set $S_{\mathrm{churn}}=0$ and use deterministic sampling with an Euler prediction followed by a second-order Heun correction~\citep{karras2022elucidating,hairer1993solving}. 
The Heun correction is omitted for the final step ending at $\sigma=0$. 
Thus, an 18-step trajectory requires 35 denoiser evaluations for a single view.

\section{Additional Experimental Details and Results}

\subsection{Datasets} \label{appendix:datasets}
\begin{figure*}[t]
    \centering
    \begin{subfigure}[t]{0.48\textwidth}
        \centering
        \includegraphics[width=\linewidth]{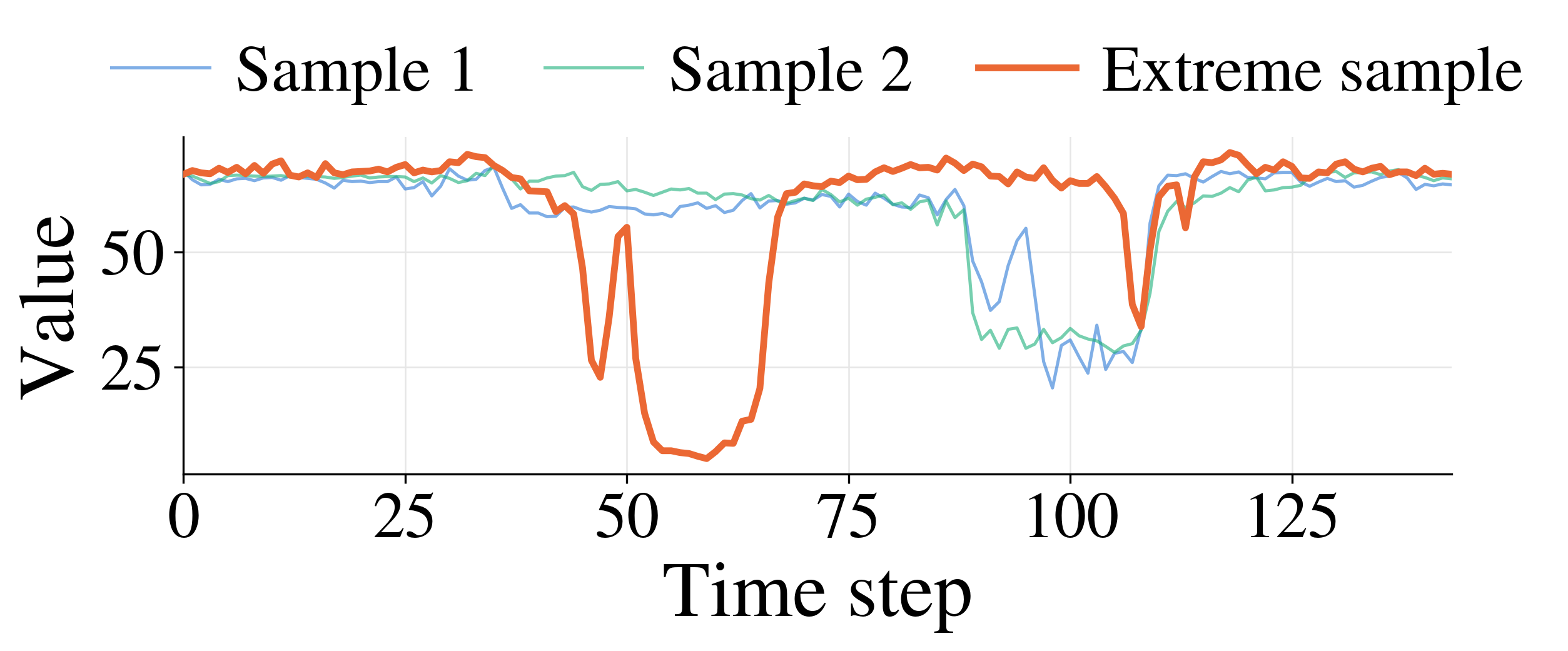}
        \caption{Traffic}\label{fig:Traffic_data_sample}
    \end{subfigure}\hfill
    \begin{subfigure}[t]{0.48\textwidth}
        \centering
        \includegraphics[width=\linewidth]{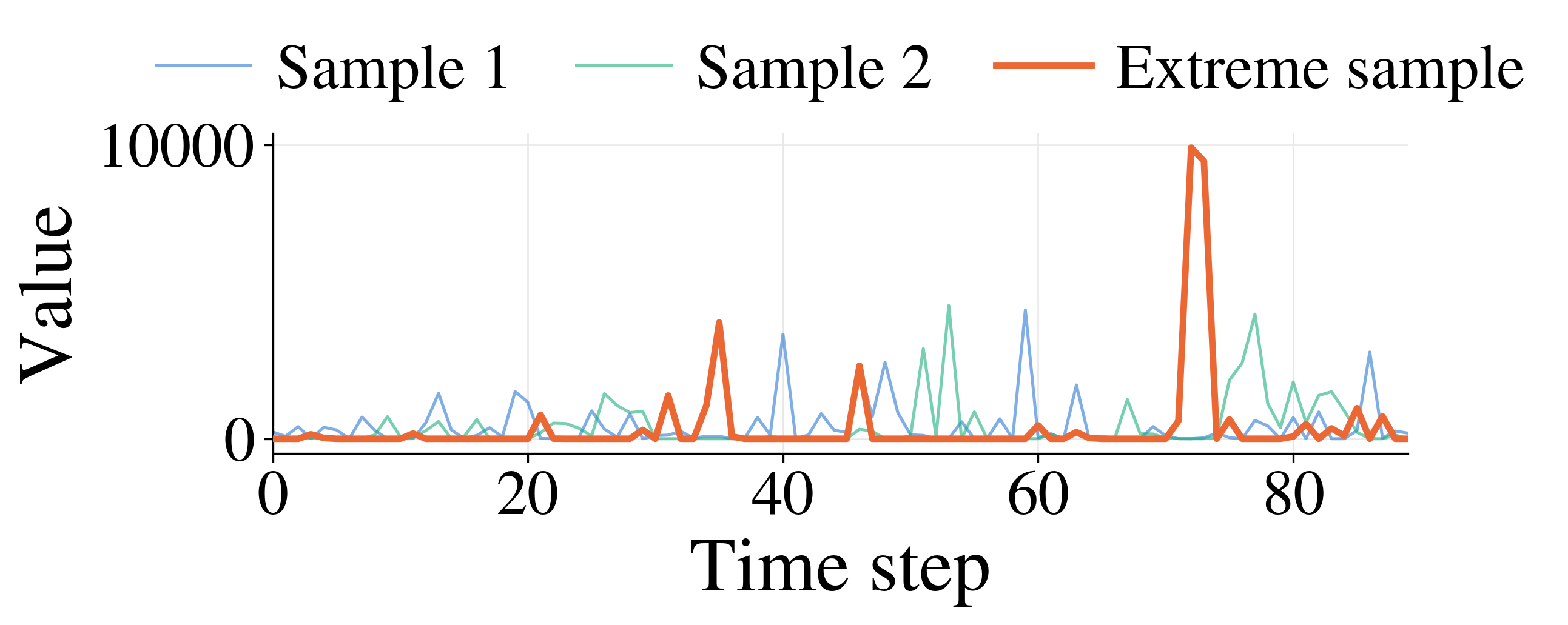}
        \caption{Precipitation}\label{fig:Precipitation_data_sample}
    \end{subfigure}

    \begin{subfigure}[t]{0.48\textwidth}
        \centering
        \includegraphics[width=\linewidth]{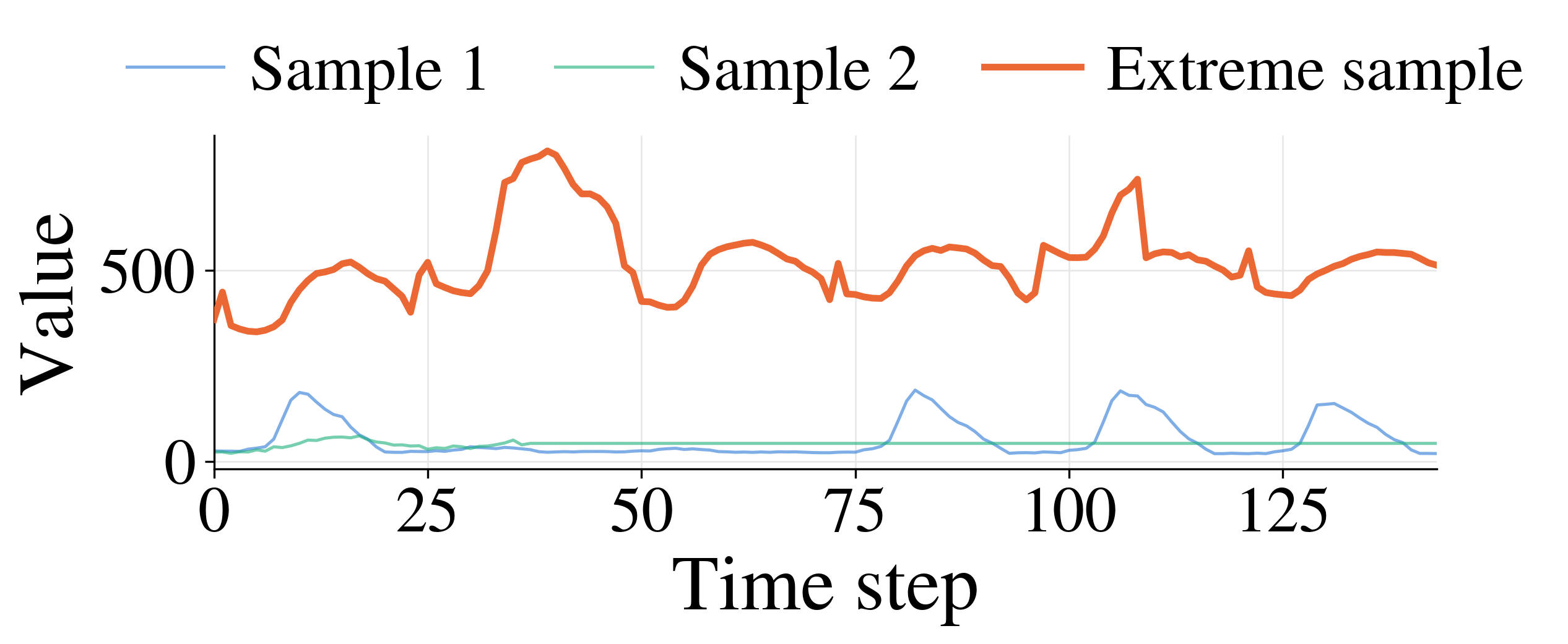}
        \caption{Building power}\label{fig:Building_data_sample}
    \end{subfigure}\hfill
    \begin{subfigure}[t]{0.48\textwidth}
        \centering
        \includegraphics[width=\linewidth]{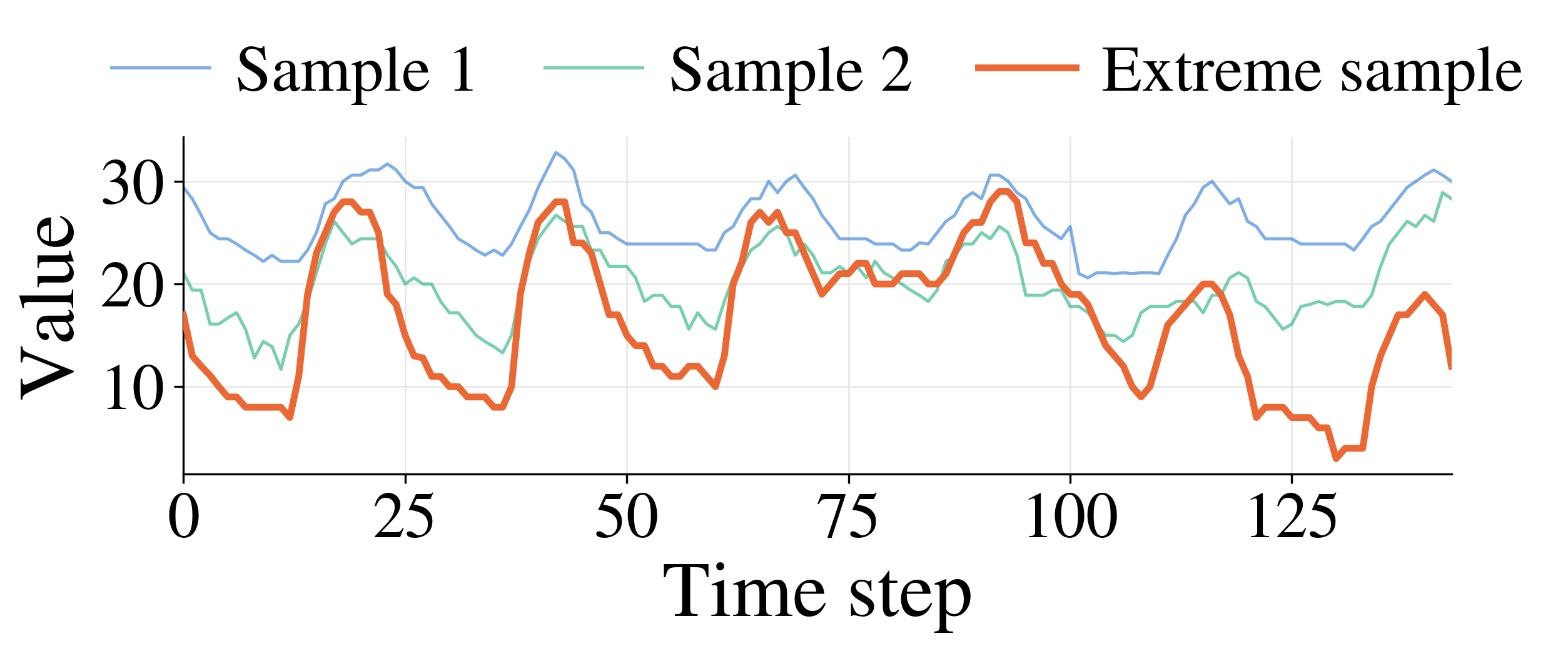}
        \caption{Temperature}\label{fig:Temperature_data_sample}
    \end{subfigure}

    \begin{subfigure}[t]{0.48\textwidth}
        \centering
        \includegraphics[width=\linewidth]{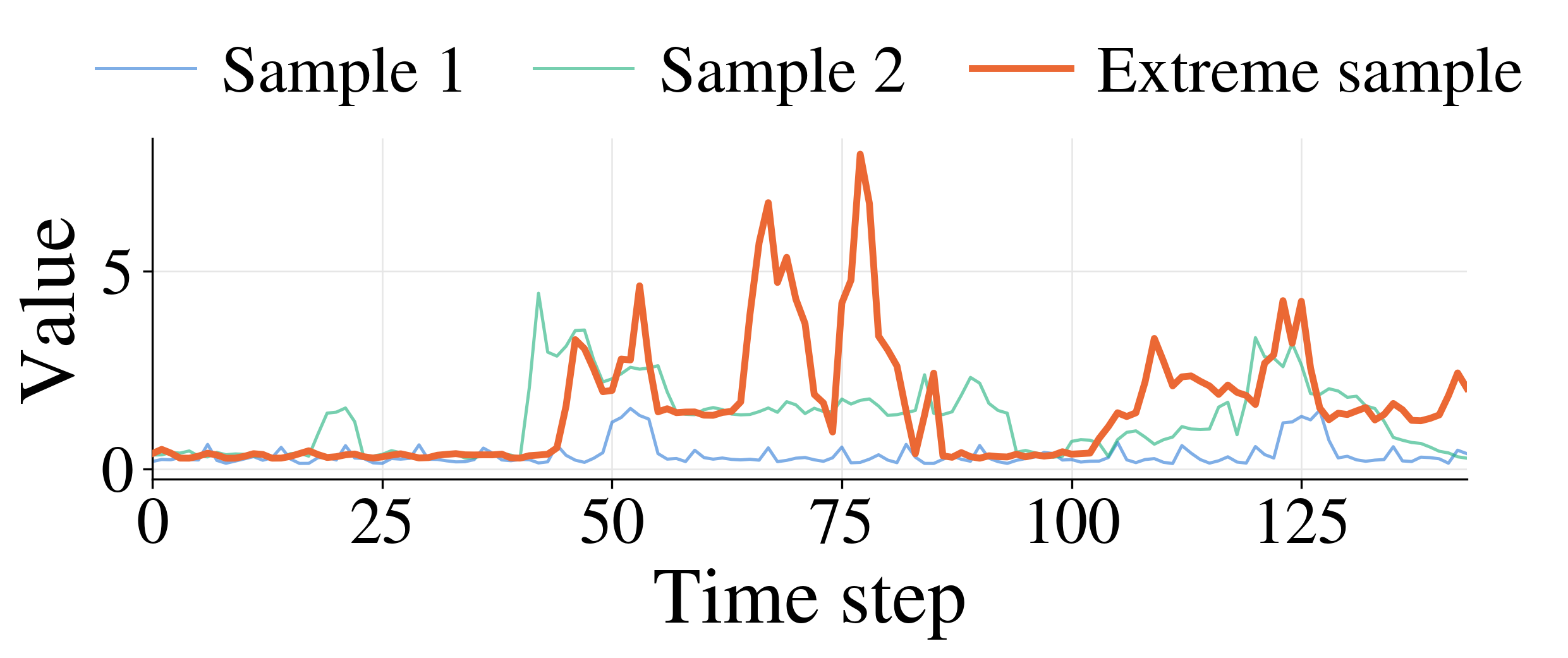}
        \caption{Electricity}\label{fig:Electricity_data_sample}
    \end{subfigure}\hfill
    \begin{subfigure}[t]{0.48\textwidth}
        \centering
        \includegraphics[width=\linewidth]{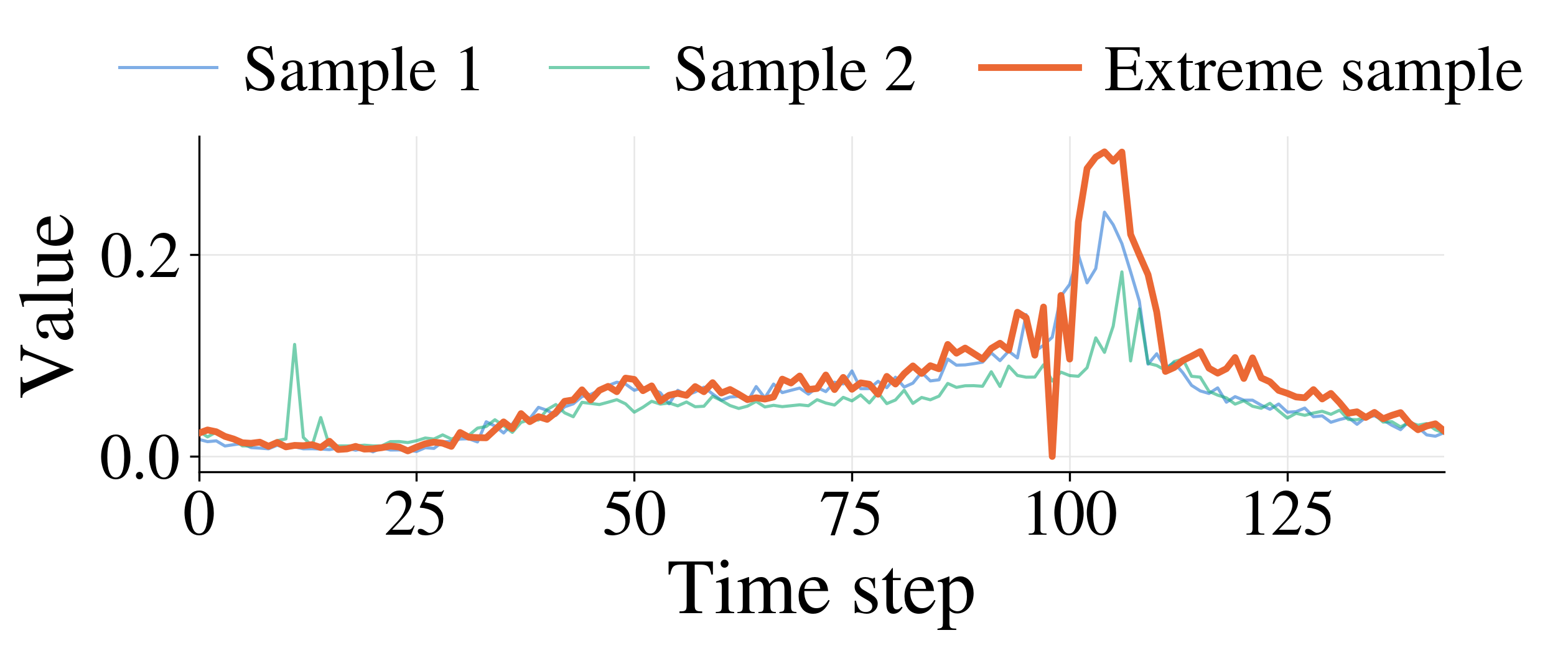}
        \caption{Transportation}\label{fig:Transportation_data_sample}
    \end{subfigure}

    \begin{subfigure}[t]{0.48\textwidth}
        \centering
        \includegraphics[width=\linewidth]{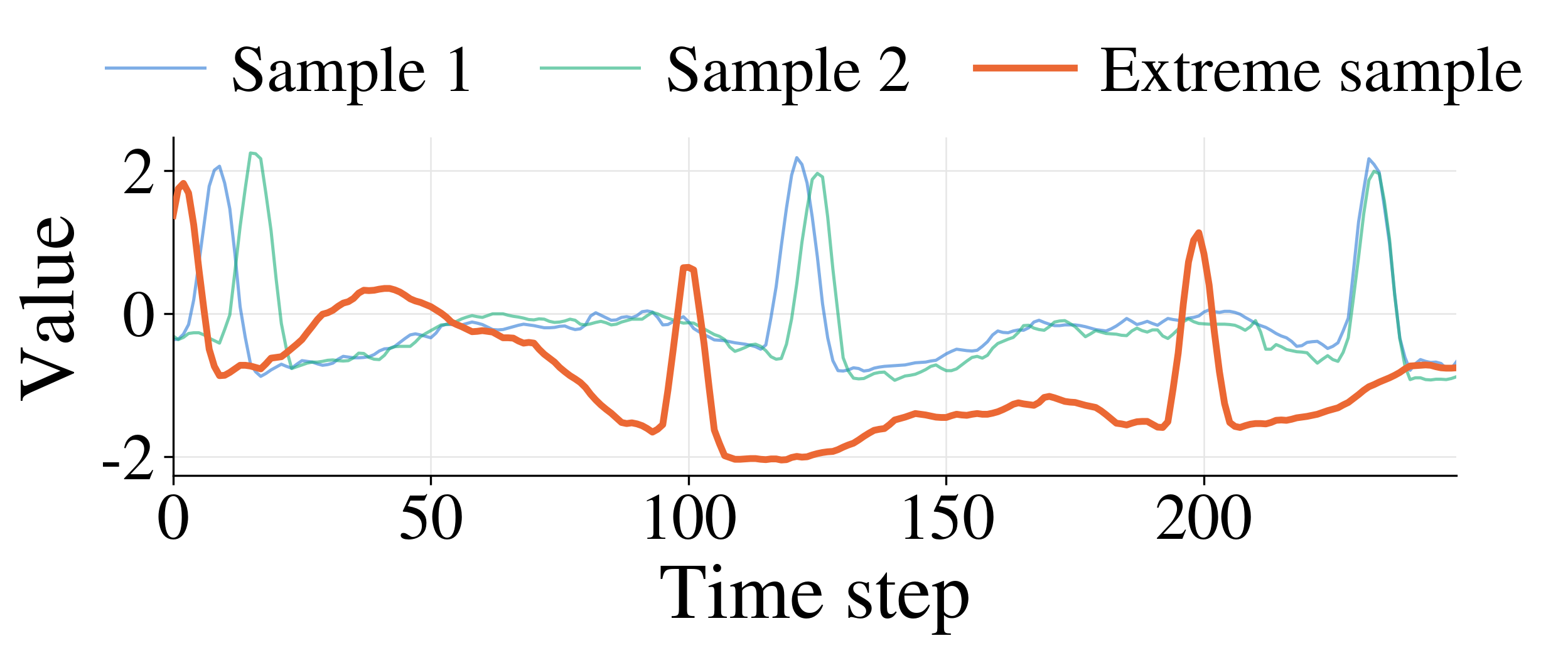}
        \caption{Electrocardiogram}\label{fig:ECG_data_sample}
    \end{subfigure}
    \caption{Representative sequences from the seven datasets described above. Each panel presents two typical samples and one sample exhibiting pronounced extreme behavior.}
\end{figure*}

\begin{itemize}

\item \textbf{Traffic}: We use the \emph{PEMS-BAY} dataset released with DCRNN~\citep{li2018dcrnn,caltranspems}, which records freeway traffic speeds from 325 loop detectors in the San Francisco Bay Area. The extremes in this dataset are the congestion events, e.g., the abrupt collapses of speed at the onset of a jam, which take a detector from free-flow velocity to stop-and-go within a few intervals and hold it there through the peak period. Speed is reported in km/h and represents the mean vehicle velocity observed by a detector within a given interval. The data are sampled every 5 minutes from January 1, 2017 to June 30, 2017. To construct our benchmark, we randomly select 10 detectors (seed 42) whose fraction of missing readings is below $0.001$, partition each detector record into daily windows, and subsample to a 10-minute resolution, so each day is represented by a sequence of length 144. Since zero denotes a missing reading, any day containing zero values is discarded (20 of 1810). Treating each detector-day pair as one sample yields a dataset of shape $(1790,144,1)$. This construction retains the strong daily commuting pattern while preserving the sharp speed drops associated with congestion. The sample visualization is shown in Fig.~\ref{fig:Traffic_data_sample}.

\item \textbf{Precipitation}: We use the \emph{Daily Weather Data} provided by Dewey~\citep{customweather2022daily}, which contains daily weather observations from ground stations. Extreme events in this dataset consist of heavy-rainfall episodes, including those associated with tropical systems, that produce substantial precipitation over one or a few days amid otherwise dry conditions. We select 88 observation stations across Florida and use daily precipitation totals, as the hourly records contain substantial missing data. Records from January 1, 2023, to December 31, 2025, are partitioned into 90-day samples, yielding a dataset of shape $(1056,90,1)$. Among the seven datasets, Precipitation exhibits the strongest zero inflation, with 59.8\% of all time steps equal to zero. Its marginal distribution therefore combines a large point mass at zero with a long right tail, and its extremes typically appear as isolated bursts rather than smooth excursions. The sample visualization is shown in Fig.~\ref{fig:Precipitation_data_sample}.

\item \textbf{Building Power}: We use the \emph{Building Data Genome Project 2} dataset~\citep{miller2020bdg2}, which records whole-building electricity consumption from non-residential buildings on university and institutional campuses. The extremes in this dataset are the peak-demand periods, e.g., the hours of heaviest HVAC and occupancy load, typically on hot weekday afternoons, during which a building draws far more power than it does at night or over a weekend. The value is reported in kWh and represents the electricity consumed by an entire building within a given interval. The data are sampled hourly from January 1, 2016 to December 31, 2017. To construct our benchmark, we randomly select 16 meters (seed 42) from the 113 meters with complete and non-degenerate records over the full period, and partition each record into non-overlapping 144-hour (six-day) windows; flat or incomplete windows are discarded (98 of 1936). Treating each meter-window pair as one sample yields a dataset of shape $(1838,144,1)$. This construction retains both daily occupancy patterns and weekday--weekend variation within each six-day window, at a finer spatial aggregation than a grid-level load series. The sample visualization is shown in Fig.~\ref{fig:Building_data_sample}.

\item \textbf{Temperature}: We use the \emph{Hourly Weather Data} provided by Dewey~\citep{customweather2022hourly}, which contains hourly climate observations across the United States since 2018. The extremes in this dataset are the cold snaps and heat spells that interrupt the ordinary diurnal and seasonal rhythm, lasting from several hours to a few days. We select five nearby stations around Jacksonville in northeastern Florida, a region frequently affected by extreme weather such as heavy rainfall and hurricanes. The value is reported in $^\circ$C and stored at a resolution of $0.1\,^\circ$C, inherited from an original reporting granularity of $1\,^\circ$F. The temperature records are divided into six-day samples, each containing 144 hourly observations $(24 \times 6)$. Using data from these five stations over the period from January 1, 2021 to December 31, 2025, we obtain 1,520 samples, yielding a dataset of shape $(1520,144,1)$, with values ranging from $-7.0$ to $44.4\,^\circ$C. Among the seven datasets, Temperature exhibits the clearest diurnal periodicity, and its extremes appear as broad, sustained excursions rather than isolated spikes. This construction captures both long-term temporal variation and local spatial variation across nearby stations. The sample visualization is shown in Fig.~\ref{fig:Temperature_data_sample}.

\item \textbf{Electricity}: We use the \emph{UCI Individual Household Electric Power Consumption} dataset~\citep{hebrail2006individual}, which records minute-level household electricity usage from a single residence near Paris, France over an extended period. Extreme events appear as short-duration load spikes, consistent with the operation of high-power appliances, that substantially exceed typical household demand. We focus on the \texttt{Global\_active\_power} variable, reported in kW, which represents the household's instantaneous active power demand, and select a continuous three-year period from January 1, 2007 to December 31, 2009. After temporal alignment and missing-value imputation, the data are resampled at 10-minute intervals, so that each day contains 144 observations. We then group the data into daily samples and retain only complete sequences, yielding a dataset of shape $(1096,144,1)$. The spikes are not tightly phase-locked to the daily pattern they ride on, which makes the extremes of this dataset hard to anticipate from its periodic component alone. The sample visualization is shown in Fig.~\ref{fig:Electricity_data_sample}.

\item \textbf{Transportation}: We use the \emph{PEMS-SF} dataset derived from the California Performance Measurement System (PeMS)~\citep{cuturi2011pems,caltranspems}, which records freeway traffic occupancy from monitoring stations in the San Francisco Bay Area. The extremes in this dataset are the heavily congested stretches of the day, during which a sensor stays covered by vehicles for a large fraction of each interval instead of the brief moments typical of free flow. The occupancy value ranges from 0 to 1 and represents the fraction of time that a sensor is occupied by vehicles within a given interval. The data are sampled every 10 minutes, so each day is represented by a sequence of length 144. Following the provided train-test split, we combine the \texttt{PEMS\_train} and \texttt{PEMS\_test} sets, obtaining 440 daily sequences in total. To construct our benchmark, we select three nearby monitoring stations and extract their daily occupancy sequences separately, treating each station-day pair as one sample. This yields a dataset of shape $(1320,144,1)$. This construction captures realistic daily traffic dynamics while retaining local variation across nearby monitoring locations. The sample visualization is shown in Fig.~\ref{fig:Transportation_data_sample}.

\item \textbf{Electrocardiogram}: We use the \emph{Long-Term ST Database} (LTST)~\citep{jager2003ltstdb,goldberger2000physionet}, which contains long-duration ambulatory ECG recordings with expert ST-related annotations. Extreme values in the selected signals appear as sharp, short-duration amplitude excursions, particularly the prominent peaks of QRS complexes. During preprocessing, we load the raw signals and annotation files and extract the first ECG lead as a one-dimensional time series. The annotations locate the abnormal ST-related intervals, which provide clinically meaningful references for informative regions. Based on these regions, we segment the signal into fixed-length windows of 250 time steps, corresponding to one second at 250 Hz, and generate candidate samples using a sliding-window strategy. To obtain representative windows with stronger signal intensity, we rank the candidates by the mean absolute amplitude within each window and randomly select 1{,}000 samples from the high-score set, yielding a dataset of shape $(1000,250,1)$. The value is reported in mV, stored at a $0.005$ mV quantization step, and ranges from $-2.045$ to $2.940$ mV. Electrocardiogram is the only physiological dataset in our benchmark and carries its most impulsive extremes. The sample visualization is shown in Fig.~\ref{fig:ECG_data_sample}.

\end{itemize}

\subsection{Metrics} \label{appendix:metrics}

To comprehensively evaluate the performance of our CrossTF, we use ten metrics from two perspectives: extreme fidelity and overall performance. All metrics are lower-is-better. Below, we provide the definitions and implementation details of each metric. Throughout, let $\mathcal{X}_{\mathrm{real}}=\{\mathbf{x}^{(i)}\}_{i=1}^{N}$ and $\mathcal{X}_{\mathrm{gen}}=\{\tilde{\mathbf{x}}^{(i)}\}_{i=1}^{N}$ denote the real and generated sets of time-series of length $T$, respectively, and let $P(\cdot)$ denote the corresponding distribution.

For \textbf{extreme fidelity}, we introduce four metrics that characterize extremes along two dimensions: \emph{how often} a series becomes extreme and \emph{how far} it exceeds the extreme threshold. We evaluate both dimensions for two notions of extremes: magnitude and change. For a series $\mathbf{x}\in\mathbb{R}^{T}$, we define the magnitude and change scores as
\begin{equation}
z_t(\mathbf{x})
=
\left|x_t-\mathrm{med}(\mathbf{x})\right|,
\quad t=1,\ldots,T,
\qquad
w_t(\mathbf{x})
=
\left|x_{t+1}-x_t\right|,
\quad t=1,\ldots,T-1,
\end{equation}
where $\mathrm{med}(\cdot)$ denotes the temporal median. Median centering provides a robust reference for measuring deviations from the typical level of each series. The scores remain in the original data units so that systematic magnitude discrepancies between the real and generated data remain visible.

We determine a single pair of thresholds from the pooled \emph{real} scores using a tail fraction of $\tau=0.05$:
\begin{equation}
u_{\mathrm{mag}}
=
Q_{1-\tau}\!\left(
\left\{z_t(\mathbf{x}^{(i)})\right\}_{i,t}
\right),
\qquad
u_{\mathrm{chg}}
=
Q_{1-\tau}\!\left(
\left\{w_t(\mathbf{x}^{(i)})\right\}_{i,t}
\right),
\end{equation}
where $Q_{1-\tau}(\cdot)$ denotes the empirical $(1-\tau)$-quantile. These thresholds are estimated only from the real data and are subsequently fixed for all generated sets. This design allows the exceedance frequency of generated samples to differ from that of the real data. In contrast, using a separate rank-based threshold for each series would fix its exceedance frequency at approximately $\tau$ by construction and therefore obscure frequency mismatches.

Each of the four quantities below produces one value per series. We compare the distributions of these per-series values using the $1$-Wasserstein distance, normalized by the standard deviation of the corresponding real values:
\begin{equation}
\mathrm{Dist}(q)
=
\frac{
W_1\!\left(
P\!\left(q(\mathcal{X}_{\mathrm{gen}})\right),
P\!\left(q(\mathcal{X}_{\mathrm{real}})\right)
\right)}
{
\mathrm{std}\!\left(q(\mathcal{X}_{\mathrm{real}})\right)
},
\end{equation}
where $q(\mathcal{X})$ denotes the collection of per-series values of quantity $q$ over the set $\mathcal{X}$.

\begin{itemize}
    \item \textbf{Mag-Rate:} Measures the discrepancy in how frequently a series exceeds the magnitude threshold. For each series, we compute a smooth exceedance rate:
    \begin{equation}
    r_{\mathrm{mag}}(\mathbf{x})
    =
    \frac{1}{T}
    \sum_{t=1}^{T}
    \sigma\!\left(
    \frac{
    z_t(\mathbf{x})-u_{\mathrm{mag}}
    }{
    \kappa\,u_{\mathrm{mag}}
    }
    \right),
    \qquad
    \mathrm{Mag\text{-}Rate}
    =
    \mathrm{Dist}\!\left(r_{\mathrm{mag}}\right),
    \end{equation}
    where $\sigma(\cdot)$ is the logistic function and $\kappa$ controls the relative width of the transition around the threshold. The hard exceedance rate $|\{t:z_t>u_{\mathrm{mag}}\}|/T$ is piecewise constant with respect to the data and therefore provides no useful gradient. The logistic relaxation provides a smooth approximation, allowing the same quantity to be used both for evaluation and as a differentiable optimization signal.

    \item \textbf{Chg-Rate:} Measures the corresponding discrepancy in the frequency of extreme changes by replacing $z_t$ and $u_{\mathrm{mag}}$ with $w_t$ and $u_{\mathrm{chg}}$:
    \begin{equation}
    r_{\mathrm{chg}}(\mathbf{x})
    =
    \frac{1}{T-1}
    \sum_{t=1}^{T-1}
    \sigma\!\left(
    \frac{
    w_t(\mathbf{x})-u_{\mathrm{chg}}
    }{
    \kappa\,u_{\mathrm{chg}}
    }
    \right),
    \qquad
    \mathrm{Chg\text{-}Rate}
    =
    \mathrm{Dist}\!\left(r_{\mathrm{chg}}\right).
    \end{equation}

    \item \textbf{Mag-Excess:} Measures the discrepancy in how far magnitude extremes extend beyond the threshold. For each series, we compute the mean excess over all exceeding time points:
    \begin{equation}
    e_{\mathrm{mag}}(\mathbf{x})
    =
    \frac{
    1
    }{
    \max\!\left(
    \left|\mathcal{T}_{\mathrm{mag}}\right|,1
    \right)
    }
    \sum_{t\in\mathcal{T}_{\mathrm{mag}}}
    \left(
    z_t(\mathbf{x})-u_{\mathrm{mag}}
    \right),
    \qquad
    \mathrm{Mag\text{-}Excess}
    =
    \mathrm{Dist}\!\left(e_{\mathrm{mag}}\right),
    \end{equation}
    where $\mathcal{T}_{\mathrm{mag}}=\{t:z_t(\mathbf{x})>u_{\mathrm{mag}}\}$ denotes the set of exceeding time points. A series with no exceedance is assigned a value of $0$ rather than being discarded. This retains the absence of extreme occurrences as part of the series-level distribution and ensures that the real and generated quantities are evaluated over the same populations.

    \item \textbf{Chg-Excess:} Measures the corresponding discrepancy in the severity of extreme changes, with $\mathcal{T}_{\mathrm{chg}}=\{t:w_t(\mathbf{x})>u_{\mathrm{chg}}\}$:
    \begin{equation}
    e_{\mathrm{chg}}(\mathbf{x})
    =
    \frac{
    1
    }{
    \max\!\left(
    \left|\mathcal{T}_{\mathrm{chg}}\right|,1
    \right)
    }
    \sum_{t\in\mathcal{T}_{\mathrm{chg}}}
    \left(
    w_t(\mathbf{x})-u_{\mathrm{chg}}
    \right),
    \qquad
    \mathrm{Chg\text{-}Excess}
    =
    \mathrm{Dist}\!\left(e_{\mathrm{chg}}\right).
    \end{equation}
\end{itemize}

For \textbf{overall performance}, we first report three fidelity metrics that measure the distributional and temporal similarity between the generated and real data, following common practice in time-series generation:

\begin{itemize}
    \item \textbf{Wasserstein Distance (Wass.)~\citep{villani2009optimal}}: Measures the discrepancy between the pooled value distributions of the real and generated data. We use the $1$-Wasserstein distance and normalize it by the standard deviation of the real values to improve comparability across datasets:
    \begin{equation}
    \mathrm{Wass.}
    =
    \frac{1}{
    \mathrm{std}\!\left(\mathcal{X}_{\mathrm{real}}\right)
    }
    \inf_{\gamma\in\Pi(P,Q)}
    \mathbb{E}_{(x,y)\sim\gamma}
    \left[
    |x-y|
    \right],
    \end{equation}
    where $P=P(\mathcal{X}_{\mathrm{gen}})$ and $Q=P(\mathcal{X}_{\mathrm{real}})$ denote the distributions of pooled generated and real values, respectively, and $\Pi(P,Q)$ is the set of joint distributions with marginals $P$ and $Q$.

\item \textbf{Marginal Distribution Difference (Marg.)~\citep{naiman2024utilizing}:} Measures the discrepancy between the marginal value distributions of the generated and real data using histogram densities. We divide the range of the real data into $B$ equal-width bins and use the same fixed bins for both sets. The score is the mean absolute difference between their binned densities:
    \begin{equation}
    \mathrm{Marg.}
    =
    \frac{1}{B}
    \sum_{b=1}^{B}
    \left|
    \hat{p}^{\,\mathrm{gen}}_{b}
    -
    \hat{p}^{\,\mathrm{real}}_{b}
    \right|,
    \end{equation}
    where $\hat{p}_{b}$ denotes the density of the $b$-th bin. We use $B=50$.

    \item \textbf{Autocorrelation Discrepancy (ACD)~\citep{ang2023tsgbench}:} Measures the discrepancy in temporal dependencies by comparing the autocorrelation functions of the real and generated data:
    \begin{equation}
    \mathrm{ACD}
    =
    \frac{1}{L}
    \sum_{l=1}^{L}
    \left|
    \rho_{\mathrm{real}}(l)
    -
    \rho_{\mathrm{gen}}(l)
    \right|,
    \end{equation}
    where $\rho(l)$ denotes the sample autocorrelation at lag $l$, averaged across all series. We use $L=50$.
\end{itemize}

We further evaluate the downstream utility of the generated data using three metrics. For these metrics, all data are min--max normalized using the per-channel range of the real set, following the evaluation conventions adopted in prior work:

\begin{itemize}
    \item \textbf{Discriminative Score (Disc)~\citep{naiman2024utilizing}:} Evaluates how distinguishable the generated sequences are from the real ones. A GRU classifier $g_\phi$ is trained to distinguish real from generated sequences, and the score is defined as the deviation of its held-out classification accuracy from chance:
    \begin{equation}
    \mathrm{Disc}
    =
    \left|
    \mathrm{Acc}\!\left(
    g_\phi;
    \mathcal{X}^{\mathrm{test}}_{\mathrm{real}},
    \mathcal{X}^{\mathrm{test}}_{\mathrm{gen}}
    \right)
    -
    0.5
    \right|,
    \end{equation}
    where $\mathrm{Acc}(\cdot)$ denotes classification accuracy on the held-out split. A score of $0$ indicates that the classifier cannot distinguish the generated data from the real data better than chance.

    \item \textbf{Predictive Score (Pred)~\citep{naiman2024utilizing}:} Evaluates whether the generated data preserve predictive temporal dynamics under the Train-on-Synthetic, Test-on-Real protocol. A GRU forecaster $f_\theta$ is trained exclusively on generated data to predict the next value from the preceding sequence:
    \begin{equation}
    \hat{x}_{t+1}
    =
    f_\theta\!\left(\mathbf{x}_{1:t}\right).
    \end{equation}
    The trained model is then evaluated on real data using the mean absolute one-step-ahead prediction error:
    \begin{equation}
    \mathrm{Pred}
    =
    \frac{1}{N(T-1)}
    \sum_{i=1}^{N}
    \sum_{t=1}^{T-1}
    \left|
    \hat{x}^{(i)}_{t+1}
    -
    x^{(i)}_{t+1}
    \right|.
    \end{equation}
    Lower values indicate that models trained on generated data generalize better to the predictive dynamics of real data.

    \item \textbf{Context-FID~\citep{yuan2024diffusion}:} Measures the Fréchet distance between contextual representations of the real and generated time series. We first train a TS2Vec encoder on the real data and then encode each real and generated sequence into a sequence-level representation. Let $\mu_r,\Sigma_r$ and $\mu_g,\Sigma_g$ denote the mean vectors and covariance matrices of the real and generated representations, respectively. The score is computed as
    \begin{equation}
    \mathrm{Context\text{-}FID}
    =
    \|\mu_r-\mu_g\|_2^2
    +
    \mathrm{Tr}
    \left(
    \Sigma_r+\Sigma_g
    -
    2(\Sigma_r\Sigma_g)^{1/2}
    \right).
    \end{equation}
\end{itemize}

\subsection{Baselines} \label{appendix:baselines}

We compare CrossTF with eleven baseline methods from six different categories. An
overview of these baselines is provided in Table~\ref{tab:baseline_statistics}. Their
implementations with detailed parameter settings are included in the
\href{https://github.com/LinJiang18/CrossTF/tree/main/baselines}{\textcolor{blue}{\textbf{baselines}}}
directory of our codebase. The details are summarized as follows:

\begin{table*}[htbp]
\centering
\caption{Overview of the baselines.}
\label{tab:baseline_statistics}
\renewcommand{\arraystretch}{1.12}
\footnotesize
\begin{tabular}{llll}
\toprule
Baseline & Category & Venue & Link \\
\midrule

TimeGAN~\citep{yoon2019time} & GAN & NeurIPS'19 &
\href{https://github.com/AlexanderVNikitin/tsgm/blob/main/tsgm/models/timeGAN.py}{Implementation Used} \\

\arrayrulecolor{gray!60}\cmidrule(lr){1-4}\arrayrulecolor{black}

KoVAE~\citep{naimangenerative} & VAE & ICLR'24 &
\href{https://github.com/azencot-group/KoVAE}{Official Implementation} \\

\arrayrulecolor{gray!60}\cmidrule(lr){1-4}\arrayrulecolor{black}

SDForger~\citep{rousseau2025forging} & LLM & NeurIPS'25 &
\href{https://github.com/IBM/fms-dgt/tree/main/fms_dgt/public/databuilders/time_series}{Official Implementation} \\

\arrayrulecolor{gray!60}\cmidrule(lr){1-4}\arrayrulecolor{black}

DiffTime~\citep{coletta2023constrained} & Diffusion & NeurIPS'23 &
\href{https://github.com/ermongroup/CSDI}{Implementation Used} \\
Diffusion-TS~\citep{yuan2024diffusion} & Diffusion & ICLR'24 &
\href{https://github.com/Y-debug-sys/Diffusion-TS}{Official Implementation} \\
FourierDiff~\citep{crabbe2024time} & Diffusion & ICML'24 &
\href{https://github.com/JonathanCrabbe/FourierDiffusion}{Official Implementation} \\
ImagenTime~\citep{naiman2024utilizing} & Diffusion & NeurIPS'24 &
\href{https://github.com/azencot-group/ImagenTime}{Official Implementation} \\

\arrayrulecolor{gray!60}\cmidrule(lr){1-4}\arrayrulecolor{black}

RF-Diffusion~\citep{chi2024rf} & TF-Based & MobiCom'24 &
\href{https://github.com/mobicom24/RF-Diffusion}{Official Implementation} \\

\arrayrulecolor{gray!60}\cmidrule(lr){1-4}\arrayrulecolor{black}

FIDE~\citep{galib2024fide} & Extreme-Aware & NeurIPS'24 &
\href{https://github.com/galib19/FIDE}{Official Implementation} \\
HeavyDiff~\citep{pandey2025heavy} & Extreme-Aware & ICLR'25 &
\href{https://github.com/Y-debug-sys/Diffusion-TS}{Implementation Used} \\
E4GEN~\citep{jiang2026e4gen} & Extreme-Aware & NeurIPS'26 &
\href{https://github.com/LinJiang18/E4GEN}{Official Implementation} \\

\bottomrule
\end{tabular}
\renewcommand{\arraystretch}{1.0}
\end{table*}

Unless otherwise stated, every baseline is trained from scratch on the full array of each
dataset with the hyperparameters of its public repository, normalizes the data internally
and inverts that transform before writing, so all generated series are evaluated in the
original units. Each baseline is run with three seeds and generates as many series as the
real set contains; no sample is ever duplicated to reach that count.

\begin{itemize}
    \item \textbf{TimeGAN}~\citep{yoon2019time}: TimeGAN embeds sequences into a latent space using a recurrent autoencoder and trains a generator with an adversarial objective together with a supervised next-step prediction loss. The original implementation is based on TensorFlow 1.x and is incompatible with our current GPU environment, so we use the \texttt{tsgm} port linked in Table~\ref{tab:baseline_statistics}. We use it without architectural modifications and follow the original repository defaults: a three-layer GRU with hidden size 24, $\gamma=1.0$, batch size 128, and Adam with a learning rate of $10^{-3}$ for all five optimizers. Each of the three training phases runs for 1000 iterations. We keep the training budget fixed in iterations rather than epochs so that datasets of different sizes receive the same number of updates.

    \item \textbf{KoVAE}~\citep{naimangenerative}: KoVAE is a variational autoencoder whose latent sequence is constrained to evolve under a linear Koopman operator. Its objective combines reconstruction, KL regularization, and a prior-prediction term that aligns the latent dynamics with the learned operator. We use the official implementation with its default settings: 600 epochs, learning rate $7\times10^{-4}$, batch size 64, three layers, latent dimension 16, hidden dimension 20, and weights $1.0/0.007/0.005$ for the three loss terms. The paper reports tuned KL and prediction weights for only two datasets, neither of which is included in our evaluation, so we use the default values for all datasets.

    \item \textbf{SDForger}~\citep{rousseau2025forging}: SDForger compresses each series into a low-dimensional ICA representation, converts the representation into text, fine-tunes a language model on the resulting corpus, and decodes sampled text back into time series. We run IBM's \texttt{fms-dgt} databuilder without modification, fine-tuning GPT-2 for 100 epochs with a learning rate of $8\times10^{-5}$ and batch size 32, and sampling with temperature 1.3. The ICA representation is selected to explain 70\% of the variance. We modify two generation-side settings to make its output comparable with the other baselines: the upstream limit of 50--100 generated series is increased to $N$, and the diversity threshold that can terminate generation early is disabled. Because the decoder performs de-duplication and interquartile filtering, the final number of generated samples can be slightly smaller than $N$; we report the number actually obtained.

    \item \textbf{DiffTime}~\citep{coletta2023constrained}: DiffTime is a denoising diffusion model for time series that supports both soft and hard constraints. We use its unconstrained generator, which is directly comparable with the other baselines. Since no public implementation is available, we take the diffusion hyperparameters from Table~5 of the paper and use the CSDI architecture on which DiffTime is based. Specifically, we use four residual layers, 64 channels, eight attention heads, a diffusion embedding dimension of 128, and 50 noise steps with a quadratic schedule from $10^{-6}$ to $0.5$, trained for 10{,}000 gradient steps with batch size 16. We use CSDI's learning rate of $10^{-3}$ rather than the $10^{-4}$ reported for DiffTime, since the latter leaves the generated mean biased by approximately one standard deviation under this training budget.

    \item \textbf{Diffusion-TS}~\citep{yuan2024diffusion}: Diffusion-TS uses an interpretable transformer that decomposes each sequence into trend and seasonal components and predicts the clean signal under a Fourier-weighted objective. We use the official implementation with three encoder and three decoder layers, $d_{\text{model}}=64$, four attention heads, 1000 diffusion steps with a cosine schedule, an $\ell_1$ objective, and 20{,}000 gradient steps with batch size 64. The released configurations target univariate sequences of length 24 and therefore do not directly cover the sequence lengths of 90--250 used in our evaluation. We extrapolate the model capacity according to how the released configurations scale with sequence length and channel count, and use the upper end of the resulting range to avoid under-parameterizing the baseline.

    \item \textbf{FourierDiff}~\citep{crabbe2024time}: FourierDiff trains a transformer-based score model with a variance-preserving SDE over the discrete Fourier coefficients of a series and maps generated samples back to the time domain through an inverse transform. We use the official implementation with $d_{\text{model}}=72$, ten layers, twelve attention heads, learning rate $10^{-3}$, batch size 64, and 1000 sampling steps. We modify two settings from the released configuration. First, we enable the frequency-domain path because the released default corresponds to the time-domain ablation rather than the proposed frequency-domain model. Second, we train for 1000 rather than 200 epochs, since 200 epochs correspond to only a few thousand updates on datasets of our size and lead to divergence on ECG. This training budget is selected once and applied consistently across all datasets rather than tuned per dataset.

    \item \textbf{ImagenTime}~\citep{naiman2024utilizing}: ImagenTime converts each time series into a two-dimensional image through an invertible transformation, generates samples using a standard image diffusion model, and maps the generated images back to time series. We use the official implementation with delay embedding (delay 4 and embedding dimension 32) at a resolution of $32\times32$, a U-Net with 128 base channels, 18 diffusion steps, batch size 8, and learning rate $10^{-4}$. CrossTF is built on the same codebase and backbone architecture, making ImagenTime a direct reference for evaluating the effect of the proposed cross-view design.

    \item \textbf{RF-Diffusion}~\citep{chi2024rf}: RF-Diffusion is a complex-valued conditional diffusion model whose forward process attenuates frequency bins at different rates, providing its time-frequency modeling mechanism. Since our data are real-valued, univariate, and unconditional, we fix the conditioning input to a zero vector and represent each real-valued series as a complex signal with zero imaginary component, without modifying the model architecture. We also increase the blur-schedule parameter from the value released for the signal-generation tasks to the value used for the other two tasks. With the former setting, the blur kernel is numerically indistinguishable from one across frequency bins on our data, effectively removing the frequency-dependent attenuation. Because the released data loader normalizes each series independently by its mean and standard deviation, we restore the amplitude of each generated series by sampling a $(\text{mean},\text{standard deviation})$ pair from the real dataset. Training otherwise follows the published settings: 32 blocks, eight attention heads, batch size 32, learning rate $10^{-3}$, and 100 diffusion steps.

    \item \textbf{FIDE}~\citep{galib2024fide}: FIDE is a conditional diffusion model that amplifies high-frequency components and regularizes the distribution of block maxima using a generalized extreme value (GEV) likelihood. After examining its released implementation, we find that \textbf{the code contains multiple implementation issues that prevent the method from being executed as described in the paper}, including: \textbf{(1)} the GEV regularization term is detached from the autograd graph before computation and therefore does not contribute gradients during backpropagation; \textbf{(2)} the ablation branch without the GEV regularizer cannot run because a returned variable is undefined; \textbf{(3)} the clean-signal estimate used for GEV regularization omits the noise-schedule scaling required by DDPM; and \textbf{(4)} some scheduled parameters and network layers are defined but never participate in training. Consequently, the training procedure executed by the released code differs from the FIDE method described in the paper, most notably because its core GEV regularization does not actually update the model parameters. To faithfully evaluate the proposed FIDE method, we correct these implementation issues while keeping the remaining model architecture, noise schedule, sampler, and major hyperparameters unchanged.

    Specifically, we recompute the GEV log-density within the autograd graph, correct the clean-signal estimate used in the denoising step, and apply a lower bound to the log-density to avoid numerical divergence on heavy-tailed datasets. The remaining model architecture, noise schedule, and sampler follow the released implementation, including a hidden size of 64, 100 diffusion steps, a regularization weight of 0.05, and a high-frequency boost of 1.1 applied to 20\% of the spectrum. We train for 2000 epochs rather than the released 400 epochs because, under its batch-size setting, each epoch contains only one full-batch update for datasets of our scale. We also do not use the released evaluation code. Its histogram-based metrics set both the number of bins and the histogram range according to the number of samples, causing the actual data to occupy only approximately 0.079\% of the specified range and roughly four effective bins. Recomputing the same samples over their actual data range increases the Jensen--Shannon divergence and KL divergence by approximately $94\times$ and $129\times$, respectively. Therefore, FIDE, like all other baselines, is evaluated using the unified evaluation pipeline described in Appendix~\ref{appendix:metrics}.

    \item \textbf{HeavyDiff}~\citep{pandey2025heavy}: HeavyDiff replaces the Gaussian noise in a diffusion model with heavy-tailed Student-$t$ noise to better accommodate heavy-tailed time-series distributions. Since no public implementation is available, we reimplement HeavyDiff on top of Diffusion-TS. The forward process draws Student-$t$ noise with $\nu=2.5$ and rescales it to match the variance assumed by the backbone's noise schedule, while the reverse process is adapted accordingly. All remaining settings follow the Diffusion-TS backbone, including 1000 diffusion steps, an $\ell_1$ objective, a cosine schedule, and the same training budget, so the primary difference from the corresponding diffusion baseline is the noise distribution.

    \item \textbf{E4GEN}~\citep{jiang2026e4gen}: E4GEN is an event-level extreme-enhanced generator that first trains a diffusion backbone, predicts extreme events, and then steers generation through a ControlNet conditioned on the predicted events. We use the released pipeline without modification except for the threshold used to define extreme events. The released configurations use the 90th percentile, which we change to the 95th percentile to match the tail fraction used by our extreme-fidelity metrics. This aligns the definition of extremes between generation and evaluation; all other method-specific settings are retained as released.
\end{itemize}

\subsection{Ablation Study} \label{appendix:ablation}
To answer RQ4, we conduct an ablation study to examine whether the performance gains of CrossTF arise from our cross-view fusion design rather than from either view alone. All settings use the same time and frequency denoisers, training configurations, and sampling procedure; the only difference is the fusion parameters $\alpha_i$ used during sampling. The full CrossTF model uses the learned step-wise fusion parameters $\alpha_i$; \textsc{Time-Only} fixes $\alpha_i=1$ at all denoising steps and uses only the time view; \textsc{Frequency-Only} fixes $\alpha_i=0$ and uses only the frequency view. This comparison directly evaluates the gains brought by cross-view fusion over single-view generation.

Tables~\ref{tab:cross_view_ablation_extreme} and Table~\ref{tab:cross_view_ablation_overall} summarize the performance of CrossTF and the two single-view variants in terms of extreme fidelity and overall fidelity, respectively. Together, the two tables show that cross-view fusion substantially improves extreme fidelity while largely maintaining overall generation quality. Specifically, for extreme fidelity, CrossTF achieves an average improvement of 18.0\% over \textsc{Time-Only} and performs better on six of the seven datasets. Compared with \textsc{Frequency-Only}, the average improvement further increases to 48.0\%, with gains observed on all seven datasets. These results indicate that the performance gains of CrossTF do not arise from either individual view alone, but from the complementarity between the two views and their cross-view fusion. In contrast, for overall fidelity, CrossTF achieves only a 0.1\% average improvement over \textsc{Time-Only}, indicating that its overall performance remains essentially unchanged. Compared with \textsc{Frequency-Only}, however, CrossTF still achieves an average improvement of 25.1\% and performs better on all seven datasets. These results are consistent with the design goal of CrossTF: leveraging cross-view complementarity to improve extreme generation while maintaining overall distributional fidelity. Table~\ref{tab:cross_view_ablation} further reports the detailed extreme- and overall-fidelity metrics across all seven datasets.

\begin{table}[htbp]
\centering
\caption{Detailed cross-view ablation on \textbf{extreme fidelity} across seven datasets. Extreme fidelity is summarized by the geometric mean of the four extreme-fidelity metrics. ``Gain'' denotes the relative improvement of CrossTF over each single-view baseline, where positive values indicate improvement. Lower is better. Best and second-best fidelity values are shown in bold and underlined, respectively.}
\label{tab:cross_view_ablation_extreme}
\fontsize{9}{10}\selectfont
\setlength{\tabcolsep}{2.5pt}
\renewcommand{\arraystretch}{1.05}
\begin{tabular}{lccccc}
\toprule
\textbf{Dataset} & \textbf{Time-Only} & \textbf{Freq.-Only} & \textbf{CrossTF} & \textbf{Gain vs. Time} & \textbf{Gain vs. Freq.} \\
\midrule
\textsc{Elect.} & 0.1531 & \underline{0.1018} & \textbf{0.0691} & +54.9\% & +32.1\% \\
\textsc{Build.} & \underline{0.1031} & 0.1280 & \textbf{0.0762} & +26.0\% & +40.4\% \\
\textsc{Precip.} & \underline{0.0963} & 0.1899 & \textbf{0.0755} & +21.7\% & +60.3\% \\
\textsc{Temp.} & \underline{0.0979} & 0.2217 & \textbf{0.0884} & +9.7\% & +60.1\% \\
\textsc{ECG} & \underline{0.1534} & 0.1667 & \textbf{0.1421} & +7.4\% & +14.8\% \\
\textsc{Transp.} & \underline{0.0512} & 0.1627 & \textbf{0.0477} & +7.0\% & +70.7\% \\
\textsc{Traffic} & \textbf{0.0341} & 0.0815 & \underline{0.0343} & -0.7\% & +57.9\% \\
\midrule
\textbf{Avg. Gain} & -- & -- & -- & \textbf{+18.0\%} & \textbf{+48.0\%} \\
\textbf{Datasets Improved} & -- & -- & -- & \textbf{6/7} & \textbf{7/7} \\
\bottomrule
\end{tabular}
\end{table}

\begin{table}[htbp]
\centering
\caption{Detailed cross-view ablation on \textbf{overall fidelity} across seven datasets. Overall fidelity is summarized by the geometric mean of the three overall-fidelity metrics. ``Gain'' denotes the relative improvement of CrossTF over each single-view baseline, where positive values indicate improvement. Lower is better. Best and second-best fidelity values are shown in bold and underlined, respectively.}
\label{tab:cross_view_ablation_overall}
\fontsize{9}{10}\selectfont
\setlength{\tabcolsep}{2.5pt}
\renewcommand{\arraystretch}{1.05}
\begin{tabular}{lccccc}
\toprule
\textbf{Dataset} & \textbf{Time-Only} & \textbf{Freq.-Only} & \textbf{CrossTF} & \textbf{Gain vs. Time} & \textbf{Gain vs. Freq.} \\
\midrule
\textsc{Elect.} & \underline{0.0142} & 0.0184 & \textbf{0.0139} & +2.3\% & +24.5\% \\
\textsc{Build.} & \textbf{0.0025} & \underline{0.0037} & \textbf{0.0025} & +0.3\% & +32.7\% \\
\textsc{Precip.} & \textbf{0.0015} & \underline{0.0031} & \textbf{0.0015} & +0.0\% & +53.3\% \\
\textsc{Temp.} & \textbf{0.0111} & 0.0133 & \underline{0.0111} & -0.6\% & +16.3\% \\
\textsc{ECG} & \textbf{0.0179} & 0.0187 & \underline{0.0180} & -0.9\% & +3.7\% \\
\textsc{Transp.} & \underline{0.0139} & 0.0189 & \textbf{0.0135} & +2.4\% & +28.5\% \\
\textsc{Traffic} & \textbf{0.0048} & 0.0059 & \underline{0.0049} & -2.7\% & +16.9\% \\
\midrule
\textbf{Avg. Gain} & -- & -- & -- & +0.1\% & \textbf{+25.1\%} \\
\textbf{Datasets Improved} & -- & -- & -- & 3/7 & \textbf{7/7} \\
\bottomrule
\end{tabular}
\vspace{-10pt}
\end{table}
\begin{table*}[t]
\centering
\caption{Detailed cross-view ablation results on extreme and overall fidelity across seven datasets. All metrics are lower-is-better. Best results are shown in bold and second-best results are underlined.}
\label{tab:cross_view_ablation}
\footnotesize
\renewcommand{\arraystretch}{1.05}
\setlength{\tabcolsep}{4.0pt}
\resizebox{\textwidth}{!}{
\begin{tabular}{llccccccc}
\toprule
\multirow{2}{*}{\textbf{Dataset}} & \multirow{2}{*}{\textbf{Method}} & \multicolumn{4}{c}{\textbf{Extreme Fidelity}} & \multicolumn{3}{c}{\textbf{Overall Fidelity}} \\
\cmidrule(lr){3-6}
\cmidrule(lr){7-9}
& & \textbf{Mag-Rate} & \textbf{Mag-Excess} & \textbf{Chg-Rate} & \textbf{Chg-Excess} & \textbf{Wass.} & \textbf{Marg.} & \textbf{ACD} \\
\midrule
\multirow{3}{*}{\textsc{Elect.}} & Time-Only & 0.1070 & 0.1991 & 0.1669 & \underline{0.1546} & \underline{0.0327} & \textbf{0.0107} & 0.0082 \\
& Frequency-Only & \textbf{0.0772} & \textbf{0.0564} & \underline{0.0818} & 0.3009 & 0.0456 & 0.0194 & \underline{0.0070} \\
& \textbf{CrossTF} & \underline{0.0802} & \underline{0.1329} & \textbf{0.0345} & \textbf{0.0619} & \textbf{0.0315} & \underline{0.0127} & \textbf{0.0067} \\
\midrule
\multirow{3}{*}{\textsc{Build.}} & Time-Only & \underline{0.0518} & \underline{0.0840} & 0.1632 & \underline{0.1587} & \underline{0.0294} & \textbf{0.0000} & \underline{0.0177} \\
& Frequency-Only & 0.0746 & 0.0952 & \underline{0.1624} & 0.2326 & 0.0388 & \textbf{0.0000} & 0.0370 \\
& \textbf{CrossTF} & \textbf{0.0401} & \textbf{0.0664} & \textbf{0.1047} & \textbf{0.1214} & \textbf{0.0292} & \textbf{0.0000} & \textbf{0.0162} \\
\midrule
\multirow{3}{*}{\textsc{Precip.}} & Time-Only & \textbf{0.0521} & 0.1496 & \textbf{0.0597} & 0.1849 & \underline{0.0283} & \textbf{0.0000} & \textbf{0.0032} \\
& Frequency-Only & 0.2823 & \underline{0.1429} & 0.2150 & \underline{0.1498} & 0.1308 & \underline{0.0001} & 0.0042 \\
& \textbf{CrossTF} & \underline{0.0555} & \textbf{0.0778} & \underline{0.0663} & \textbf{0.1134} & \textbf{0.0247} & \textbf{0.0000} & \underline{0.0035} \\
\midrule
\multirow{3}{*}{\textsc{Temp.}} & Time-Only & \underline{0.0586} & \underline{0.0503} & \textbf{0.1221} & \underline{0.2550} & \textbf{0.0461} & \textbf{0.0026} & \textbf{0.0112} \\
& Frequency-Only & 0.1238 & 0.1386 & 0.2227 & 0.6319 & 0.0505 & \underline{0.0027} & 0.0173 \\
& \textbf{CrossTF} & \textbf{0.0535} & \textbf{0.0486} & \underline{0.1703} & \textbf{0.1379} & \underline{0.0463} & \textbf{0.0026} & \underline{0.0114} \\
\midrule
\multirow{3}{*}{\textsc{ECG}} & Time-Only & 0.0787 & \textbf{0.3569} & \underline{0.0993} & 0.1988 & \textbf{0.0500} & \textbf{0.0142} & \underline{0.0081} \\
& Frequency-Only & \textbf{0.0733} & 0.5107 & 0.1291 & \underline{0.1598} & 0.0514 & 0.0160 & \textbf{0.0080} \\
& \textbf{CrossTF} & \underline{0.0738} & \underline{0.3607} & \textbf{0.0976} & \textbf{0.1568} & \underline{0.0503} & \underline{0.0143} & \underline{0.0081} \\
\midrule
\multirow{3}{*}{\textsc{Transp.}} & Time-Only & \textbf{0.0257} & \underline{0.0437} & \textbf{0.0601} & \underline{0.1022} & \underline{0.0152} & \textbf{0.0401} & 0.0044 \\
& Frequency-Only & 0.0731 & 0.1077 & 0.3410 & 0.2607 & 0.0237 & \underline{0.0702} & \textbf{0.0041} \\
& \textbf{CrossTF} & \underline{0.0270} & \textbf{0.0389} & \underline{0.0613} & \textbf{0.0804} & \textbf{0.0149} & \textbf{0.0401} & \underline{0.0042} \\
\midrule
\multirow{3}{*}{\textsc{Traffic}} & Time-Only & \underline{0.0327} & \textbf{0.0262} & \textbf{0.0387} & \underline{0.0407} & \textbf{0.0203} & \textbf{0.0014} & \textbf{0.0039} \\
& Frequency-Only & 0.0401 & 0.0322 & 0.1359 & 0.2518 & 0.0216 & \underline{0.0015} & 0.0063 \\
& \textbf{CrossTF} & \textbf{0.0323} & \underline{0.0282} & \underline{0.0416} & \textbf{0.0367} & \underline{0.0208} & \textbf{0.0014} & \underline{0.0041} \\
\bottomrule
\end{tabular}
}
\end{table*}

\subsection{Interpolation and Extrapolation of Learned Step-Wise Fusion Parameters \texorpdfstring{$\alpha_i$}{alpha\_i}}
\label{appendix:fusion_parameters}

Table~\ref{tab:stepwise_alpha} presents the learned step-wise fusion parameters $\alpha_i$ within the cross-view window across seven datasets. For clarity, interpolation and extrapolation parameters are highlighted in blue and orange, respectively. The learned parameters span both interpolation and extrapolation: among the 42 parameters, 20 fall within $0\leq\alpha_i\leq1$, while the remaining 22 satisfy $\alpha_i>1$. Notably, we do not observe any case with $\alpha_i<0$. This behavior can be explained by the different extreme-generation behaviors observed for the two views in our experiments: the time view generally achieves better extreme fidelity, whereas the frequency view tends to underestimate extremes, making optimization less likely to move beyond the frequency-view estimate along the cross-view direction. When the two views respectively overestimate and underestimate extremes, the learned parameter typically lies between them, corresponding to interpolation; when both views underestimate extremes, optimization may move beyond the time-view estimate, resulting in extrapolation with $\alpha_i>1$. Extrapolation appears on six of the seven datasets except \textsc{Elect.}, while interpolation also appears on six datasets except \textsc{Build.}, indicating that the two modes are adaptively selected across datasets and denoising steps. This suggests that restricting $\alpha_i$ to $[0,1]$ would unnecessarily constrain the search space of cross-view optimization. Furthermore, compared with using a single shared $\alpha$ across the cross-view window, the learned step-wise parameters improve extreme fidelity on all seven datasets, with gains ranging from 0.7\% to 45.4\%. These results further support the use of step-wise interpolation and extrapolation in CrossTF.

\begin{table*}[t]
\centering
\caption{Learned step-wise fusion parameters $\alpha_i$ within the cross-view window across seven datasets. The final six denoising steps ($i=6,\ldots,1$) form the cross-view window. Parameters within the interpolation range ($0\leq\alpha_i\leq1$) and extrapolation range ($\alpha_i<0$ or $\alpha_i>1$) are shown in \textcolor{interpcolor}{blue} and \textcolor{extrapcolor}{orange}, respectively. The last column reports the relative gain in extreme fidelity over using a single shared $\alpha$ across the window.}
\label{tab:stepwise_alpha}
\fontsize{9}{10}\selectfont
\setlength{\tabcolsep}{5.0pt}
\renewcommand{\arraystretch}{1.15}
\begin{tabular}{lccccccc}
\toprule
\multirow{2}{*}{\textbf{Dataset}} & \multicolumn{6}{c}{\textbf{Cross-View Window}} & \multirow{2}{*}{\textbf{Gain vs. Single $\boldsymbol{\alpha}$}} \\
\cmidrule(lr){2-7}
& $\boldsymbol{\alpha_1}$ & $\boldsymbol{\alpha_2}$ & $\boldsymbol{\alpha_3}$ & $\boldsymbol{\alpha_4}$ & $\boldsymbol{\alpha_5}$ & $\boldsymbol{\alpha_6}$ & \\
\midrule
\textsc{Elect.} & \textcolor{interpcolor}{\textbf{0.546}} & \textcolor{interpcolor}{\textbf{0.440}} & \textcolor{interpcolor}{\textbf{0.480}} & \textcolor{interpcolor}{\textbf{0.846}} & \textcolor{interpcolor}{\textbf{0.654}} & \textcolor{interpcolor}{\textbf{0.681}} & +45.4\% \\
\textsc{Build.} & \textcolor{extrapcolor}{\textbf{1.271}} & \textcolor{extrapcolor}{\textbf{1.544}} & \textcolor{extrapcolor}{\textbf{1.871}} & \textcolor{extrapcolor}{\textbf{1.905}} & \textcolor{extrapcolor}{\textbf{1.812}} & \textcolor{extrapcolor}{\textbf{1.735}} & +20.3\% \\
\textsc{Precip.} & \textcolor{interpcolor}{\textbf{0.954}} & \textcolor{extrapcolor}{\textbf{1.006}} & \textcolor{extrapcolor}{\textbf{1.101}} & \textcolor{interpcolor}{\textbf{0.979}} & \textcolor{interpcolor}{\textbf{0.701}} & \textcolor{interpcolor}{\textbf{0.404}} & +5.8\% \\
\textsc{Temp.} & \textcolor{extrapcolor}{\textbf{1.502}} & \textcolor{extrapcolor}{\textbf{1.460}} & \textcolor{extrapcolor}{\textbf{1.335}} & \textcolor{interpcolor}{\textbf{0.939}} & \textcolor{interpcolor}{\textbf{0.804}} & \textcolor{interpcolor}{\textbf{0.779}} & +8.0\% \\
\textsc{ECG} & \textcolor{interpcolor}{\textbf{0.571}} & \textcolor{interpcolor}{\textbf{0.666}} & \textcolor{interpcolor}{\textbf{0.977}} & \textcolor{extrapcolor}{\textbf{1.143}} & \textcolor{interpcolor}{\textbf{0.911}} & \textcolor{interpcolor}{\textbf{0.830}} & +0.7\% \\
\textsc{Transp.} & \textcolor{extrapcolor}{\textbf{1.061}} & \textcolor{extrapcolor}{\textbf{1.163}} & \textcolor{interpcolor}{\textbf{0.934}} & \textcolor{extrapcolor}{\textbf{1.152}} & \textcolor{extrapcolor}{\textbf{1.358}} & \textcolor{extrapcolor}{\textbf{1.207}} & +6.0\% \\
\textsc{Traffic} & \textcolor{extrapcolor}{\textbf{1.014}} & \textcolor{extrapcolor}{\textbf{1.096}} & \textcolor{interpcolor}{\textbf{0.823}} & \textcolor{extrapcolor}{\textbf{1.075}} & \textcolor{extrapcolor}{\textbf{1.252}} & \textcolor{extrapcolor}{\textbf{1.138}} & +3.6\% \\
\bottomrule
\end{tabular}
\end{table*}

\subsection{Cross-View Complementarity Beyond Model Diversity}\label{appendix:model_diversity}

To answer RQ~6, we further investigate whether the cross-view effect observed in CrossTF simply arises from general model diversity between two different denoisers, or is essentially a model ensemble effect. In other words, if we do not introduce distinct time and frequency views and instead fuse two independently trained models from the same view, can we obtain a similar extreme-sensitive direction? To examine this, we construct a same-view ensemble control. Since the time view generally achieves better extreme fidelity in our previous experiments, we perform time-time fusion using two independently trained time denoisers and compare it with time-frequency cross-view fusion. Specifically, we evaluate both settings using three independent training seeds and four sampling seeds. The cross-view and same-view settings use the same cross-view window, sampler, and $\alpha$ grid, so that the comparison mainly reflects the difference between cross-view and same-view fusion. Consistent with the aggregation used in Figure~\ref{fig:cross_view_window_effect}, we summarize extreme fidelity using the geometric mean of the four extreme-fidelity metrics. Along each direction, we measure the relative max--min range as $\alpha$ varies, as well as the endpoint gap normalized by its sampling standard error.

\begin{figure}[htbp]
    \centering
    \vspace{-10pt}
    \includegraphics[width=0.60\linewidth]{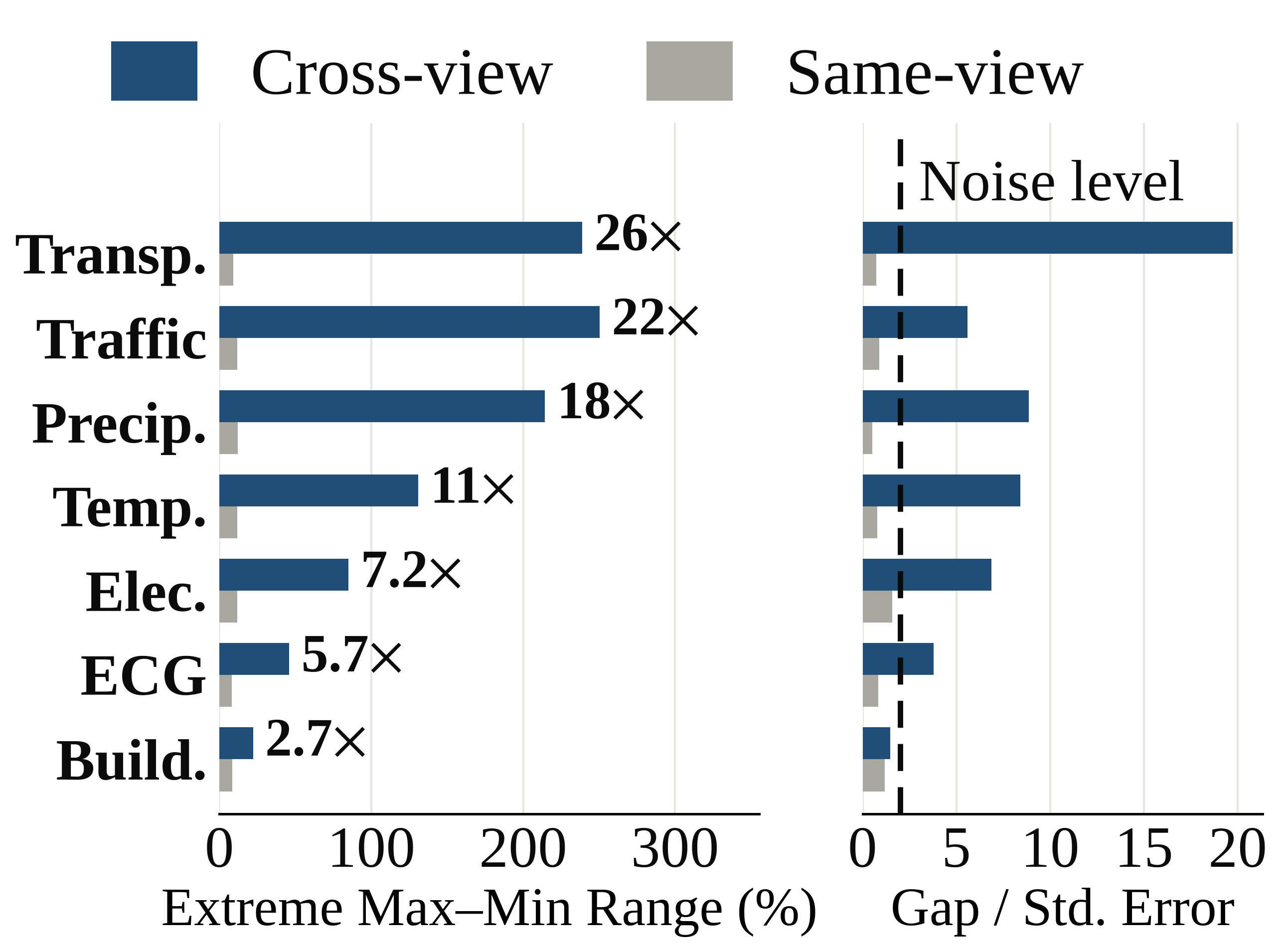}
    \caption{\textbf{Cross-view complementarity versus same-view model diversity.}
    Left: relative max--min range of extreme fidelity along cross-view and same-view directions, with annotations showing their ratio. Right: endpoint gap normalized by sampling standard error; the dashed line marks the $2$-standard-error noise threshold. Cross-view directions exhibit substantially stronger and more stable extreme-sensitive variation.}
    \label{fig:same_view_control}
    \vspace{-10pt}
\end{figure}

\begin{table*}[htbp]
\centering
\caption{Comparison of cross-view and same-view sensitivity across seven datasets. The range ratio is computed as the cross-view range divided by the same-view range, while Gap / Standard Error measures the observed gap relative to sampling variability. Values of Gap / Standard Error $\geq 2$ are considered beyond sampling noise. For the median row, the range ratio is computed from the median cross-view and
same-view ranges using the unrounded values.}
\label{tab:cross_same_view}
\fontsize{9}{10}\selectfont
\setlength{\tabcolsep}{7.0pt}
\renewcommand{\arraystretch}{1.05}
\begin{tabular}{lccccc}
\toprule
\multirow{2}{*}{\textbf{Dataset}} & \multicolumn{3}{c}{\textbf{Sensitivity Range}} & \multicolumn{2}{c}{\textbf{Gap / Standard Error}} \\
\cmidrule(lr){2-4}
\cmidrule(lr){5-6}
& \textbf{Cross-View} & \textbf{Same-View} & \textbf{Ratio} & \textbf{Cross-View} & \textbf{Same-View} \\
\midrule
\textsc{Transp.} & 238.8\% & 9.1\% & 26.28$\times$ & 19.73 & 0.72 \\
\textsc{Traffic} & 250.3\% & 11.6\% & 21.54$\times$ & 5.60 & 0.87 \\
\textsc{Precip.} & 214.3\% & 11.9\% & 18.02$\times$ & 8.85 & 0.51 \\
\textsc{Temp.} & 130.8\% & 11.7\% & 11.14$\times$ & 8.39 & 0.78 \\
\textsc{Elect.} & 85.0\% & 11.8\% & 7.17$\times$ & 6.86 & 1.59 \\
\textsc{ECG} & 45.9\% & 8.0\% & 5.73$\times$ & 3.78 & 0.82 \\
\textsc{Build.} & 22.0\% & 8.3\% & 2.66$\times$ & 1.47 & 1.17 \\
\midrule
\textbf{Median} & \textbf{130.8\%} & \textbf{11.6\%} & \textbf{11.25$\times$} & \textbf{6.86} & \textbf{0.82} \\
\midrule
\textbf{Summary} & \multicolumn{3}{c}{Cross-View $>$ Same-View: \textbf{7/7}} & \multicolumn{2}{c}{Beyond Noise ($\geq 2$): \textbf{6/7} vs. \textbf{0/7}} \\
\bottomrule
\end{tabular}
\vspace{-10pt}
\end{table*}

Figure~\ref{fig:same_view_control} and Table~\ref{tab:cross_same_view} present the comparison. The same-view ensemble direction exhibits limited variation in extreme fidelity, with a median relative range of 11.6\%, whereas the cross-view direction reaches 130.8\%, approximately 11.25$\times$ larger, and shows a greater range across all seven datasets. More importantly, the endpoint gaps along all same-view directions remain below the two-standard-error noise threshold, with a median of 0.82, indicating no distinguishable change in extreme fidelity beyond generation noise. In contrast, the cross-view direction has a median endpoint gap of 6.86 standard errors and exceeds the noise threshold on six of the seven datasets. These results show that same-view model ensembling does not produce a meaningful extreme-sensitive direction. Therefore, the \emph{Extreme-Sensitive Cross-View Asymmetry} observed in CrossTF is a characteristic of the time-frequency cross-view direction rather than a generic effect of model ensemble or model diversity.

\subsection{Computational Efficiency}\label{appendix:computation}
To evaluate the computational efficiency of CrossTF, we measure the training time and the sampling time required to generate 1,000 samples on the \textsc{Precip.} dataset, using a single \textbf{NVIDIA RTX PRO 6000 Blackwell Server} GPU for all methods. All baselines follow their original public implementations and configurations as closely as possible, without additional modifications for the efficiency comparison. As shown in Table~\ref{tab:efficiency}, \textbf{CrossTF requires only 29.5 minutes for training and 3.70 seconds to generate 1,000 samples}. Although CrossTF invokes both the time and frequency denoisers during sampling, it remains fast in practice, largely because it uses only 18 denoising steps, whereas diffusion-based methods such as E4GEN may require up to 1,000 sampling steps. Notably, methods with performance closer to CrossTF generally incur substantially higher sampling costs. For example, ImagenTime, FourierDiff, and E4GEN require approximately \textbf{6.03$\times$, 6.27$\times$, and 20.14$\times$} the sampling time of CrossTF, respectively. DiffTime is one of the few diffusion baselines that samples substantially faster than CrossTF, but its training takes 60.0 minutes, approximately 2.0$\times$ longer than CrossTF. In contrast, several lightweight models with faster sampling incur a noticeable gap in generation performance. Together with the preceding extreme- and overall-fidelity results, these findings show that \textbf{CrossTF achieves the best generation performance while maintaining practical computational efficiency}, with fast sampling and moderate training time.

\begin{table}[htbp]
\centering
\caption{Training and sampling efficiency comparison. Total training time is reported in minutes, and sampling time is measured for generating 1,000 samples. Relative sampling time is normalized by CrossTF, with CrossTF set to $1.00\times$; values above $1\times$ indicate slower sampling than CrossTF.}
\label{tab:efficiency}
\fontsize{9}{10}\selectfont
\setlength{\tabcolsep}{3.5pt}
\renewcommand{\arraystretch}{1.15}
\begin{tabular}{lccc}
\toprule
\textbf{Method} & \textbf{Training (min) $\downarrow$} & \textbf{Sampling 1K (s) $\downarrow$} & \textbf{Relative Sampling} \\
\midrule
TimeGAN & 108.0 & 2.90 & 0.78$\times$ \\
KoVAE & 6.2 & 0.03 & 0.01$\times$ \\
SDForger & 19.7 & 151.10 & 40.84$\times$ \\
DiffTime & \textbf{60.0} & 0.42 & 0.11$\times$ \\
Diffusion-TS & 8.1 & 24.10 & 6.51$\times$ \\
FourierDiff & 2.7 & 23.20 & \textbf{6.27}$\times$ \\
ImagenTime & 144.0 & 22.30 & \textbf{6.03}$\times$ \\
RF-Diffusion & 15.9 & 74.60 & 20.16$\times$ \\
FIDE & 1.4 & 0.54 & 0.15$\times$ \\
HeavyDiff & 4.5 & 17.90 & 4.84$\times$ \\
E4GEN & 6.9 & 74.50 & \textbf{20.14}$\times$ \\
\midrule
\textbf{CrossTF} & \textbf{29.5} & \textbf{3.70} & \textbf{1.00$\times$} \\
\bottomrule
\end{tabular}
\end{table}

\clearpage
\subsection{Additional Results on Extreme and Overall Fidelity}
\label{appendix:more_fidelity}

\begin{table*}[htbp]
\centering
\caption{Generation performance on the \textsc{Precip.} dataset.
All metrics are lower-is-better.}
\label{tab:precip_results}
\footnotesize
\renewcommand{\arraystretch}{0.98}
\setlength{\tabcolsep}{3.0pt}

\resizebox{\textwidth}{!}{
\begin{tabular}{cclccccccc}
\toprule

\multirow{2}{*}{\textbf{Dataset}}
& \multirow{2}{*}{\textbf{Type}}
& \multirow{2}{*}{\textbf{Method}}
& \multicolumn{4}{c}{\textbf{Extreme Fidelity}}
& \multicolumn{3}{c}{\textbf{Overall Fidelity}} \\

\cmidrule(lr){4-7}
\cmidrule(lr){8-10}

&
&
& \textbf{Mag-Rate}
& \textbf{Chg-Rate}
& \textbf{Mag-Excess}
& \textbf{Chg-Excess}
& \textbf{Wass.}
& \textbf{Marg.}
& \textbf{ACD} \\

\midrule

\multirow{12}{*}{\textsc{Precip.}}
& \textbf{GAN}
& TimeGAN
& 0.2923 & 0.4502 & 1.0694 & 1.0721 & 0.1259 & 2.18e-05 & 0.0136 \\

\cmidrule(lr){2-10}

& \textbf{VAE}
& KoVAE
& 0.9404 & 0.9779 & 0.7589 & 0.7514 & 0.2005 & 2.23e-05 & 0.0063 \\

\cmidrule(lr){2-10}

& \textbf{LLM}
& SDForger
& 0.9623 & 0.4326 & 0.9122 & 0.7071 & 0.6871 & 1.04e-04 & 0.0162 \\

\cmidrule(lr){2-10}

& \multirow{4}{*}{\textbf{Diffusion}}
& DiffTime
& 0.2816 & 0.1978 & 0.2209 & 0.1667 & 0.0607 & \underline{1.76e-05} & 0.0074 \\

&
& Diffusion-TS
& 0.9686 & 0.8395 & 0.4960 & 0.4518 & 0.2537 & 4.83e-05 & 0.0063 \\

&
& FourierDiff
& 0.5341 & 0.4998 & 0.9973 & 0.7184 & 0.5366 & 9.89e-05 & 0.0041 \\

&
& ImagenTime
& \underline{0.0815}
& \underline{0.0767}
& \underline{0.1335}
& \underline{0.1664}
& \underline{0.0347}
& 4.11e-05
& \textbf{0.0028} \\

\cmidrule(lr){2-10}

& \textbf{TF-Based}
& RF-Diffusion
& 1.7815 & 1.5310 & 1.5139 & 1.1972 & 0.6161 & 1.04e-04 & 0.0112 \\

\cmidrule(lr){2-10}

& \multirow{3}{*}{\textbf{\makecell{Extreme-\\Aware}}}
& FIDE
& 0.1792 & 0.1140 & 0.2004 & 0.2291 & 0.0668 & 3.70e-05 & 0.0101 \\

&
& HeavyDiff
& 0.4075 & 0.2475 & 2.4150 & 2.4176 & 0.1590 & \textbf{1.31e-05} & 0.0087 \\

&
& E4GEN
& 0.6380 & 0.3563 & 0.9420 & 0.7839 & 0.1896 & 3.20e-05 & 0.0073 \\

\midrule

& \textbf{Ours}
& \textbf{CrossTF}
& \textbf{0.0555}
& \textbf{0.0663}
& \textbf{0.0778}
& \textbf{0.1134}
& \textbf{0.0247}
& 3.53e-05
& \underline{0.0035} \\

\bottomrule
\end{tabular}
}
\vspace{-10pt}
\end{table*}
\begin{table*}[htbp]
\centering
\caption{Generation performance on the \textsc{Build.} dataset.
All metrics are lower-is-better.}
\label{tab:build_results}
\footnotesize
\renewcommand{\arraystretch}{0.98}
\setlength{\tabcolsep}{3.0pt}

\resizebox{\textwidth}{!}{
\begin{tabular}{cclccccccc}
\toprule

\multirow{2}{*}{\textbf{Dataset}}
& \multirow{2}{*}{\textbf{Type}}
& \multirow{2}{*}{\textbf{Method}}
& \multicolumn{4}{c}{\textbf{Extreme Fidelity}}
& \multicolumn{3}{c}{\textbf{Overall Fidelity}} \\

\cmidrule(lr){4-7}
\cmidrule(lr){8-10}

&
&
& \textbf{Mag-Rate}
& \textbf{Chg-Rate}
& \textbf{Mag-Excess}
& \textbf{Chg-Excess}
& \textbf{Wass.}
& \textbf{Marg.}
& \textbf{ACD} \\

\midrule

\multirow{12}{*}{\textsc{Build.}}
& \textbf{GAN}
& TimeGAN
& 0.3299 & 0.4519 & 0.1554 & 1.1183 & 0.1900 & 1.82e-04 & 0.2302 \\

\cmidrule(lr){2-10}

& \textbf{VAE}
& KoVAE
& 0.3504 & 0.5416 & 0.2563 & 0.4590 & 0.4135 & 3.00e-04 & 0.1417 \\

\cmidrule(lr){2-10}

& \textbf{LLM}
& SDForger
& 0.2244 & 0.5211 & 0.2174 & 0.4442 & 0.3577 & 2.23e-04 & 0.0910 \\

\cmidrule(lr){2-10}

& \multirow{4}{*}{\textbf{Diffusion}}
& DiffTime
& 0.1902 & 0.2676 & 0.1616 & 0.2347 & 0.1552 & 6.52e-05 & 0.0478 \\

&
& Diffusion-TS
& 0.2428 & 0.4228 & 0.2093 & 0.4149 & 0.2102 & 6.74e-05 & 0.0777 \\

&
& FourierDiff
& 0.1278 & 0.1830 & 0.1783 & 0.2630 & 0.1411 & 6.86e-05 & 0.0209 \\

&
& ImagenTime
& 0.0862
& \textbf{0.0638}
& 0.1162
& \underline{0.1642}
& \underline{0.0570}
& 3.40e-05
& \underline{0.0203} \\

\cmidrule(lr){2-10}

& \textbf{TF-Based}
& RF-Diffusion
& \underline{0.0606}
& 3.4284
& \textbf{0.0612}
& 1.1259
& \textbf{0.0292}
& \textbf{2.85e-05}
& 0.2370 \\

\cmidrule(lr){2-10}

& \multirow{3}{*}{\textbf{\makecell{Extreme-\\Aware}}}
& FIDE
& 0.1465 & 1.2235 & 0.1503 & 0.3541 & 0.3941 & 1.50e-04 & 0.0589 \\

&
& HeavyDiff
& 0.0838 & 0.2901 & 0.1696 & 0.3793 & 0.2409 & 1.90e-04 & 0.1667 \\

&
& E4GEN
& 0.2377 & 0.4180 & 0.2229 & 0.4254 & 0.2414 & 8.82e-05 & 0.1268 \\

\midrule

& \textbf{Ours}
& \textbf{CrossTF}
& \textbf{0.0401}
& \underline{0.1047}
& \underline{0.0664}
& \textbf{0.1214}
& \textbf{0.0292}
& \underline{3.36e-05}
& \textbf{0.0162} \\

\bottomrule
\end{tabular}
}
\vspace{-10pt}
\end{table*}
\begin{table*}[htbp]
\centering
\caption{Generation performance on the \textsc{Temp.} dataset.
All metrics are lower-is-better.}
\label{tab:temp_results}
\footnotesize
\renewcommand{\arraystretch}{0.98}
\setlength{\tabcolsep}{3.0pt}

\resizebox{\textwidth}{!}{
\begin{tabular}{cclccccccc}
\toprule

\multirow{2}{*}{\textbf{Dataset}}
& \multirow{2}{*}{\textbf{Type}}
& \multirow{2}{*}{\textbf{Method}}
& \multicolumn{4}{c}{\textbf{Extreme Fidelity}}
& \multicolumn{3}{c}{\textbf{Overall Fidelity}} \\

\cmidrule(lr){4-7}
\cmidrule(lr){8-10}

&
&
& \textbf{Mag-Rate}
& \textbf{Chg-Rate}
& \textbf{Mag-Excess}
& \textbf{Chg-Excess}
& \textbf{Wass.}
& \textbf{Marg.}
& \textbf{ACD} \\

\midrule

\multirow{12}{*}{\textsc{Temp.}}
& \textbf{GAN}
& TimeGAN
& 1.4295 & 15.7082 & 0.6670 & 5.7697 & 0.2666 & 0.0089 & 0.2837 \\

\cmidrule(lr){2-10}

& \textbf{VAE}
& KoVAE
& 0.6441 & 1.0221 & 0.7101 & 1.5098 & 0.2467 & 0.0077 & 0.1475 \\

\cmidrule(lr){2-10}

& \textbf{LLM}
& SDForger
& 0.1175 & 0.6380 & 0.1528 & 1.2320 & 0.2247 & 0.0058 & 0.0804 \\

\cmidrule(lr){2-10}

& \multirow{4}{*}{\textbf{Diffusion}}
& DiffTime
& 0.1903 & \underline{0.1545} & 0.1360 & 0.3976 & 0.3028 & 0.0065 & 0.0308 \\

&
& Diffusion-TS
& 0.1410 & 0.3274 & 0.2299 & 0.4479 & 0.1726 & 0.0040 & 0.0253 \\

&
& FourierDiff
& \underline{0.0824} & 0.2359 & 0.0899 & 0.9614 & 0.0710 & 0.0033 & 0.0185 \\

&
& ImagenTime
& 0.1018
& \textbf{0.1210}
& \underline{0.0620}
& \underline{0.2915}
& 0.1347
& 0.0033
& \underline{0.0168} \\

\cmidrule(lr){2-10}

& \textbf{TF-Based}
& RF-Diffusion
& 0.2914
& 12.0715
& 0.5461
& 3.3805
& \textbf{0.0435}
& \underline{0.0031}
& 0.2795 \\

\cmidrule(lr){2-10}

& \multirow{3}{*}{\textbf{\makecell{Extreme-\\Aware}}}
& FIDE
& 0.1382 & 0.6491 & 0.1187 & 0.5057 & 0.0779 & 0.0036 & 0.0197 \\

&
& HeavyDiff
& 0.2722 & 0.6402 & 0.1739 & 0.3053 & 0.1698 & 0.0040 & 0.0872 \\

&
& E4GEN
& 0.1672 & 0.2352 & 0.1527 & 0.6367 & 0.1741 & 0.0038 & 0.0354 \\

\midrule

& \textbf{Ours}
& \textbf{CrossTF}
& \textbf{0.0535}
& 0.1703
& \textbf{0.0486}
& \textbf{0.1379}
& \underline{0.0463}
& \textbf{0.0026}
& \textbf{0.0114} \\

\bottomrule
\end{tabular}
}

\end{table*}
\begin{table*}[htbp]
\centering
\caption{Generation performance on the \textsc{Elect.} dataset.
All metrics are lower-is-better.}
\label{tab:elect_results}
\footnotesize
\renewcommand{\arraystretch}{1.05}
\setlength{\tabcolsep}{3.0pt}

\resizebox{\textwidth}{!}{
\begin{tabular}{cclccccccc}
\toprule

\multirow{2}{*}{\textbf{Dataset}}
& \multirow{2}{*}{\textbf{Type}}
& \multirow{2}{*}{\textbf{Method}}
& \multicolumn{4}{c}{\textbf{Extreme Fidelity}}
& \multicolumn{3}{c}{\textbf{Overall Fidelity}} \\

\cmidrule(lr){4-7}
\cmidrule(lr){8-10}

&
&
& \textbf{Mag-Rate}
& \textbf{Chg-Rate}
& \textbf{Mag-Excess}
& \textbf{Chg-Excess}
& \textbf{Wass.}
& \textbf{Marg.}
& \textbf{ACD} \\

\midrule

\multirow{12}{*}{\textsc{Elect.}}
& \textbf{GAN}
& TimeGAN
& 0.2936 & 1.0724 & 0.8155 & 1.3465 & 0.1261 & 0.0522 & 0.0694 \\

\cmidrule(lr){2-10}

& \textbf{VAE}
& KoVAE
& 0.9229 & 1.1475 & 0.7631 & 0.9310 & 0.3385 & 0.0778 & 0.0422 \\

\cmidrule(lr){2-10}

& \textbf{LLM}
& SDForger
& 0.4007 & 1.4693 & 0.7653 & 1.8697 & 0.2040 & 0.0575 & 0.0408 \\

\cmidrule(lr){2-10}

& \multirow{4}{*}{\textbf{Diffusion}}
& DiffTime
& 0.2456 & 0.1872 & 0.1802 & 0.1525 & 0.1143 & 0.0256 & 0.0246 \\

&
& Diffusion-TS
& 0.3145 & 0.6035 & 0.3155 & 0.4491 & 0.3212 & 0.0510 & 0.0286 \\

&
& FourierDiff
& 0.6082 & 0.8399 & 1.0678 & 1.4358 & 0.2036 & 0.0607 & 0.0091 \\

&
& ImagenTime
& \textbf{0.0445}
& \underline{0.0621}
& \textbf{0.1278}
& \underline{0.0821}
& \underline{0.0356}
& \textbf{0.0089}
& \textbf{0.0060} \\

\cmidrule(lr){2-10}

& \textbf{TF-Based}
& RF-Diffusion
& 0.6494 & 9.0197 & 1.0747 & 0.5879 & 0.1287 & 0.0433 & 0.1610 \\

\cmidrule(lr){2-10}

& \multirow{3}{*}{\textbf{\makecell{Extreme-\\Aware}}}
& FIDE
& 0.2682 & 0.7331 & 0.1928 & 0.1588 & 0.0738 & 0.0262 & 0.0220 \\

&
& HeavyDiff
& 1.3662 & 0.6383 & 1.5234 & 0.8290 & 0.1751 & 0.0382 & 0.0247 \\

&
& E4GEN
& 0.4506 & 0.3388 & 0.3829 & 0.1602 & 0.2145 & 0.0521 & 0.0258 \\

\midrule

& \textbf{Ours}
& \textbf{CrossTF}
& \underline{0.0802}
& \textbf{0.0345}
& \underline{0.1329}
& \textbf{0.0619}
& \textbf{0.0315}
& \underline{0.0127}
& \underline{0.0067} \\

\bottomrule
\end{tabular}
}

\end{table*}
\begin{table*}[htbp]
\centering
\caption{Generation performance on the \textsc{Transp.} dataset.
All metrics are lower-is-better.}
\label{tab:transp_results}
\footnotesize
\renewcommand{\arraystretch}{1.05}
\setlength{\tabcolsep}{3.0pt}

\resizebox{\textwidth}{!}{
\begin{tabular}{cclccccccc}
\toprule

\multirow{2}{*}{\textbf{Dataset}}
& \multirow{2}{*}{\textbf{Type}}
& \multirow{2}{*}{\textbf{Method}}
& \multicolumn{4}{c}{\textbf{Extreme Fidelity}}
& \multicolumn{3}{c}{\textbf{Overall Fidelity}} \\

\cmidrule(lr){4-7}
\cmidrule(lr){8-10}

&
&
& \textbf{Mag-Rate}
& \textbf{Chg-Rate}
& \textbf{Mag-Excess}
& \textbf{Chg-Excess}
& \textbf{Wass.}
& \textbf{Marg.}
& \textbf{ACD} \\

\midrule

\multirow{12}{*}{\textsc{Transp.}}
& \textbf{GAN}
& TimeGAN
& 0.2161 & 0.8154 & 0.4333 & 0.7877 & 0.1951 & 0.5131 & 0.0298 \\

\cmidrule(lr){2-10}

& \textbf{VAE}
& KoVAE
& 0.6015 & 0.8455 & 0.6152 & 0.7580 & 0.2056 & 0.5539 & 0.0635 \\

\cmidrule(lr){2-10}

& \textbf{LLM}
& SDForger
& 0.1201 & 0.3637 & 0.4108 & 0.6160 & 0.1127 & 0.2715 & 0.0268 \\

\cmidrule(lr){2-10}

& \multirow{4}{*}{\textbf{Diffusion}}
& DiffTime
& 0.2473 & 0.2002 & 0.2661 & 0.1386 & 0.1560 & 0.3159 & 0.0194 \\

&
& Diffusion-TS
& 0.2835 & 0.5090 & 0.2599 & 0.3218 & 0.1393 & 0.1938 & 0.0391 \\

&
& FourierDiff
& 0.4169 & 0.5539 & 0.5616 & 0.7058 & 0.0974 & 0.1928 & 0.0591 \\

&
& ImagenTime
& \underline{0.0760}
& \underline{0.0687}
& \underline{0.0425}
& \underline{0.1032}
& \underline{0.0272}
& \underline{0.0431}
& \textbf{0.0027} \\

\cmidrule(lr){2-10}

& \textbf{TF-Based}
& RF-Diffusion
& 0.6752 & 9.3573 & 0.4048 & 1.0099 & 0.1680 & 0.7449 & 0.3004 \\

\cmidrule(lr){2-10}

& \multirow{3}{*}{\textbf{\makecell{Extreme-\\Aware}}}
& FIDE
& 0.2395 & 0.5854 & 0.2632 & 0.2071 & 0.1054 & 0.1259 & 0.0221 \\

&
& HeavyDiff
& 0.1595 & 0.1009 & 0.3101 & 0.4885 & 0.0569 & 0.1948 & 0.0332 \\

&
& E4GEN
& 0.1359 & 0.3536 & 0.1970 & 0.2078 & 0.0539 & 0.2168 & 0.0285 \\

\midrule

& \textbf{Ours}
& \textbf{CrossTF}
& \textbf{0.0270}
& \textbf{0.0613}
& \textbf{0.0389}
& \textbf{0.0804}
& \textbf{0.0149}
& \textbf{0.0401}
& \underline{0.0042} \\

\bottomrule
\end{tabular}
}

\end{table*}
\begin{table*}[htbp]
\centering
\caption{Generation performance on the \textsc{ECG} dataset.
All metrics are lower-is-better.}
\label{tab:ecg_results}
\footnotesize
\renewcommand{\arraystretch}{1.05}
\setlength{\tabcolsep}{3.0pt}

\resizebox{\textwidth}{!}{
\begin{tabular}{cclccccccc}
\toprule

\multirow{2}{*}{\textbf{Dataset}}
& \multirow{2}{*}{\textbf{Type}}
& \multirow{2}{*}{\textbf{Method}}
& \multicolumn{4}{c}{\textbf{Extreme Fidelity}}
& \multicolumn{3}{c}{\textbf{Overall Fidelity}} \\

\cmidrule(lr){4-7}
\cmidrule(lr){8-10}

&
&
& \textbf{Mag-Rate}
& \textbf{Chg-Rate}
& \textbf{Mag-Excess}
& \textbf{Chg-Excess}
& \textbf{Wass.}
& \textbf{Marg.}
& \textbf{ACD} \\

\midrule

\multirow{12}{*}{\textsc{ECG}}
& \textbf{GAN}
& TimeGAN
& 0.8675 & 0.4654 & 0.9233 & 1.7869 & 0.0886 & 0.0581 & 0.0282 \\

\cmidrule(lr){2-10}

& \textbf{VAE}
& KoVAE
& 2.7715 & 1.9206 & 3.7157 & 2.1064 & 0.3909 & 0.1586 & 0.1084 \\

\cmidrule(lr){2-10}

& \textbf{LLM}
& SDForger
& 1.9948 & 1.8672 & 3.4791 & 2.1706 & 0.3934 & 0.1243 & 0.0546 \\

\cmidrule(lr){2-10}

& \multirow{4}{*}{\textbf{Diffusion}}
& DiffTime
& 0.5198 & 0.6658 & 0.3663 & 0.7390 & 0.2807 & 0.0723 & 0.0298 \\

&
& Diffusion-TS
& 0.4689 & 0.5354 & \underline{0.2841} & 0.5571 & 0.1496 & 0.0440 & \underline{0.0153} \\

&
& FourierDiff
& 2.2049 & 1.7647 & 3.6607 & 1.8019 & 0.3620 & 0.1252 & \underline{0.0153} \\

&
& ImagenTime
& \underline{0.1214}
& \underline{0.1249}
& \textbf{0.2705}
& \underline{0.2603}
& \underline{0.0637}
& \underline{0.0149}
& 0.0157 \\

\cmidrule(lr){2-10}

& \textbf{TF-Based}
& RF-Diffusion
& 2.7087 & 19.3703 & 4.6889 & 11.1231 & 0.3463 & 0.1566 & 0.1122 \\

\cmidrule(lr){2-10}

& \multirow{3}{*}{\textbf{\makecell{Extreme-\\Aware}}}
& FIDE
& 0.7168 & 21.8706 & 0.6882 & 14.4202 & 0.4429 & 0.1240 & 0.1232 \\

&
& HeavyDiff
& 0.2136 & 0.7832 & 0.7308 & 0.7663 & 0.0822 & 0.0500 & 0.0515 \\

&
& E4GEN
& 0.2054 & 0.5869 & 0.8813 & 0.4068 & 0.1212 & 0.0446 & 0.0331 \\

\midrule

& \textbf{Ours}
& \textbf{CrossTF}
& \textbf{0.0738}
& \textbf{0.0976}
& 0.3607
& \textbf{0.1568}
& \textbf{0.0503}
& \textbf{0.0143}
& \textbf{0.0081} \\

\bottomrule
\end{tabular}
}

\end{table*}

\clearpage
\subsection{Additional Results about Downstream Utility}
Table~\ref{tab:downstream_utility} reports the downstream utility results across seven datasets. Since \textsc{Disc}, \textsc{Pred}, and \textsc{Context-FID} involve stochastic training, their scores can vary across random seeds. To avoid over-interpreting small stochastic differences, we additionally consider the performance range across seeds: a result is regarded as tied with the best method when their ranges overlap, as indicated by $\dagger$ in the table. This is particularly important for \textsc{Disc}, where high-quality generated samples can make the classifier approach random guessing at 50\% accuracy; consequently, small seed-level variations in classification accuracy can noticeably affect the reported score. Under this criterion, CrossTF achieves the best or tied-best result in 18 of the 21 dataset--metric comparisons, including all seven datasets for \textsc{Context-FID}, six for \textsc{Pred}, and five for \textsc{Disc}.
\label{appendix:downstream}

\begin{table*}[htbp]
\centering
\caption{\textbf{Downstream task utility} across seven real-world datasets.
\textbf{Bold} denotes the best or tied-best result.
A result is regarded as tied with the best method ($\dagger$) when their performance ranges across random seeds overlap.
The last column reports the number of best or tied-best results.}
\label{tab:downstream_utility}

\fontsize{8}{10}\selectfont
\setlength{\tabcolsep}{2.5pt}
\renewcommand{\arraystretch}{1.10}

\begin{tabular*}{\textwidth}{
@{\extracolsep{\fill}}
llcccccccc
@{}
}
\toprule
\textbf{Metric} & \textbf{Method}
& \textbf{Traffic}
& \textbf{Precip.}
& \textbf{Build.}
& \textbf{Temp.}
& \textbf{Elect.}
& \textbf{Transp.}
& \textbf{ECG}
& \textbf{\#Best} \\
\midrule

\multirow{12}{*}{\textsc{Disc} $\downarrow$}
& \textbf{CrossTF (Ours)}
& \textbf{0.0126}$^{\dagger}$
& \textbf{0.0212}$^{\dagger}$
& 0.0466
& 0.0088
& \textbf{0.0470}$^{\dagger}$
& \textbf{0.0189}$^{\dagger}$
& \textbf{0.0342}$^{\dagger}$
& 5 \\

& ImagenTime
& \textbf{0.0396}$^{\dagger}$
& \textbf{0.0157}$^{\dagger}$
& \textbf{0.0086}$^{\dagger}$
& 0.0214
& \textbf{0.0295}$^{\dagger}$
& \textbf{0.0076}$^{\dagger}$
& \textbf{0.0067}$^{\dagger}$
& 6 \\

& DiffTime
& 0.0968
& \textbf{0.1053}$^{\dagger}$
& 0.0784
& 0.1266
& \textbf{0.1000}$^{\dagger}$
& 0.1307
& \textbf{0.1867}$^{\dagger}$
& 3 \\

& FIDE
& 0.3897
& 0.0354
& 0.2595
& \textbf{0.0016}
& \textbf{0.0364}$^{\dagger}$
& 0.1913
& 0.2550
& 2 \\

& FourierDiff
& \textbf{0.0209}$^{\dagger}$
& 0.4929
& 0.0082
& 0.0263
& 0.1091
& 0.1989
& 0.3025
& 1 \\

& E4GEN
& 0.3240
& 0.3325
& 0.0992
& 0.1053
& 0.2038
& \textbf{0.1780}$^{\dagger}$
& 0.2658
& 1 \\

& Diffusion-TS
& 0.1280
& 0.3176
& 0.0466
& 0.0482
& 0.2023
& 0.0928
& 0.1425
& 0 \\

& HeavyDiff
& 0.1676
& 0.3514
& 0.1889
& 0.1941
& 0.3682
& 0.2557
& 0.3175
& 0 \\

& SDForger
& \textbf{0.0140}$^{\dagger}$
& 0.4929
& \textbf{0.0054}$^{\dagger}$
& 0.0082
& 0.3318
& 0.1648
& \textbf{0.0125}$^{\dagger}$
& 3 \\

& KoVAE
& 0.3198
& 0.1108
& 0.0815
& 0.2056
& 0.2341
& 0.1761
& 0.4700
& 0 \\

& RF-Diffusion
& 0.3687
& 0.4882
& 0.3614
& 0.4737
& 0.4318
& 0.4886
& 0.0325
& 0 \\

& TimeGAN
& 0.4763
& \textbf{0.0283}$^{\dagger}$
& 0.4728
& 0.4342
& 0.4364
& 0.3769
& 0.0750
& 1 \\

\midrule

\multirow{12}{*}{\textsc{Pred} $\downarrow$}
& \textbf{CrossTF (Ours)}
& \textbf{0.0198}$^{\dagger}$
& \textbf{0.0325}$^{\dagger}$
& \textbf{0.0026}$^{\dagger}$
& 0.0161
& \textbf{0.0298}$^{\dagger}$
& \textbf{0.0095}$^{\dagger}$
& \textbf{0.0069}
& 6 \\

& ImagenTime
& \textbf{0.0199}$^{\dagger}$
& \textbf{0.0326}$^{\dagger}$
& \textbf{0.0026}$^{\dagger}$
& 0.0161
& \textbf{0.0298}$^{\dagger}$
& \textbf{0.0095}$^{\dagger}$
& 0.0072
& 5\\

& DiffTime
& 0.0201
& 0.0327
& \textbf{0.0027}$^{\dagger}$
& 0.0160
& 0.0300
& 0.0096
& 0.0080
& 1 \\

& FIDE
& 0.0378
& 0.0333
& 0.0036
& 0.0169
& 0.0338
& 0.0103
& 0.0644
& 0 \\

& FourierDiff
& 0.0199
& 0.0439
& 0.0027
& 0.0158
& 0.0310
& 0.0097
& 0.0070
& 0 \\

& E4GEN
& \textbf{0.0199}$^{\dagger}$
& 0.0328
& 0.0028
& 0.0159
& 0.0299
& 0.0096
& 0.0078
& 1 \\

& Diffusion-TS
& 0.0200
& 0.0329
& 0.0027
& \textbf{0.0158}
& \textbf{0.0298}$^{\dagger}$
& 0.0096
& 0.0077
& 2 \\

& HeavyDiff
& 0.0202
& 0.0328
& 0.0031
& 0.0161
& 0.0299
& 0.0096
& 0.0083
& 0\\

& SDForger
& 0.0226
& 0.0442
& 0.0031
& 0.0176
& 0.0368
& 0.0100
& 0.0100
& 0\\

& KoVAE
& 0.0229
& 0.0346
& 0.0028
& 0.0185
& 0.0310
& 0.0109
& 0.0111
& 0\\

& RF-Diffusion
& 0.0501
& 0.0427
& 0.0085
& 0.0620
& 0.0726
& 0.0433
& 0.0675
& 0 \\

& TimeGAN
& 0.0283
& 0.0329
& 0.0040
& 0.1077
& 0.0362
& 0.0149
& 0.0085
& 0 \\

\midrule

\multirow{12}{*}{\textsc{Context-FID} $\downarrow$}
& \textbf{CrossTF (Ours)}
& \textbf{0.0112}
& \textbf{0.0074}$^{\dagger}$
& \textbf{0.0019}
& \textbf{0.0254}
& \textbf{0.0303}$^{\dagger}$
& \textbf{0.0068}$^{\dagger}$
& \textbf{0.0485}$^{\dagger}$
& 7\\

& ImagenTime
& 0.0325
& \textbf{0.0089}$^{\dagger}$
& 0.0109
& 0.1110
& \textbf{0.0306}$^{\dagger}$
& \textbf{0.0098}$^{\dagger}$
& \textbf{0.0710}$^{\dagger}$
& 4\\

& DiffTime
& 0.1379
& 0.0462
& 0.0372
& 0.6744
& 0.1356
& 0.1525
& 0.3918
& 0 \\

& FIDE
& 1.8391
& 0.0590
& 0.1642
& 0.1308
& 0.1071
& 0.2239
& 5.3355
& 0\\

& FourierDiff
& 0.0394
& 0.6732
& 0.1911
& 0.0697
& 1.4993
& 0.4780
& 1.4689
& 0\\

& E4GEN
& 0.1550
& 0.0480
& 0.0517
& 0.2065
& 0.2987
& 0.0841
& 0.2622
& 0\\

& Diffusion-TS
& 0.4602
& 0.3729
& 0.0470
& 0.2125
& 0.7585
& 0.2326
& 0.1749
& 0\\

& HeavyDiff
& 0.5811
& 1.0210
& 0.0344
& 0.6680
& 0.7900
& 0.1154
& 0.5741
& 0\\

& SDForger
& 11.6116
& 1.5288
& 0.0996
& 0.9860
& 7.3520
& 0.4737
& 2.4966
& 0 \\

& KoVAE
& 3.1085
& 0.3765
& 0.1826
& 0.8379
& 0.9892
& 0.9565
& 4.4922
& 0 \\

& RF-Diffusion
& 0.7904
& 1.0919
& 0.0106
& 1.9811
& 0.9143
& 1.5682
& 2.0688
& 0 \\

& TimeGAN
& 2.0369
& 0.6872
& 0.0485
& 4.5985
& 0.6189
& 0.6625
& 0.5203
& 0 \\

\bottomrule
\end{tabular*}
\end{table*}

\clearpage
\subsection{Visualization of Generated Distributions}
\label{appendix:visualization}
For each dataset, we compare the distribution of real sequences with those generated by CrossTF and four representative baselines, using three complementary views: a PCA projection, a t-SNE embedding, and a kernel density estimate of the marginal value distribution. We visualize a representative subset of baselines for clarity, while quantitative results for the full set are reported in Appendix~\ref{appendix:more_fidelity}. Real data are shown in blue and generated data in orange in the projection plots, and as solid and dashed curves, respectively, in the density plots.

\begin{figure}[htbp]
\centering
\begin{minipage}[t]{\textwidth}\centering
\includegraphics[width=0.9\linewidth]{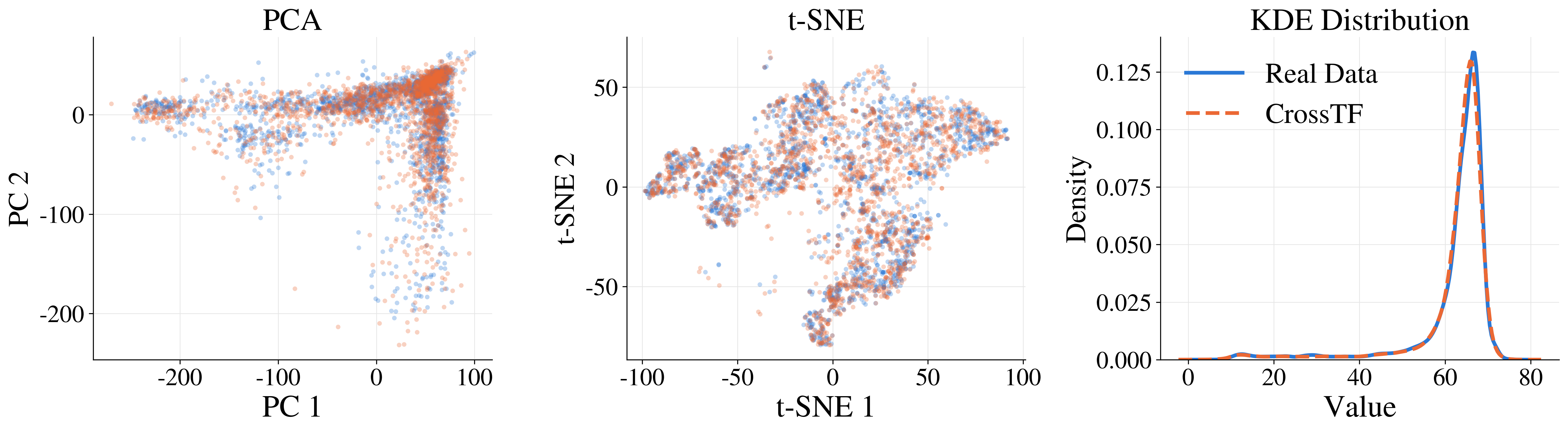}\\[-0pt]{\small \textbf{CrossTF (Ours)}}
\end{minipage}\\[8pt]
\begin{minipage}[t]{\textwidth}\centering
\includegraphics[width=0.9\linewidth]{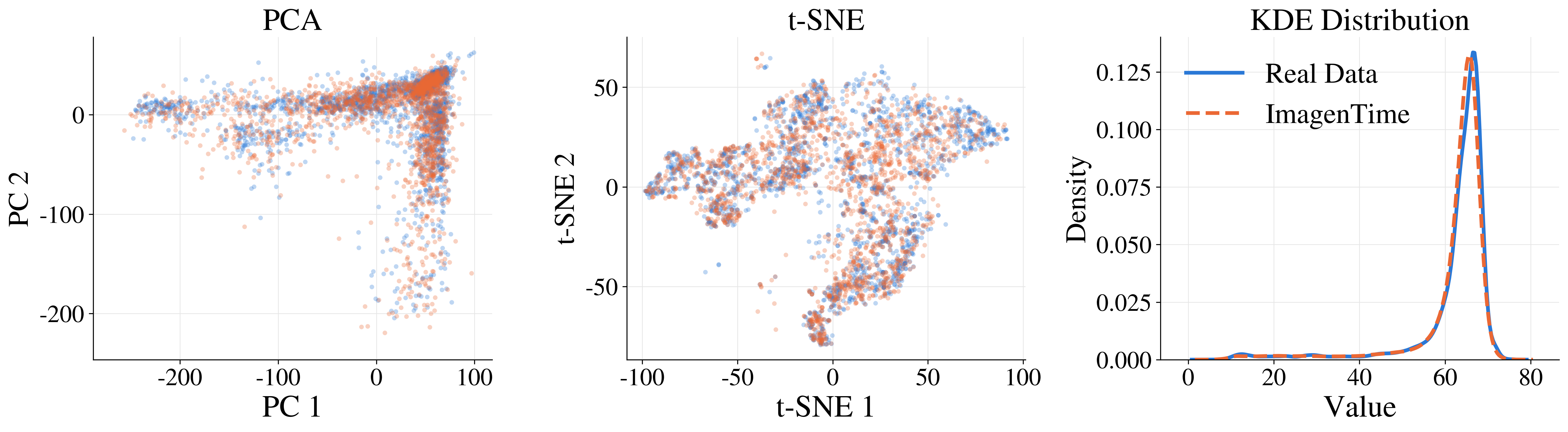}\\[-0pt]{\small ImagenTime}
\end{minipage}\\[8pt]
\begin{minipage}[t]{\textwidth}\centering
\includegraphics[width=0.9\linewidth]{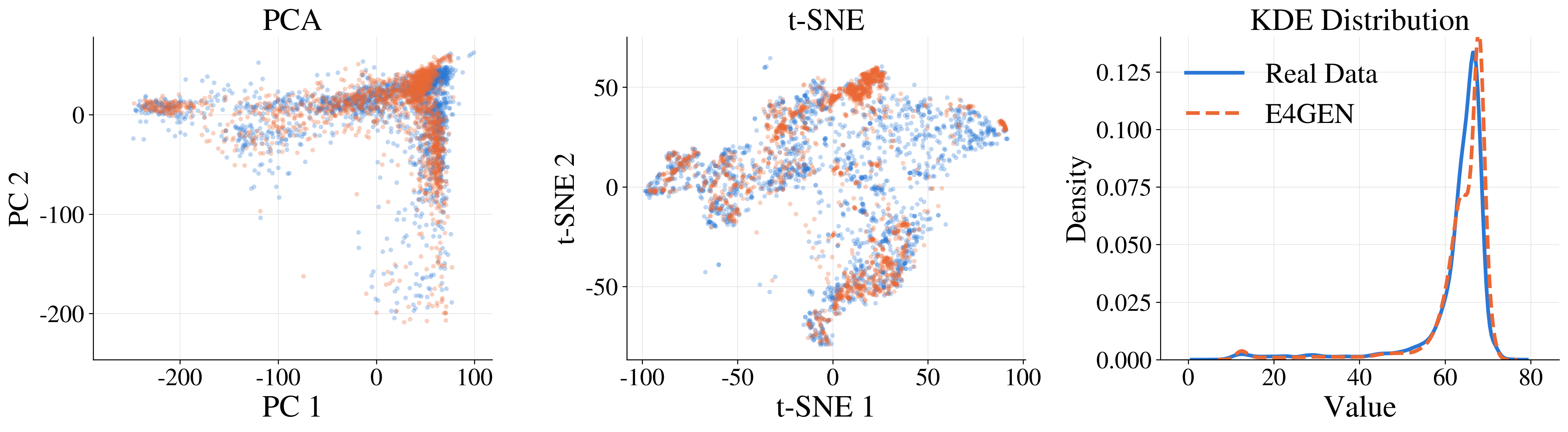}\\[-0pt]{\small E4GEN}
\end{minipage}\\[8pt]
\begin{minipage}[t]{\textwidth}\centering
\includegraphics[width=0.9\linewidth]{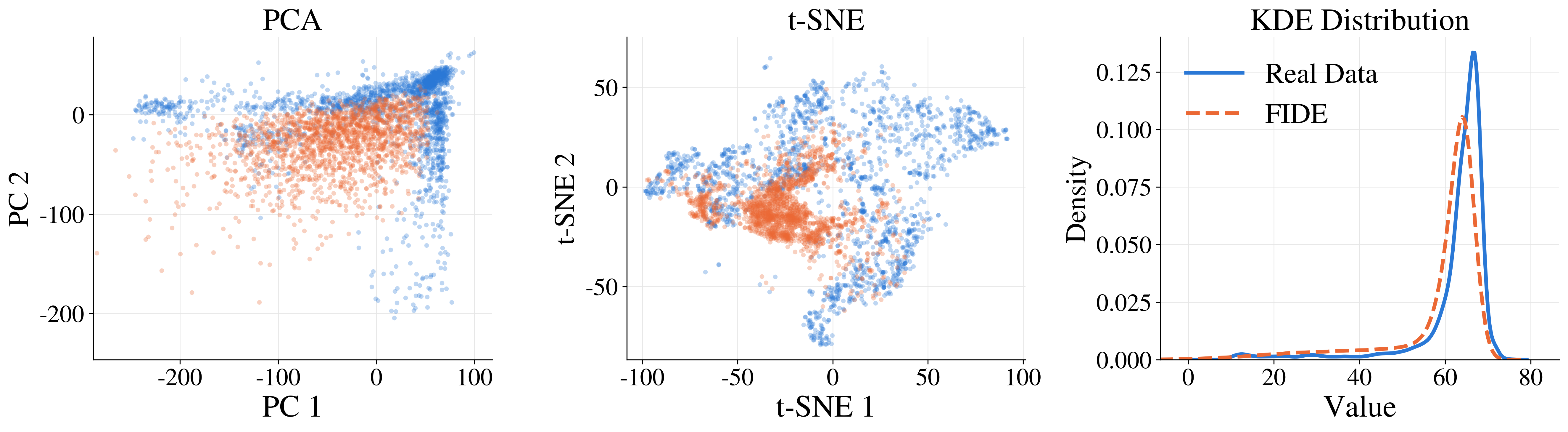}\\[-0pt]{\small FIDE}
\end{minipage}\\[8pt]
\caption{Distribution visualization on the Traffic dataset. Each panel triplet shows the PCA projection, t-SNE embedding, and kernel density estimate for one method against the real data.}
\label{fig:vis_traffic}
\end{figure}

\clearpage

\begin{figure}[p]
\centering
\begin{minipage}[t]{\textwidth}\centering
\includegraphics[width=0.9\linewidth]{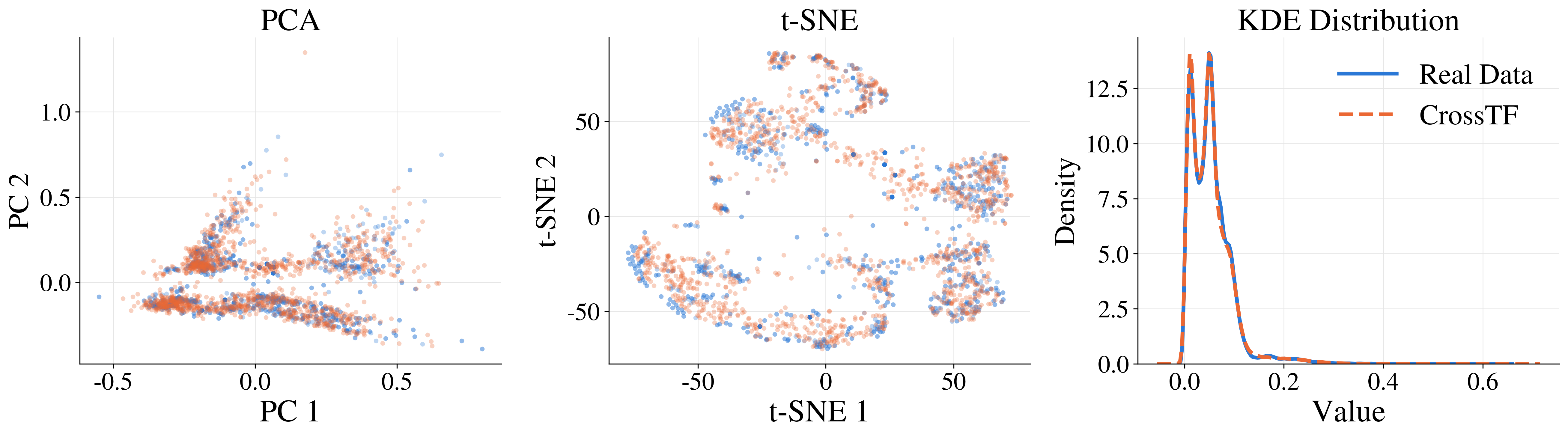}\\[-0pt]{\small \textbf{CrossTF (Ours)}}
\end{minipage}\\[8pt]
\begin{minipage}[t]{\textwidth}\centering
\includegraphics[width=0.9\linewidth]{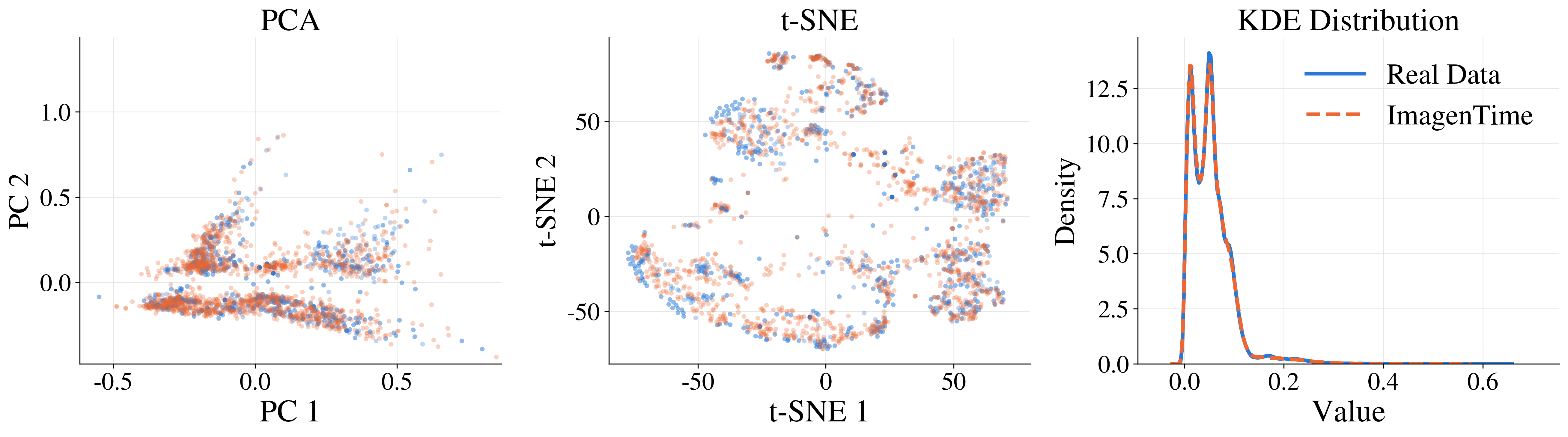}\\[-0pt]{\small ImagenTime}
\end{minipage}\\[8pt]
\begin{minipage}[t]{\textwidth}\centering
\includegraphics[width=0.9\linewidth]{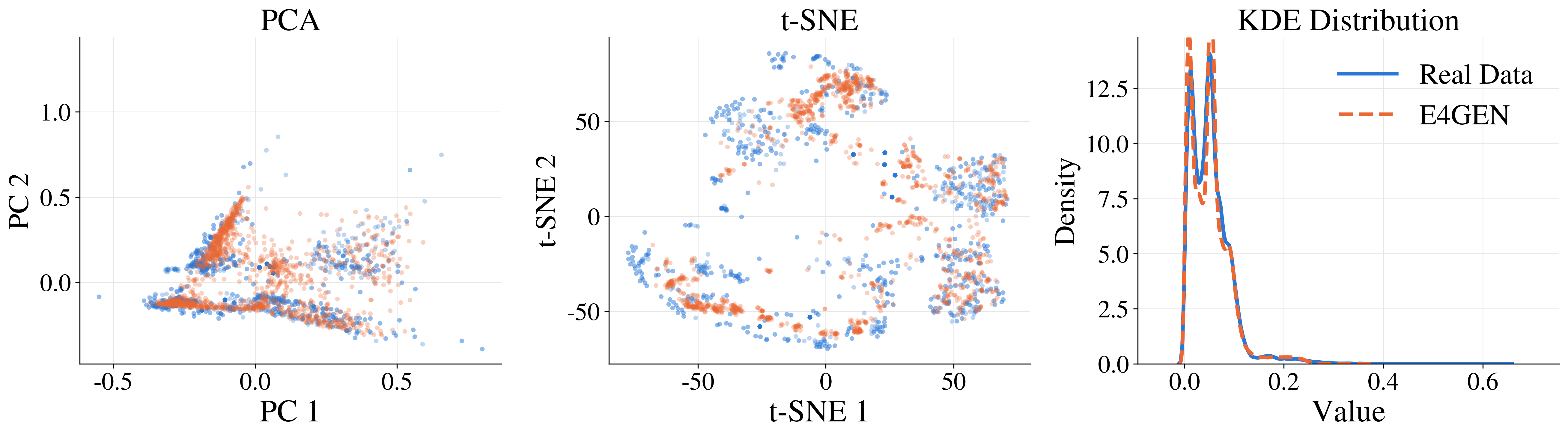}\\[-0pt]{\small E4GEN}
\end{minipage}\\[8pt]
\begin{minipage}[t]{\textwidth}\centering
\includegraphics[width=0.9\linewidth]{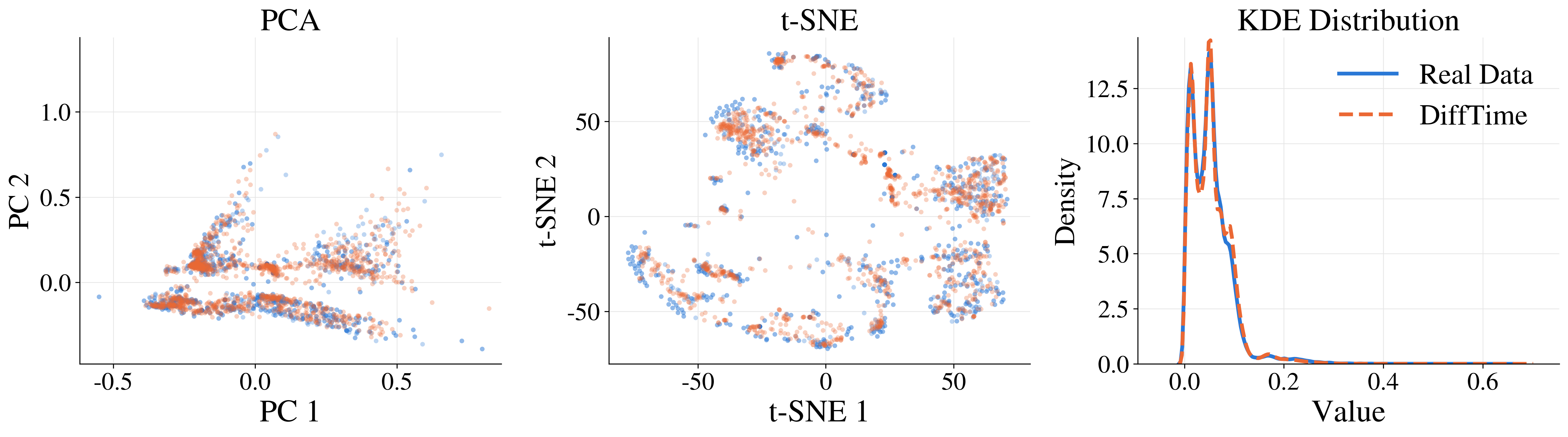}\\[-0pt]{\small DiffTime}
\end{minipage}\\[8pt]
\begin{minipage}[t]{\textwidth}\centering
\includegraphics[width=0.9\linewidth]{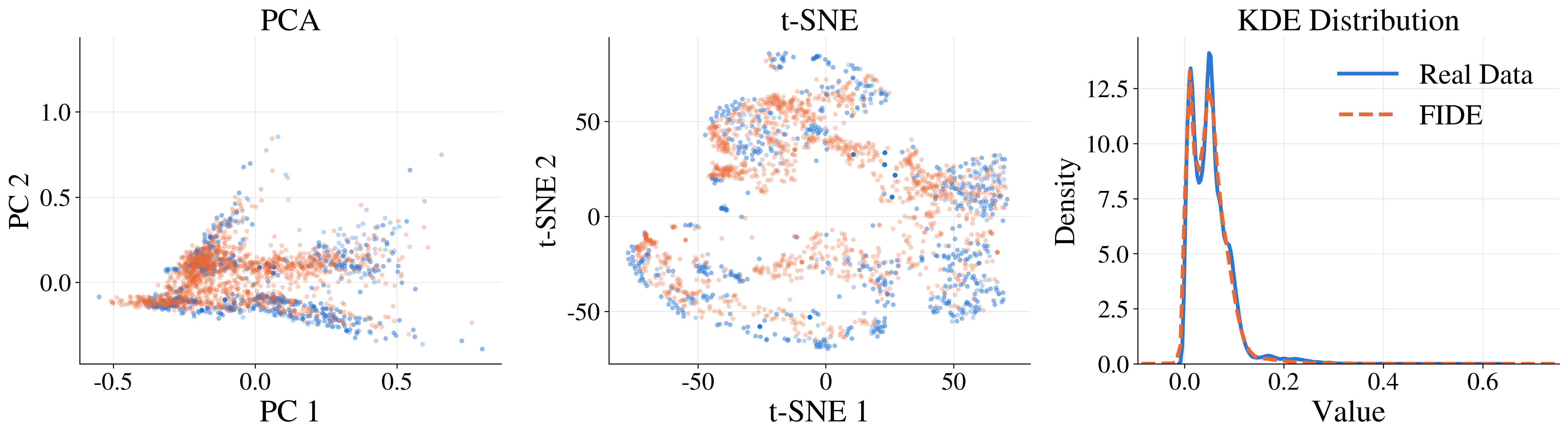}\\[-0pt]{\small FIDE}
\end{minipage}\\[8pt]
\caption{Distribution visualization on the Transportation dataset. Each panel triplet shows the PCA projection, t-SNE embedding, and kernel density estimate for one method against the real data.}
\label{fig:vis_transportation}
\end{figure}

\clearpage

\begin{figure}[p]
\centering
\begin{minipage}[t]{\textwidth}\centering
\includegraphics[width=0.9\linewidth]{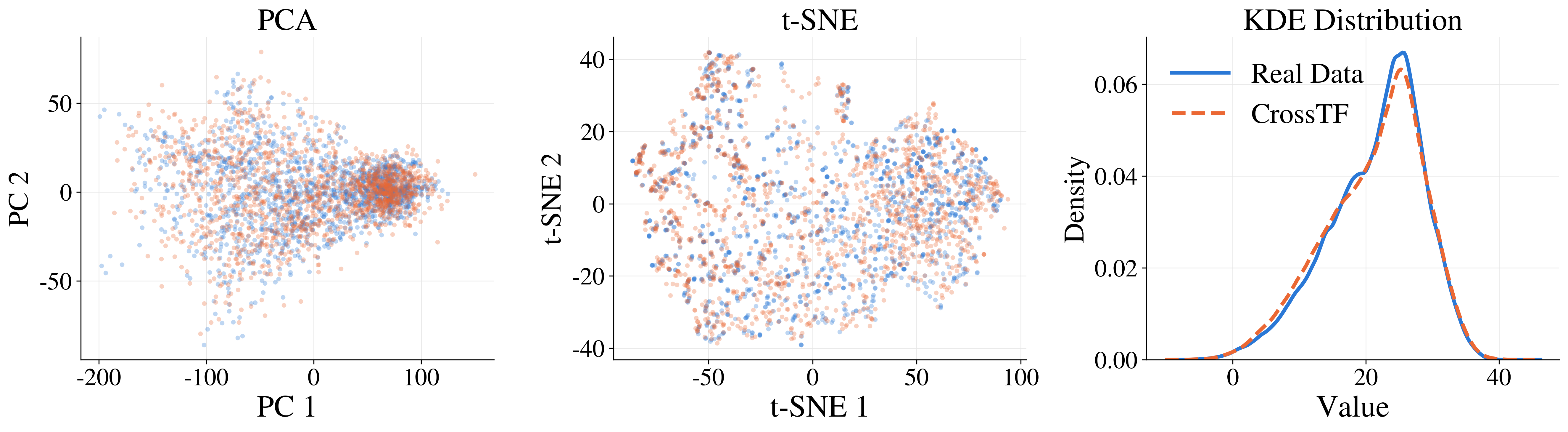}\\[-0pt]{\small \textbf{CrossTF (Ours)}}
\end{minipage}\\[8pt]
\begin{minipage}[t]{\textwidth}\centering
\includegraphics[width=0.9\linewidth]{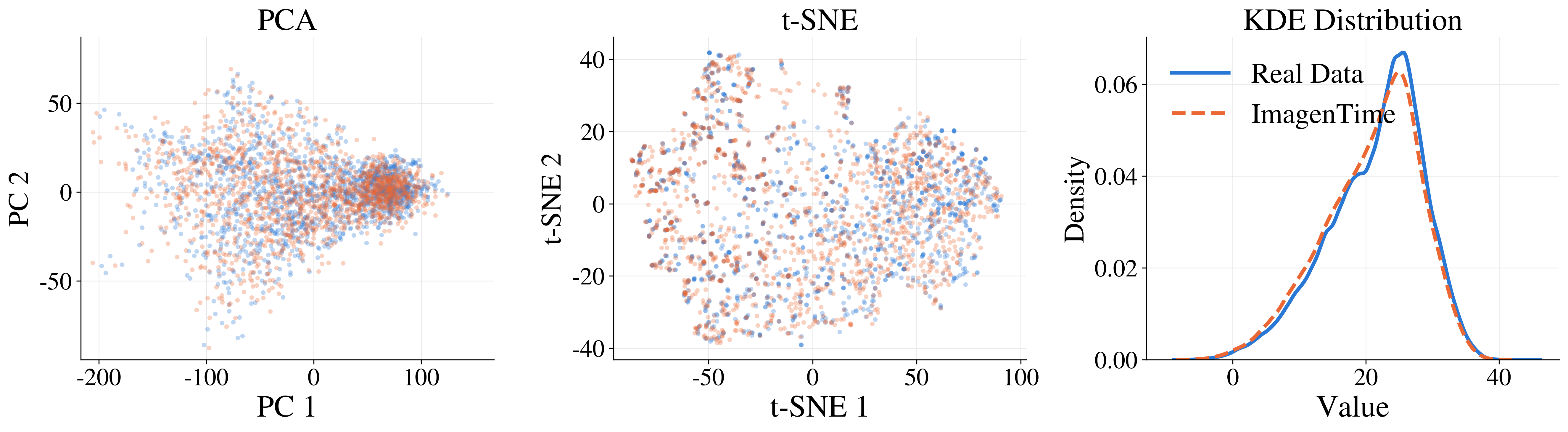}\\[-0pt]{\small ImagenTime}
\end{minipage}\\[8pt]
\begin{minipage}[t]{\textwidth}\centering
\includegraphics[width=0.9\linewidth]{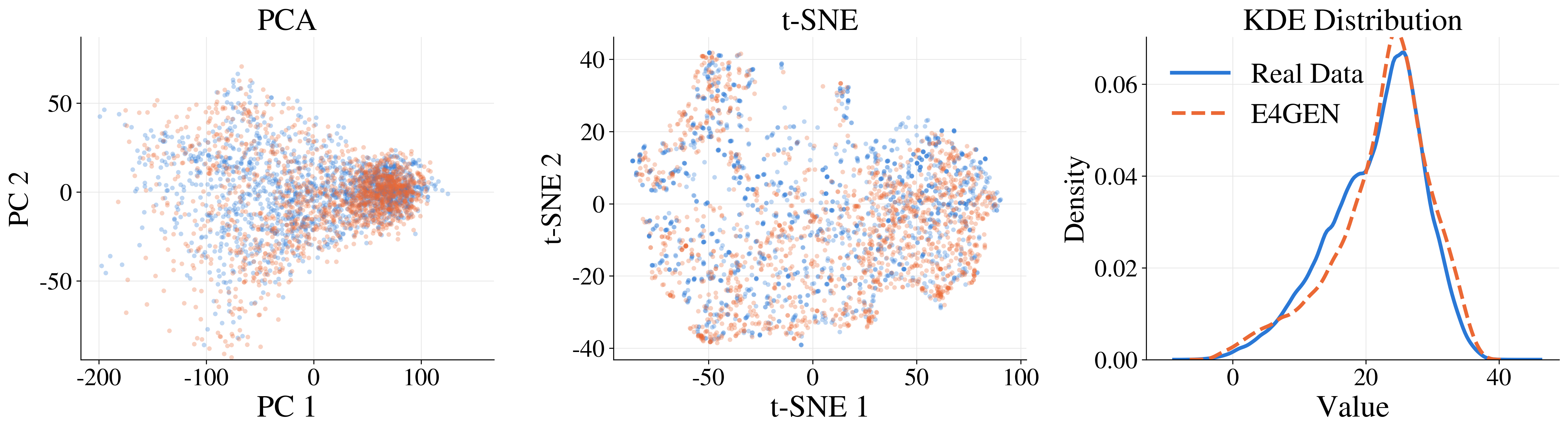}\\[-0pt]{\small E4GEN}
\end{minipage}\\[8pt]
\begin{minipage}[t]{\textwidth}\centering
\includegraphics[width=0.9\linewidth]{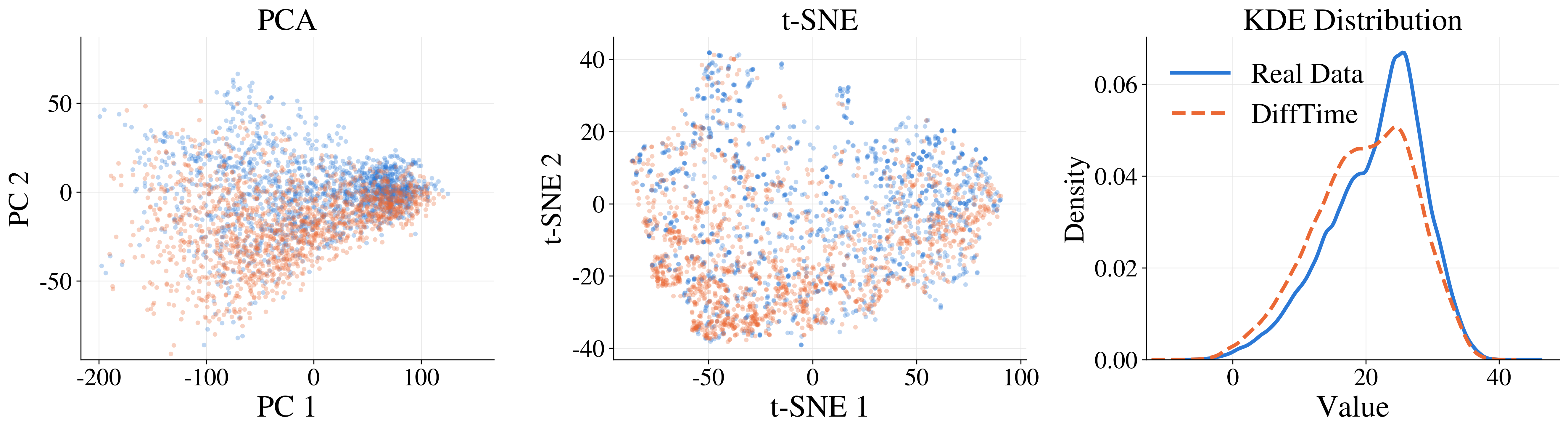}\\[-0pt]{\small DiffTime}
\end{minipage}\\[8pt]
\begin{minipage}[t]{\textwidth}\centering
\includegraphics[width=0.9\linewidth]{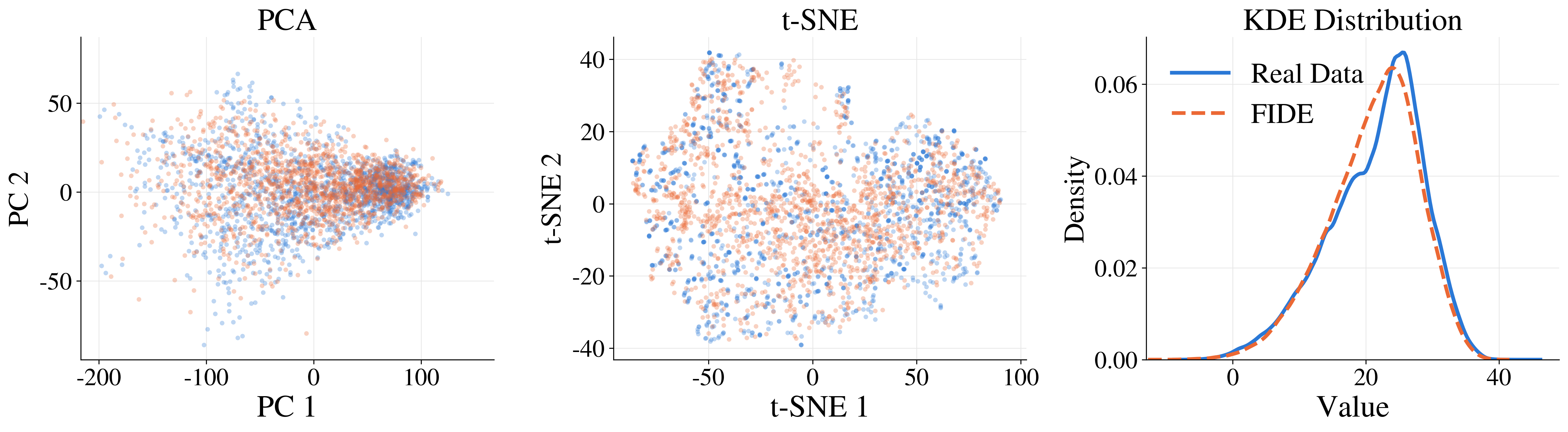}\\[-0pt]{\small FIDE}
\end{minipage}\\[8pt]
\caption{Distribution visualization on the Temperature dataset. Each panel triplet shows the PCA projection, t-SNE embedding, and kernel density estimate for one method against the real data.}
\label{fig:vis_temperature}
\end{figure}

\clearpage

\begin{figure}[p]
\centering
\begin{minipage}[t]{\textwidth}\centering
\includegraphics[width=0.9\linewidth]{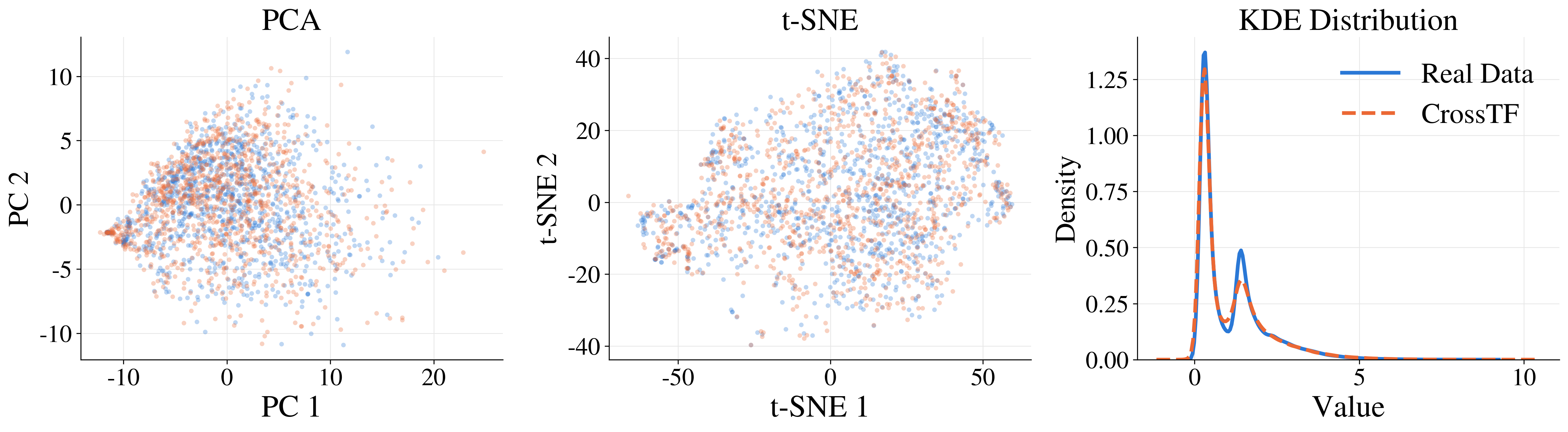}\\[-0pt]{\small \textbf{CrossTF (Ours)}}
\end{minipage}\\[8pt]
\begin{minipage}[t]{\textwidth}\centering
\includegraphics[width=0.9\linewidth]{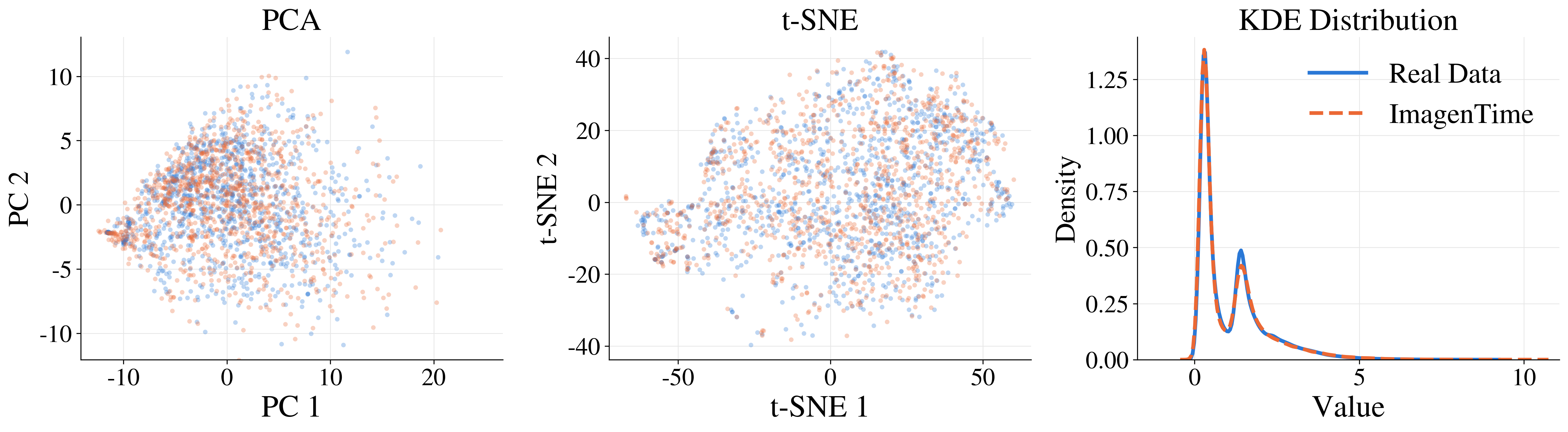}\\[-0pt]{\small ImagenTime}
\end{minipage}\\[8pt]
\begin{minipage}[t]{\textwidth}\centering
\includegraphics[width=0.9\linewidth]{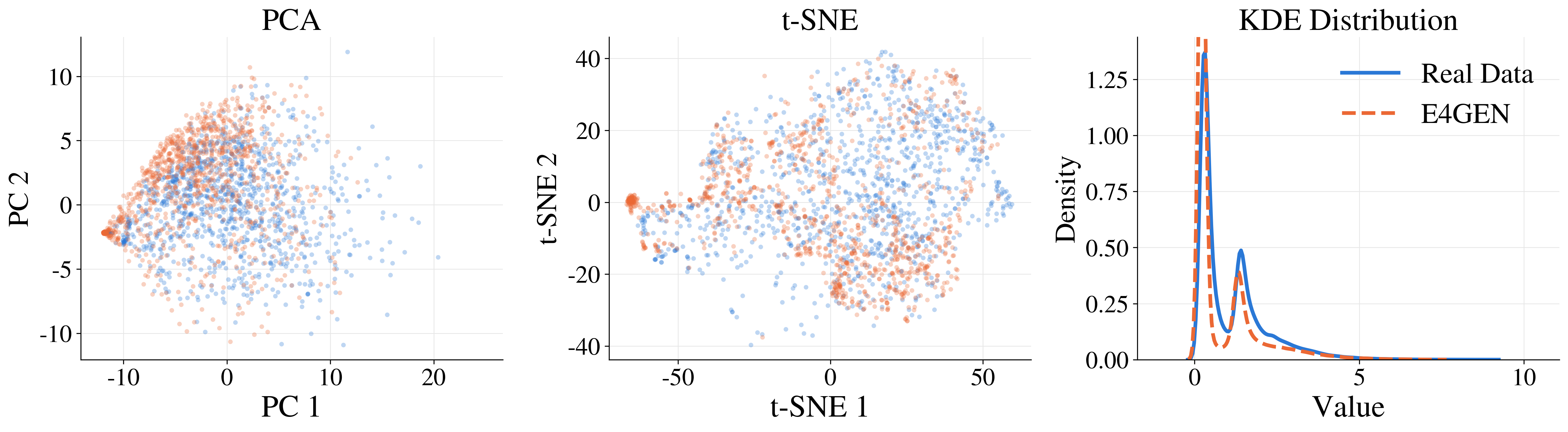}\\[-0pt]{\small E4GEN}
\end{minipage}\\[8pt]
\begin{minipage}[t]{\textwidth}\centering
\includegraphics[width=0.9\linewidth]{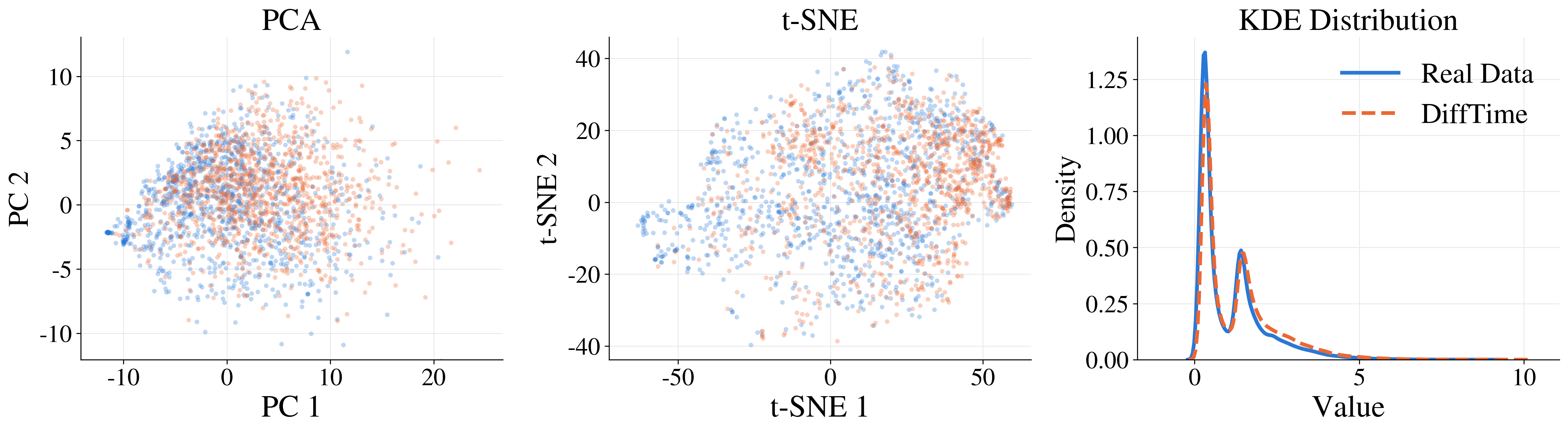}\\[-0pt]{\small DiffTime}
\end{minipage}\\[8pt]
\begin{minipage}[t]{\textwidth}\centering
\includegraphics[width=0.9\linewidth]{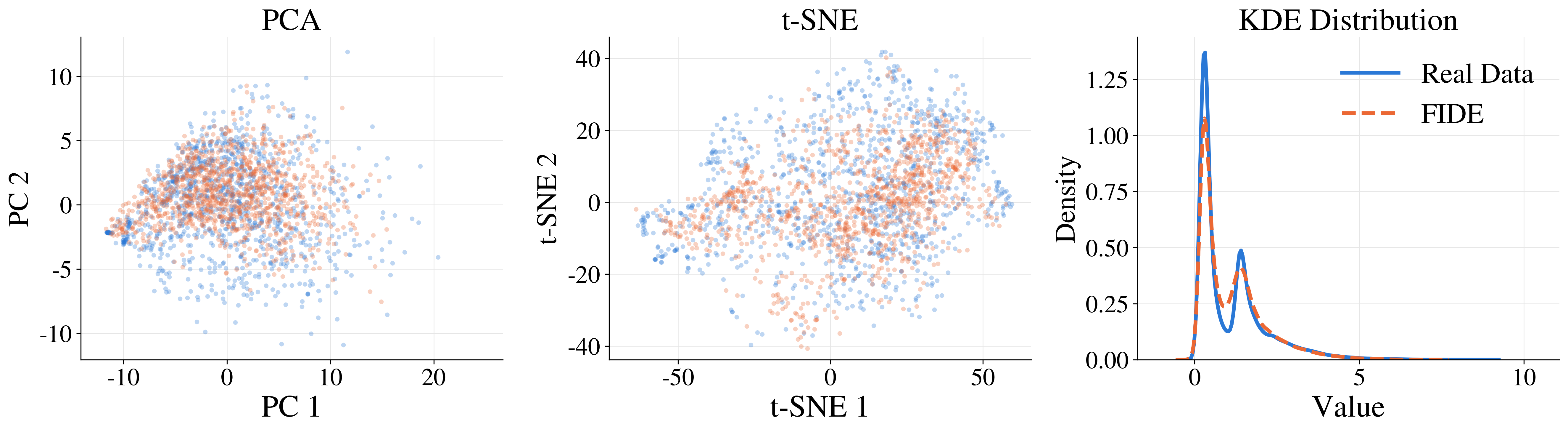}\\[-0pt]{\small FIDE}
\end{minipage}\\[8pt]
\caption{Distribution visualization on the Electricity dataset. Each panel triplet shows the PCA projection, t-SNE embedding, and kernel density estimate for one method against the real data.}
\label{fig:vis_electricity}
\end{figure}

\clearpage

\begin{figure}[p]
\centering
\begin{minipage}[t]{\textwidth}\centering
\includegraphics[width=0.9\linewidth]{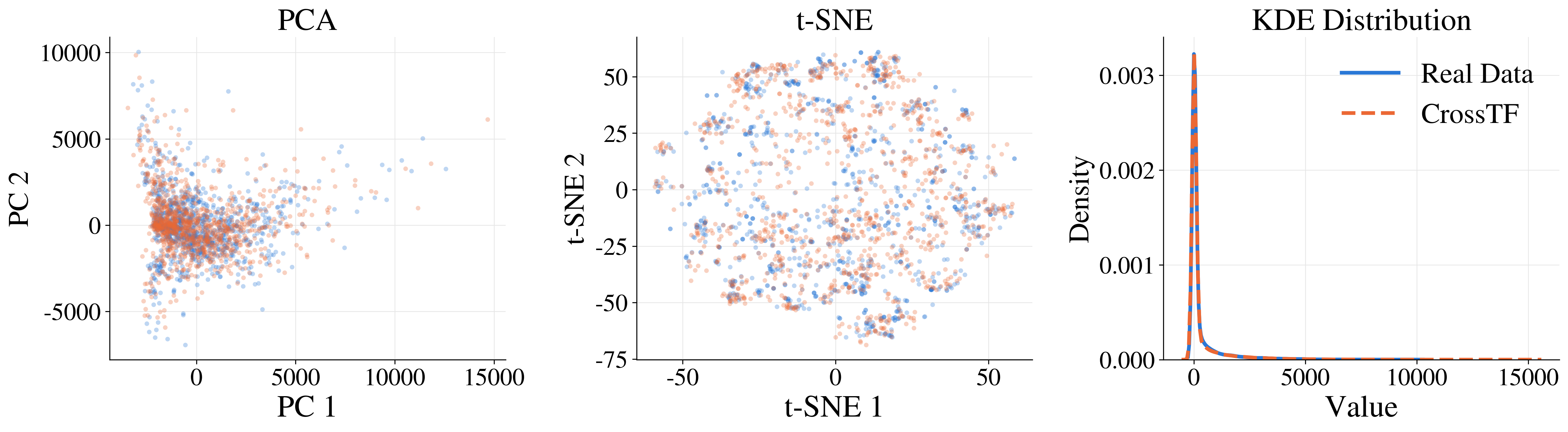}\\[-0pt]{\small \textbf{CrossTF (Ours)}}
\end{minipage}\\[8pt]
\begin{minipage}[t]{\textwidth}\centering
\includegraphics[width=0.9\linewidth]{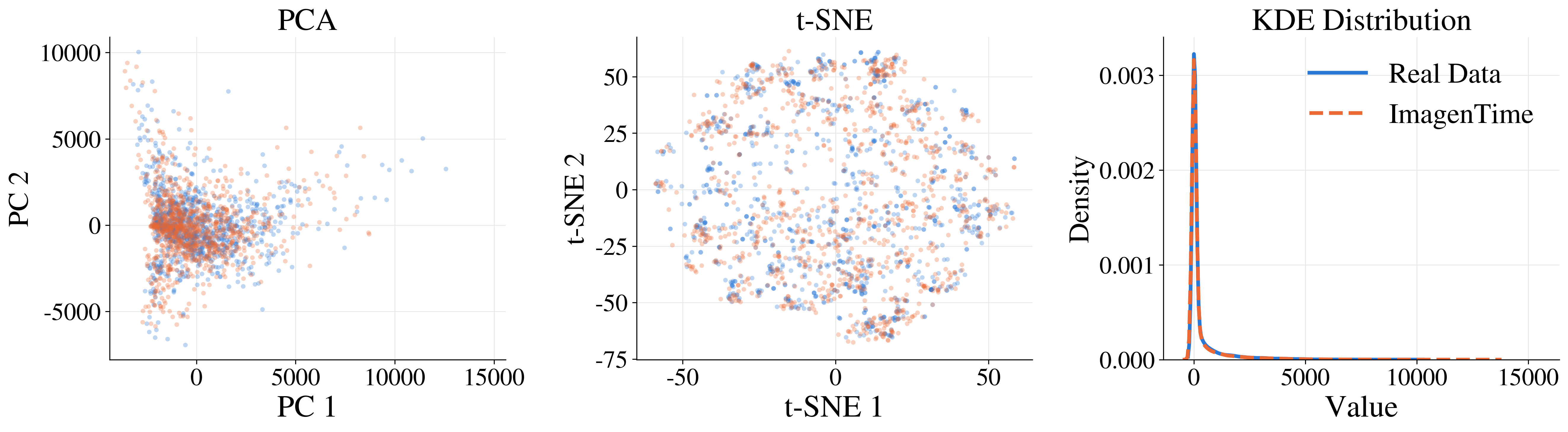}\\[-0pt]{\small ImagenTime}
\end{minipage}\\[8pt]
\begin{minipage}[t]{\textwidth}\centering
\includegraphics[width=0.9\linewidth]{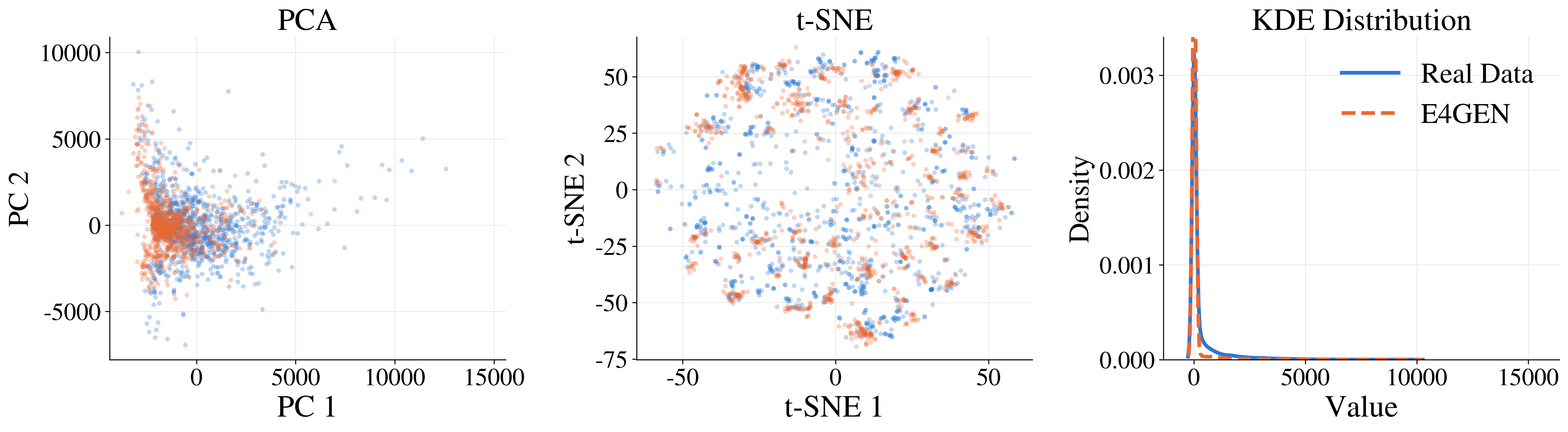}\\[-0pt]{\small E4GEN}
\end{minipage}\\[8pt]
\begin{minipage}[t]{\textwidth}\centering
\includegraphics[width=0.9\linewidth]{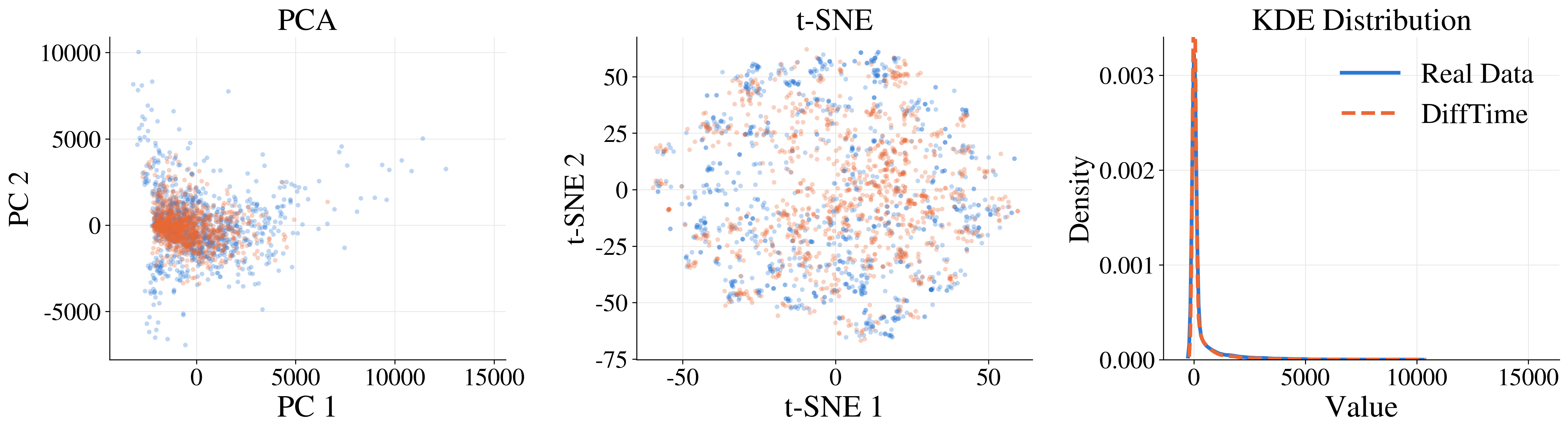}\\[-0pt]{\small DiffTime}
\end{minipage}\\[8pt]
\begin{minipage}[t]{\textwidth}\centering
\includegraphics[width=0.9\linewidth]{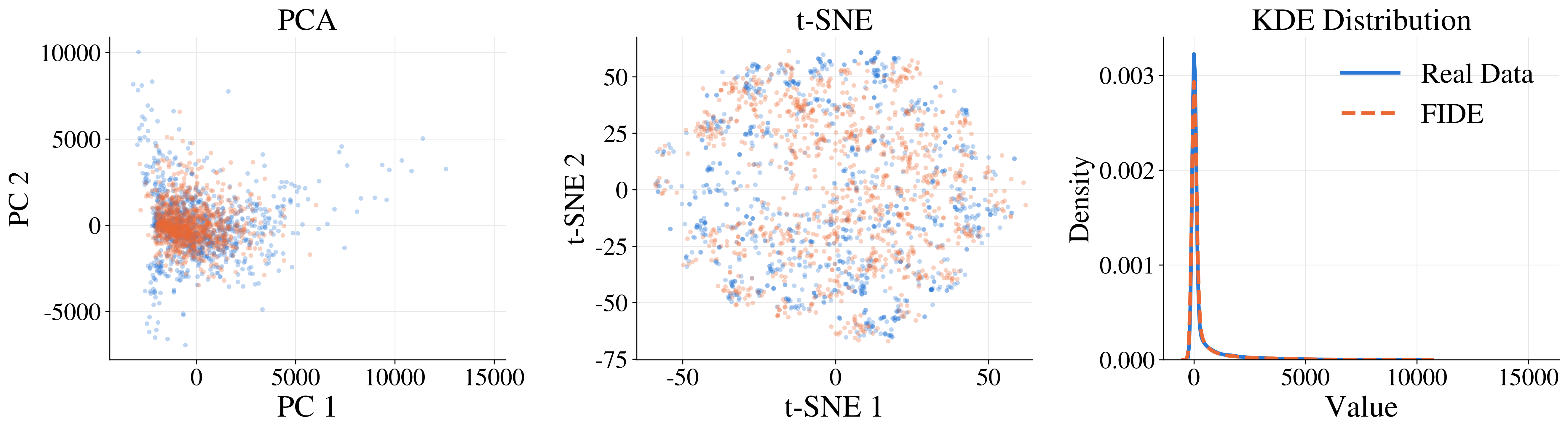}\\[-0pt]{\small FIDE}
\end{minipage}\\[8pt]
\caption{Distribution visualization on the Precipitation dataset. Each panel triplet shows the PCA projection, t-SNE embedding, and kernel density estimate for one method against the real data.}
\label{fig:vis_precipitation}
\end{figure}

\clearpage

\begin{figure}[p]
\centering
\begin{minipage}[t]{\textwidth}\centering
\includegraphics[width=0.9\linewidth]{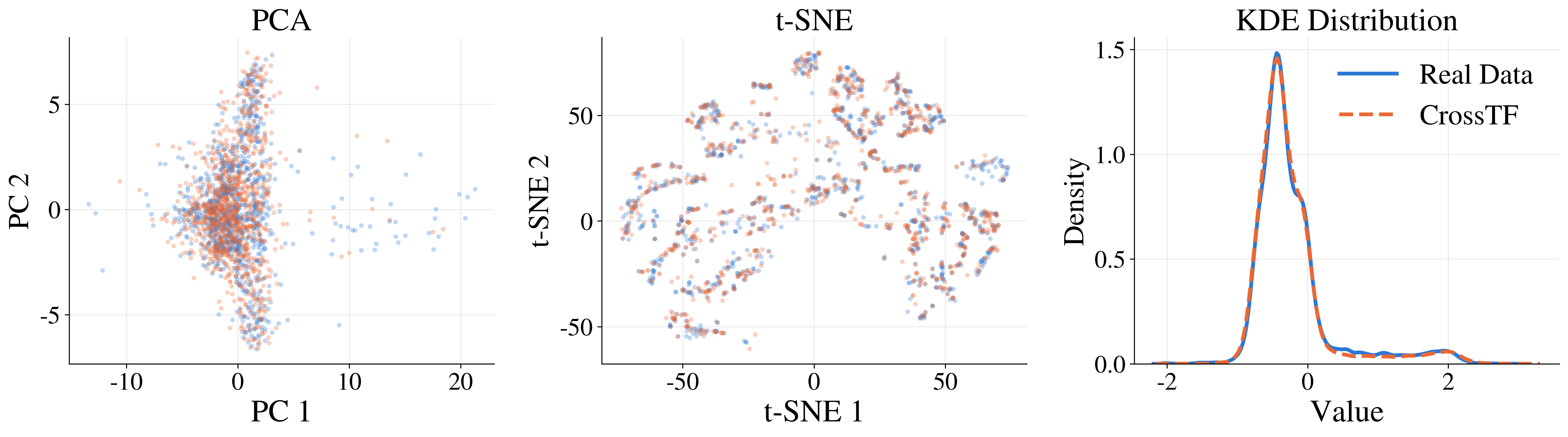}\\[-0pt]{\small \textbf{CrossTF (Ours)}}
\end{minipage}\\[8pt]
\begin{minipage}[t]{\textwidth}\centering
\includegraphics[width=0.9\linewidth]{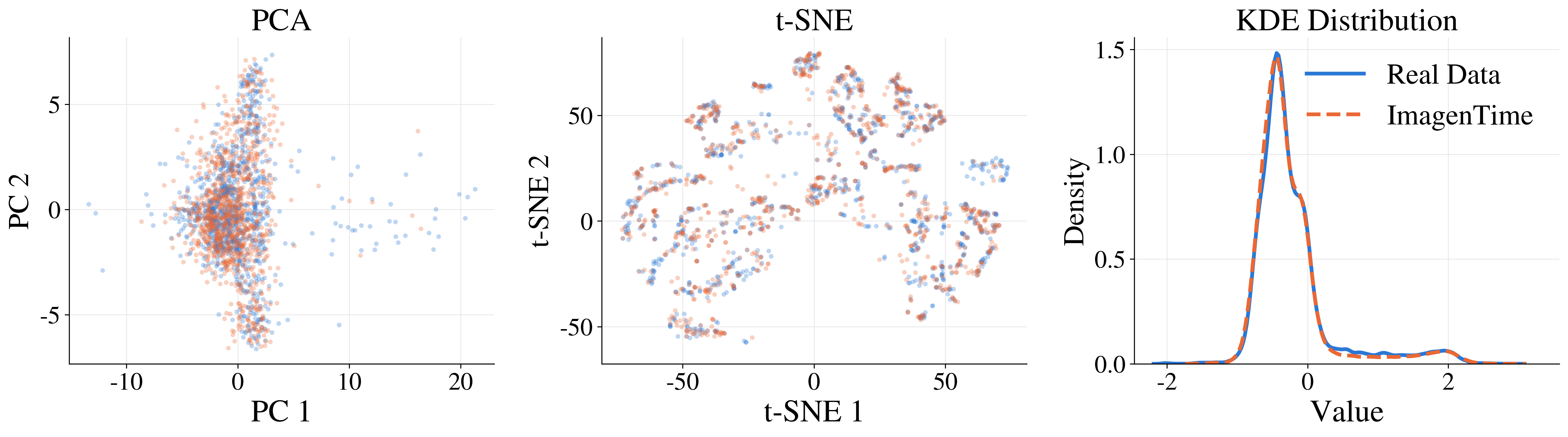}\\[-0pt]{\small ImagenTime}
\end{minipage}\\[8pt]
\begin{minipage}[t]{\textwidth}\centering
\includegraphics[width=0.9\linewidth]{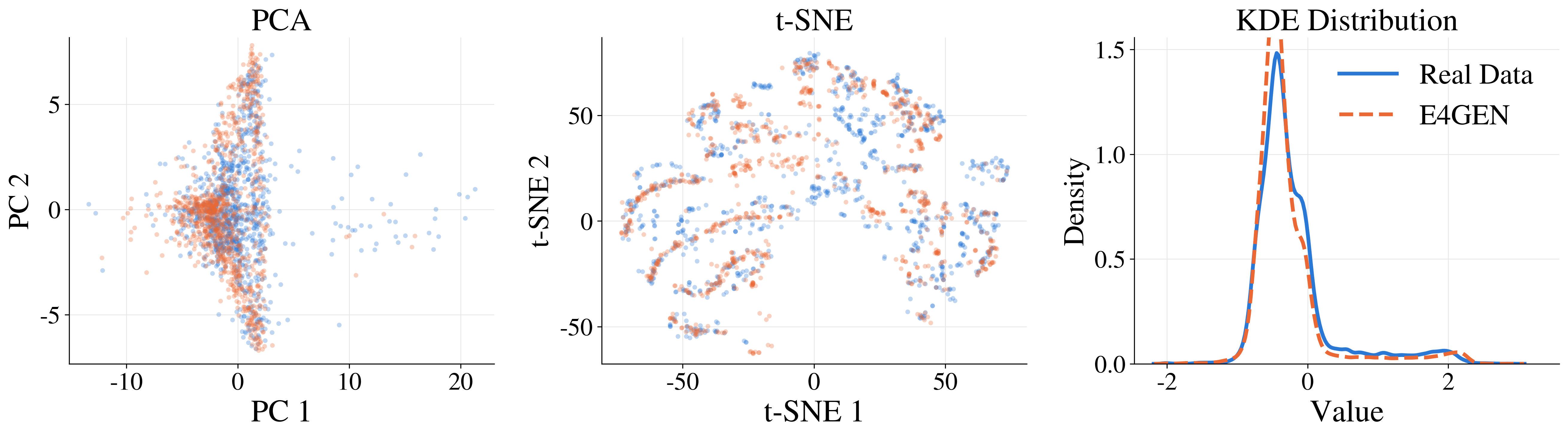}\\[-0pt]{\small E4GEN}
\end{minipage}\\[8pt]
\begin{minipage}[t]{\textwidth}\centering
\includegraphics[width=0.9\linewidth]{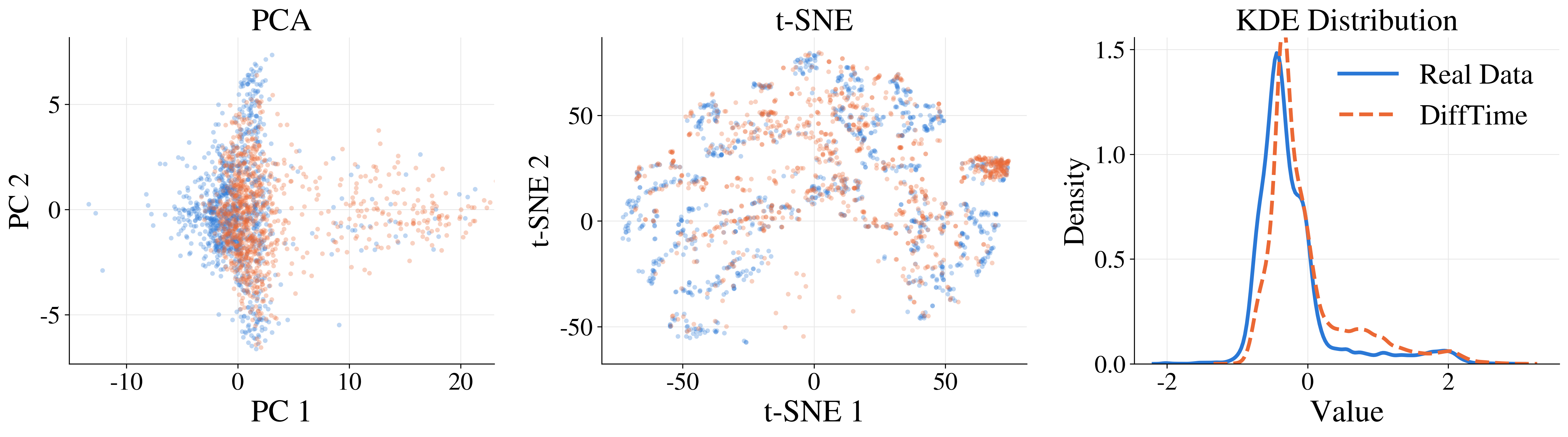}\\[-0pt]{\small DiffTime}
\end{minipage}\\[8pt]
\begin{minipage}[t]{\textwidth}\centering
\includegraphics[width=0.9\linewidth]{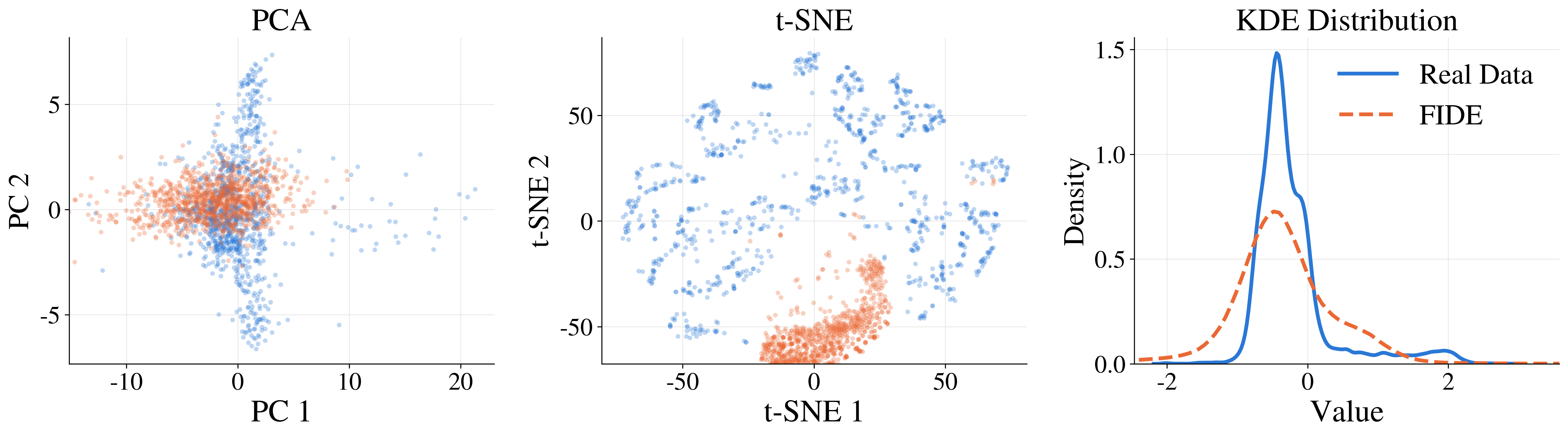}\\[-0pt]{\small FIDE}
\end{minipage}\\[8pt]
\caption{Distribution visualization on the ECG dataset. Each panel triplet shows the PCA projection, t-SNE embedding, and kernel density estimate for one method against the real data.}
\label{fig:vis_ecg}
\end{figure}

\clearpage

\begin{figure}[p]
\centering
\begin{minipage}[t]{\textwidth}\centering
\includegraphics[width=0.9\linewidth]{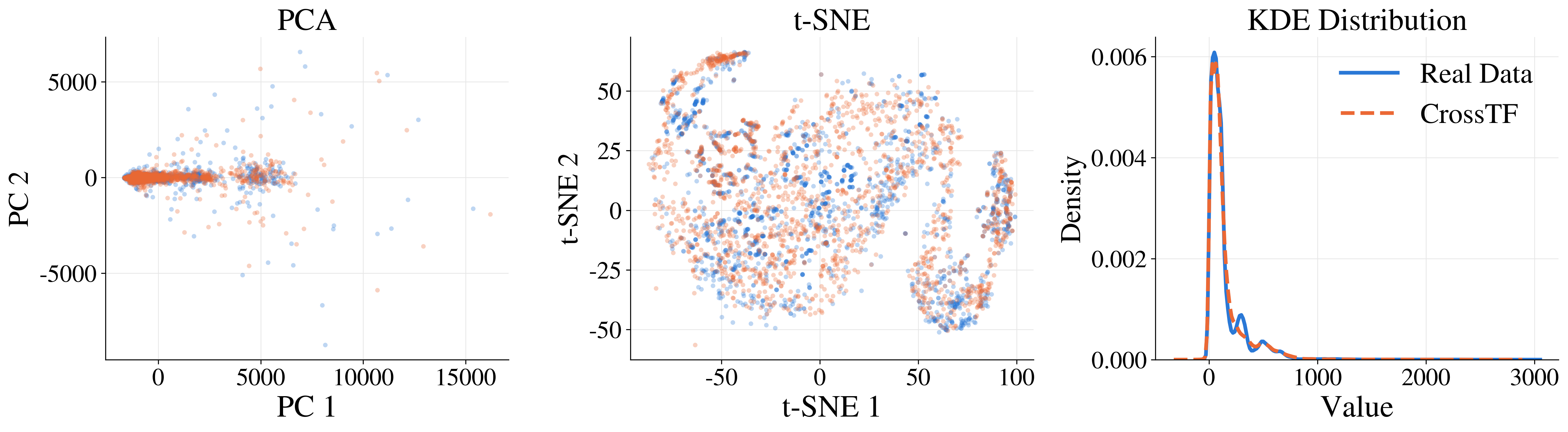}\\[-0pt]{\small \textbf{CrossTF (Ours)}}
\end{minipage}\\[8pt]
\begin{minipage}[t]{\textwidth}\centering
\includegraphics[width=0.9\linewidth]{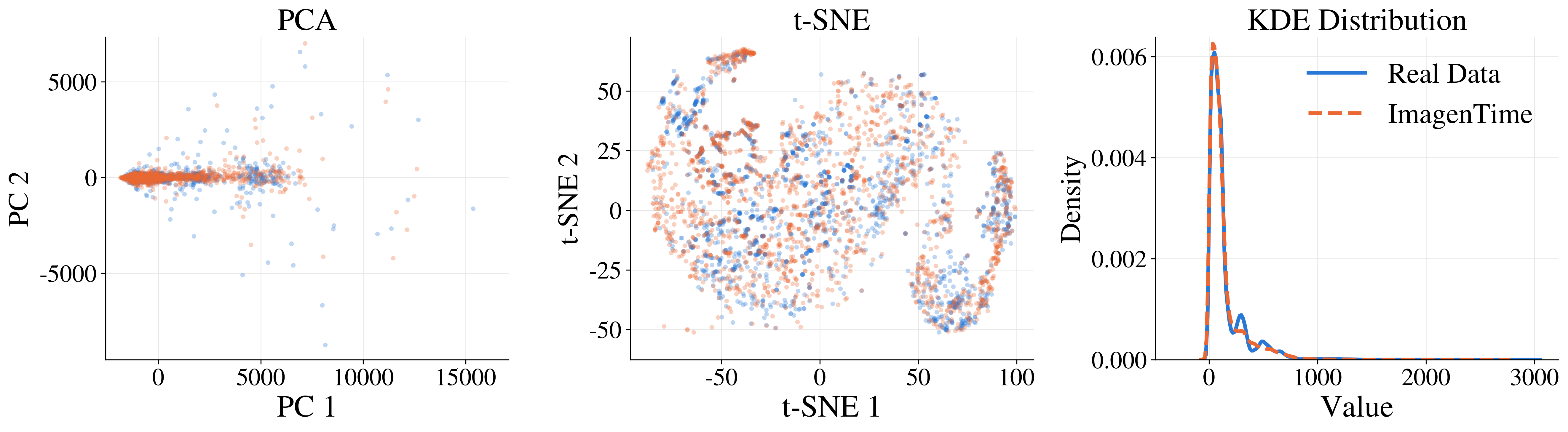}\\[-0pt]{\small ImagenTime}
\end{minipage}\\[8pt]
\begin{minipage}[t]{\textwidth}\centering
\includegraphics[width=0.9\linewidth]{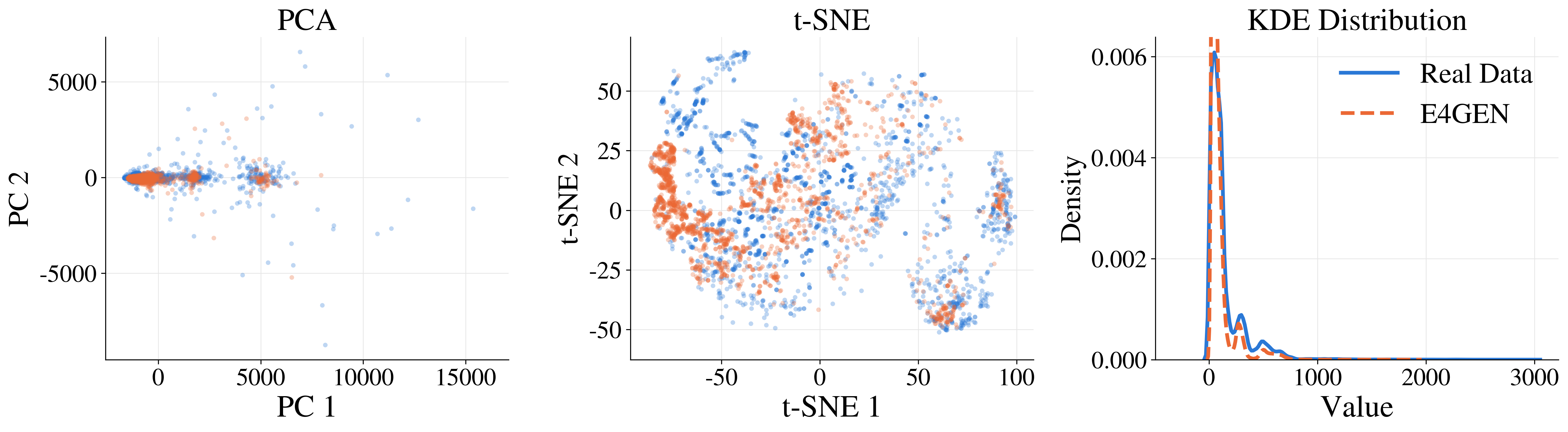}\\[-0pt]{\small E4GEN}
\end{minipage}\\[8pt]
\begin{minipage}[t]{\textwidth}\centering
\includegraphics[width=0.9\linewidth]{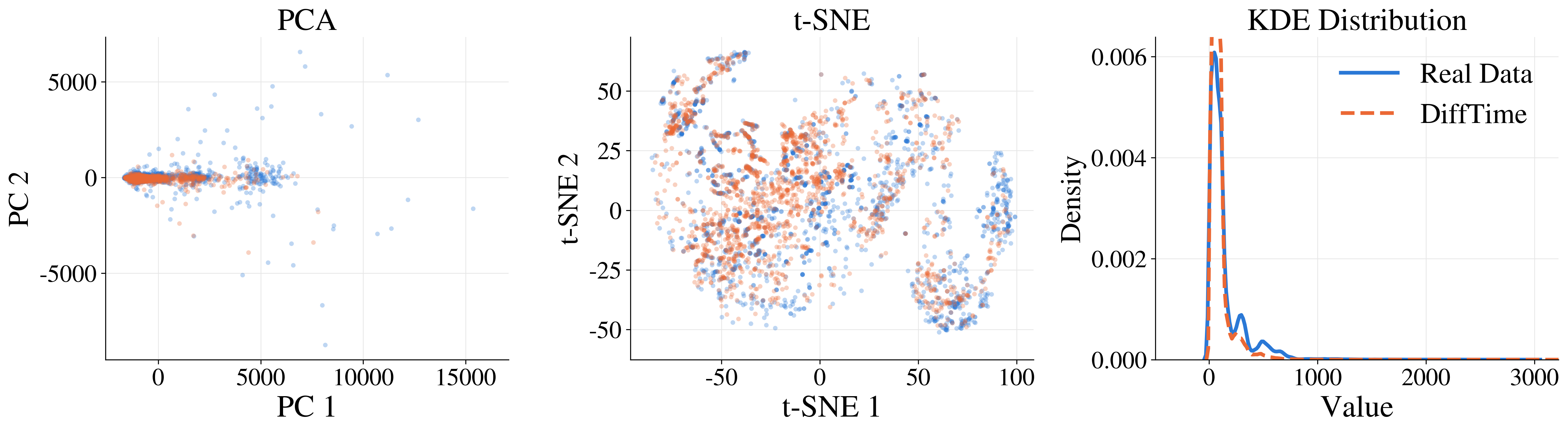}\\[-0pt]{\small DiffTime}
\end{minipage}\\[8pt]
\begin{minipage}[t]{\textwidth}\centering
\includegraphics[width=0.9\linewidth]{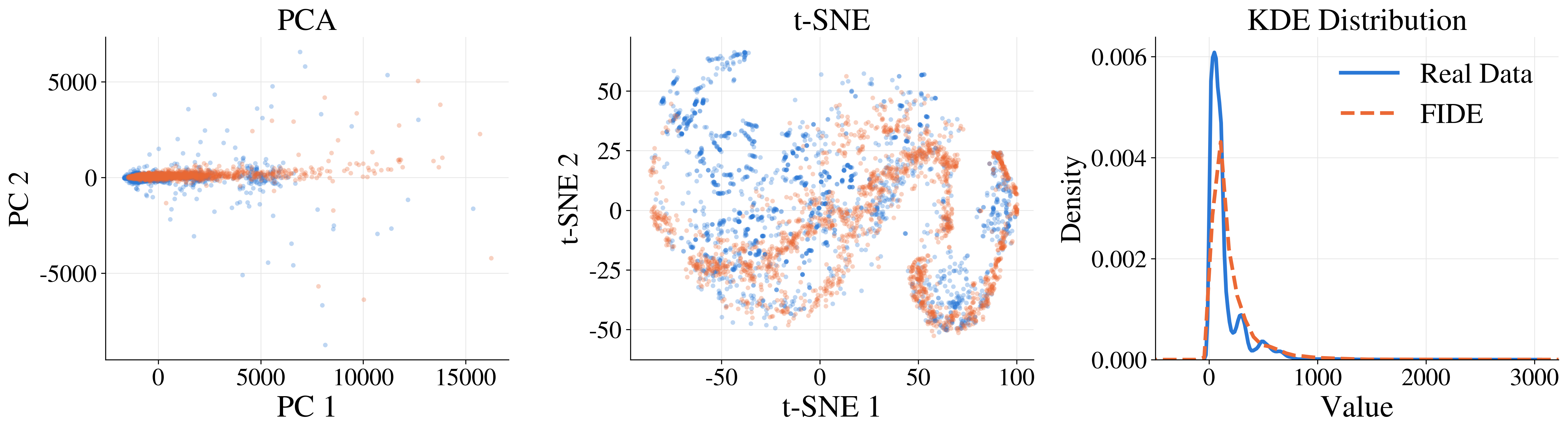}\\[-0pt]{\small FIDE}
\end{minipage}\\[8pt]
\caption{Distribution visualization on the Building dataset. Each panel triplet shows the PCA projection, t-SNE embedding, and kernel density estimate for one method against the real data.}
\label{fig:vis_building}
\end{figure}

\clearpage

\end{document}